%% file: v1.tex
\documentclass{article} % For LaTeX2e
\usepackage{iclr2024_conference,times}
\usepackage{mdframed}
\usepackage{titlecaps}
\input{math_commands.tex}

\title{MindAgent: Emergent Gaming Interaction}
\author{
Ran Gong$^{1\dagger*}$, 
Qiuyuan Huang$^{2\ddagger*}$, 
Xiaojian Ma$^{1*}$, 
Hoi Vo$^{3}$, 
Zane Durante$^{4\dagger}$,
\textbf{Yusuke Noda}$^{3}$,\\ 
\textbf{Zilong Zheng}$^5$,  
\textbf{Song-Chun Zhu}$^{1567}$, 
\textbf{Demetri Terzopoulos}$^{1}$, 
\textbf{Li Fei-Fei}$^4$, 
\textbf{Jianfeng Gao}$^{2}$ \\
$^1$UCLA; $^2$Microsoft Research, Redmond; $^3$Xbox Team, Microsoft; $^4$Stanford;$^5$BIGAI; $^6$PKU; $^7$THU
}

\newcommand\blfootnote[1]{%
  \begingroup
  \renewcommand\thefootnote{}\footnote{#1}%
  \addtocounter{footnote}{-1}%
  \endgroup
}
\iclrfinalcopy % Uncomment for camera-ready version, but NOT for submission.
\input{config}

\begin{document}

\maketitle
\begin{center}
    \centering
    \vspace{-15pt}
    \captionsetup{type=figure}
    \includegraphics[width=0.95\linewidth]{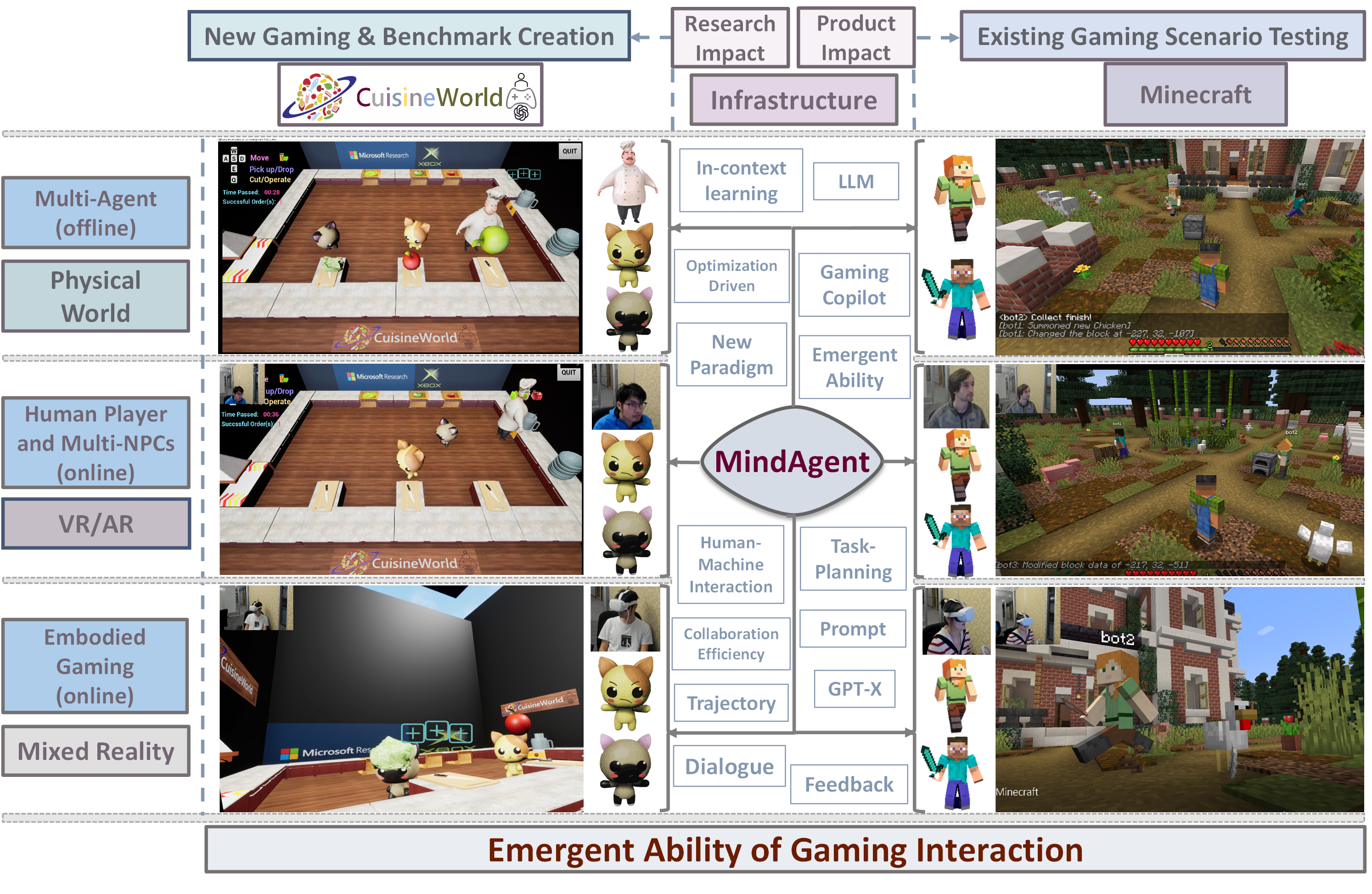}
    \captionof{figure}{ The \mindagent system for gaming interactions. \mindagent enables complex task planning in a multi-agent system and human-AI collaborated infrastructure across different domains.
    }
    \label{fig:game_overview}
\end{center}

\begin{abstract}
\vspace{-3mm}
%We find that pre-trained 
Large Language Models (LLMs) have the capacity of performing complex scheduling in a multi-agent system and can coordinate these agents into completing sophisticated tasks that require extensive collaboration.
However, despite the introduction of numerous gaming frameworks, the community has insufficient benchmarks towards building general multi-agents collaboration infrastructure that encompass both LLM and human-NPCs collaborations. In this work, we propose a novel infrastructure - \textbf{MindAgent} - to evaluate planning and coordination emergent capabilities for gaming interaction. In particular, our infrastructure leverages existing gaming framework, to i) require understanding of the coordinator for a multi-agent system, ii) collaborate with human players via un-finetuned proper instructions, and iii) establish an in-context learning on few-shot prompt with feedback.
%We establish a new multi-agent collaboration benchmark, \textbf{\overcook}, 
% in which several agents are placed inside a virtual kitchen together with cooking tools and ingredients and must work together to complete as many orders as possible within a limited period of time. 
% Zane: I commented out these lines and removed them since I think it is too much detail for the abstract
%The goal is to fulfill as many incoming orders as possible in a limited period of time. 
%With new orders flooding in and old ones expiring, these robots have to properly coordinate to maximize overall productivity. 
%which to evaluate planning and coordination capabilities, 
%Our infrastructure dispatchs and supervise multiple agents playing the game simultaneously. 
Furthermore, we introduce \textbf{\textsc{CuisineWorld}}, a new gaming scenario and related benchmark that dispatch a multi-agent collaboration efficiency and supervise multiple agents playing the game simultaneously.
We conduct comprehensive evaluations with new auto-metric \textit{collaboration score} \textbf{CoS} for calculating the collaboration efficiency. 
%Our results indicate that our infrastructure can serve as a coordinator for a considerable amount of agents, and even collaborate with human players without any additional fine-tuning when they are prompted with proper instructions, few-shot examples (via \textit{in-context learning}), and game feedback. 
%Further, we show that LLMs exhibit impressive generalization capabilities by being able to coordinate a large number of agents when only given example of a small number, and adaptation to novel game domains.
%%% -----Qiuyuan stop here at 5pm Sep.09
Finally, our infrastructure can be deployed into real-world gaming scenarios in a customized VR version of \textbf{\textsc{CuisineWorld}} and adapted in existing broader ``Minecraft'' gaming domain. %Even deploying them in real-world game systems at scale can still be challenging due to computation costs, context length limitations, \etc. %We hope our findings can pave the path for future gaming interaction systems.  
We hope our findings on LLMs and the new infrastructure for general-purpose scheduling and coordination can help shed light on how such skills can be obtained by learning from large language corpora. Project webpage:  \href{https://mindagent.github.io}{https://mindagent.github.io}.
\end{abstract}
%%%%%%%%%%%%%%%
\blfootnote{$*$ Equal Contribution. $\ddagger$ Project Leader. \\$\dagger$ Work done while Ran and Zane interning at Microsoft Research, Redmond. }
%%%%%%%%%%%%%%
\section{Introduction}
\label{sec:Introduction}

% -intro
% 1. LLM as piloting many tasks, including planning, thanks to zero-shot and few-shot learning (ICL)
% 2. We further explore MA planning, explain what is MA planning, why is it important and its difference from existing planning scheme stuided with LLM
% 3. tech details briefing, the game, zero-shot and ICL based prompting
% 4. the result, inc. genearlization and collaborating with  humans
% 5. contributions

Large language Models (LLMs) have been piloting the effort of developing general intelligent machines\citep{bubeck2023sparks, mirchandani2023large} . Although they are trained in large text corpora, their superior problem-solving capacity is not limited to canonical language processing domains. 
% maybe we should remove this part
LLMs already demonstrate the potential to tackle complex tasks that were previously presumed exclusive to domain-specific algorithms or human experts, ranging from mathematical reasoning~\citep{imani2023mathprompter, wei2022chain, zhu2022solving} to answering questions of professional law~\citep{blair2023can, choi2023chatgpt, nay2022law} and medicine~\citep{nov2023putting, yang2023evaluations, jeblick2022chatgpt}. 
More recently, some research has shown the possibility of using LLMs to generate complex plans for robots and game AI~\citep{codeaspolicies2022, wang2023describe, wang2023voyager, yao2023react, huang2023ark}, marking an important milestone for LLMs as generalist intelligent agents.

In this work, we would like to further investigate the planning capacity of LLMs. Specifically, we are interested in planning in a multi-agent system~\citep{stone2000multiagent}, \ie {multi-agent planning}. Compared to planning for a single agent, which has been extensively studied by previous research~\citep{wang2023describe, wang2023voyager}, multi-agent planning imposes much higher problem-solving complexity due to the exponentially growing action space (\wrt number of agents). The planner has to simultaneously control multiple agents, avoid possible conflicts, and coordinate them into completing a shared goal that requires sophisticated collaborations. To understand to which extent can LLMs obtain multi-agent planning skills, we first establish a new benchmark, \textbf{\overcook} as illustrated in \autoref{fig:game_overview}. 

%It is a game that emulates a virtual kitchen, where several robots are commanded to use various cooking tools and ingredients to prepare as many dish orders as possible in a limited period of time. To facilitate collaboration, new orders will keep flooding in while the existing ones should be completed before expiration. Therefore, LLMs need to properly coordinate these robots to maximize overall productivity. \overcook also offers game levels with a wide range of planning difficulty: dishes with different complexity (number of ingredients and tools involved), number of agents, order frequency and lifetime, etc, making it an ideal test bed for LLM-based multi-agent planning.

To incorporate agent AI into video games, we main design an infrastructure - \textbf{\mindagent} - inspired by multi-agent task allocation optimization theories to facilitate LLM multi-agent planning capabilities. Our infrastructure enables LLMs to perform complex coordination and scheduling with multiple different agents. 
% 1. zero-shot
% 2. prompting techniques
% 3. generalization and other perks
We conduct comprehensive evaluations with recently introduced LLMs playing our game with our infrastructure, including GPT-4, Claude, and LLaMA. Through the proposed \mindagent interactive multi-agent planning framework for LLMs, we make the following key observations: 1) \textbf{zero shot multi-agent planning}: Without bells and whistles, powerful pretrained LLMs like GPT-4 are capable of scheduling multiple agents (ranging from 2 to 4) into completing dishes, and even collaborate with human players, by merely reading simple game instructions and recipes; 2) \textbf{planning with advanced prompting}: We are able to significantly boost their multi-agent planning performances by leveraging the emergent \textbf{\textit{in-context learning}} capability~\citep{brown2020language, wei2021finetuned}: adding very few expert demonstrations even from different game levels to the prompt, explaining the rationale of certain actions as in Chain-of-Thought prompting~\citep{wei2022chain}, and providing on-the-fly feedback to the LLMs during planning; 3) \textbf{generalist potentials}: LLMs exhibits great potentials of being generalist multi-agent planner as it has strong generalization to coordinate more agents with examples of fewer agents, and adaptation to new game domains like Minecraft. 

While compared to canonical domain-specific automated planning systems, multi-agent planning with LLMs can still be bottlenecked by challenging computation cost, context length limitation, non-optimal plans, \etc, it has the potential of improving from data without fine-tuning (via \textit{in-context learning}), seamlessly adapting to planning problems from different domains and offering more flexible interfaces.
We hope our findings on LLMs for general-purpose scheduling and coordination can help shed some light on how such skills can be obtained by learning from large text corpora, and facilitate the emergence of better LLM planners.

To summarize, our key contributions are as follows:
% game
% prompting
% benchmarking and results
\setlength{\leftmargini}{0.85em}
\begin{itemize}[topsep=0.8pt]
 \setlength\itemsep{0.6pt}
 \item We establish a new gaming scenario and related benchmark based on a multi-agent virtual kitchen environment, \overcook. It adopts a minimal text-based game format and supports various planning task structures and difficulties, making it an ideal test bed for the emergent multi-agent planning (scheduling and coordination) capacity of LLMs.
 \item We introduce \mindagent, an infrastructure for interactive multi-agent planning with LLMs, which demonstrates the in-context learning multi-agent planning capacity of LLMs and brings several prompting techniques that help facilitate their planning ability, including providing few-shot demonstrations, planning rationals, and environmental feedback. 
 \item We conduct extensive evaluations with multiple LLMs and prompting settings on our benchmark. Experimental results confirm their potential on being generalist multi-agent planners in terms of generalizing to more agents. %and adapting to novel game domains.

 \item We deploy our system into real-world gaming scenarios and demonstrate its capabilities in human-AI interactions. 

\end{itemize}

%%%%%%%%%%%%%%%%%%%%%%%%
\section{Related Work}
\label{sec:RelatedWork}
% focused on collaboration efficiency, multiple tool, multiple task.  generative agents no evaluation metrics . 
\textbf{Multi-Agent Coordination.}~~The field of multi-agent collaborations boasts a comprehensive body of literature. Traditionally, such collaborations have been modeled using MDP/POMDP \citep{lowe2017multi, rashid2020monotonic, jain2019two} frameworks.

However, there has been a recent shift towards utilizing Large Language Models (LLMs) for these collaborations. For instance, \citet{zhang2023building} delved into how large language models might communicate and cooperate in a watch-and-help (WAH) task. Meanwhile, \citet{zhang2023proagent} investigated a two-agent collaboration game inspired by the simpler dynamics of the two-agent Overcooked-style game. Notably, their research chiefly concentrated on the task success rate, with most studies typically anchored to a singular task objective. In contrast, we emphasize the importance of collaboration efficiency in scenarios encompassing multiple task objectives. Further, our research uniquely focuses on evaluating the collaborative efficiency of more than two agents. Additionally, while other works like \citet{park2023generative} simulate each agent individually, we employ a centralized system. This approach not only significantly reduces the number of API calls but also reduces context length, making it more appropriate for gaming applications.

\textbf{Planning with LLMs.}~~There exists a number of works that leverage LLMs to perform task planning \citep{pmlr-v162-huang22a, wang2023voyager, yao2023react}. They leverage the LLMs' internet-scale domain knowledge and emergent zero-shot planning abilities to perform complex task planning and reasoning. Recent works in robotics also leverage LLMs to perform task planning, they decompose a natural language instruction into a sequence of subtasks, either in natural language form or in python code \citep{saycan2022arxiv, huang2022inner, codeaspolicies2022}. Then they use a low-level controller to execute these subtasks. Additionally, \citep{huang2022inner, codeaspolicies2022, wang2023describe} also incorporate environment feedback to improve task performance. 

\textbf{Benchmarks using Games.}~~
Numerous games have been developed to study task planning \cite{baker2022video, carroll2019utility}, yet only a handful delve into multi-agent collaborations. Even within this limited subset, the focus predominantly remains on two-agent interactions where responsibilities are not evenly distributed. As evidenced by \citep{ wan2022handmethat, puig2020watch}, it's common for one player to assume a dominant role while the other provides support. In contrast, our paper assumes equal responsibilities across agents, and we expand our investigation to encompass collaborations involving more than just two agents, even with human players. While some previous studies have ventured into multi-task settings, none have delved into scenarios where agents must complete multiple distinct tasks using competing resources within a single episode. Furthermore, our game presents tasks with varied levels of difficulty.

Additionally, our work distinguishes itself from \citet{carroll2019utility}. Contrary to their settings, our game settings feature a diverse array of tools and task objectives, thereby generating an exponentially larger task space. A comparison between our work and other related games is shown in \autoref{tab:related_work}.

\begin{table*}[t!]
\centering
\setlength\tabcolsep{2.5pt}
\large
\resizebox{\textwidth}{!}{%
\begin{tabular}{ccccccccccc}
\toprule 
Benchmark & Multi-task & \makecell{Object\\Interaction} & \makecell{Tool\\Use} & \makecell{Maximum\\Agents} & \makecell{Collabo-\\ration} & \makecell{Human\\in-the-loop} & \makecell{Procedural\\Level Generation} \\
\midrule
ALFWorld \citep{shridhar2020alfworld} & \cmark & \cmark & \cmark & 1 & \xmark & \xmark & \xmark \\
WAH \citep{puig2020watch} & \cmark & \cmark & \xmark & 2 & \cmark & \cmark & \xmark \\
TextWorld \citep{cote2019textworld} & \cmark & \cmark & \cmark & 1 & \xmark & \xmark & \cmark \\ 
Generative Agents \citep{park2023generative} & \cmark & \cmark & \cmark & 25 & \xmark & \xmark & \cmark \\
EMATP \citep{liu2022embodied} & \cmark & \cmark & \cmark & 2 & \cmark & \xmark & \xmark \\
Overcooked-AI \citep{carroll2019utility} & \xmark & \cmark & \cmark & 2 & \cmark & \cmark & \xmark \\
HandMeThat \citep{wan2022handmethat} & \cmark & \cmark & \cmark & 2 & \cmark & \xmark & \xmark \\
DialFRED \citep{gao2022dialfred} & \cmark & \cmark & \cmark & 2 & \cmark$^{*}$ & \xmark & \xmark \\
TEACH \citep{padmakumar2022teach} & \cmark & \cmark & \cmark & 2 & \cmark$^{*}$ & \xmark & \xmark \\
CerealBar \citep{suhr2019executing} & \xmark & \xmark & \xmark & 2 & \cmark & \xmark & \xmark \\
LIGHT \citep{urbanek2019learning} & \cmark & \xmark & \xmark & 1369 & \xmark & \cmark & \cmark \\
Diplomacy \citep{meta2022human}  & \xmark & \xmark & \xmark & 7 & \cmark & \cmark & \xmark \\ \midrule
\overcook (Ours) & \cmark & \cmark & \cmark & 4+ & \cmark & \cmark & \cmark \\
\bottomrule
\end{tabular}
}
\caption{Comparsion between \overcook and other related benchmarks. \textbf{Multi-task}: The benchmark contains multiple different tasks.  \textbf{Object Interaction}: Agents have to manipulate or engage with different items or environmental elements to achieve certain goals with irreversible actions. \textbf{Tool Use}: Completing tasks necessitates the use of specific tools by the agents. \textbf{Maximum Agents}: This denotes the upper limit of agents that can be present in a single experiment. \textbf{Collaboration}: Many tasks mandate teamwork and collaboration between different agents. \textbf{Human in-the-loop}: The framework allows humans to join the game and collaborate actively with the agents. \textbf{Procedural Level Generation}: There's flexibility in adding new tasks, making the game dynamic and adaptable. $^*$: Notably, even though multiple agents can be present, the second agent is limited to communicating with the first agent. The second agent cannot interact with the environment in an active gaming capacity.}
\label{tab:related_work}

\end{table*}

\begin{table}[t!]
\begin{minipage}{0.532\textwidth}
        \setlength\tabcolsep{4pt}
\resizebox{\textwidth}{!}{
        \begin{tabular}{ccc}
            \toprule
            Type & Arguments & Description \\
            \midrule
            \multirow{2}{*}{\texttt{goto}} & \makecell[tc]{\agent\\\loc} & \makecell[tl]{Move \agent to\\\loc}  \\
            \midrule
            \multirow{3}{*}{\texttt{get}} & \makecell[tc]{\agent\\\loc\\\texttt{(item)}} & \multirow{3}{*}{\makecell[tl]{\agent obtain \texttt{item}\\ from \loc }} \\
            \midrule
            \multirow{2}{*}{\texttt{put}} & \makecell[tc]{\agent\\\loc} & \makecell[tl]{\agent put everything \\it holds to \loc}  \\
            \midrule
            \multirow{2}{*}{\texttt{activate}} & \makecell[tc]{\agent\\\loc} & \makecell[tl]{\agent turn on \\\loc}  \\
            \midrule
            \texttt{noop} & \agent & not dispatching \agent \\
            \bottomrule
        \end{tabular}}
        \captionof{table}{\label{tab:action_space}Action space in \overcook.}
\end{minipage}
    \hfill
\begin{minipage}{0.45\textwidth}
        \centering
        \includegraphics[width=\textwidth]{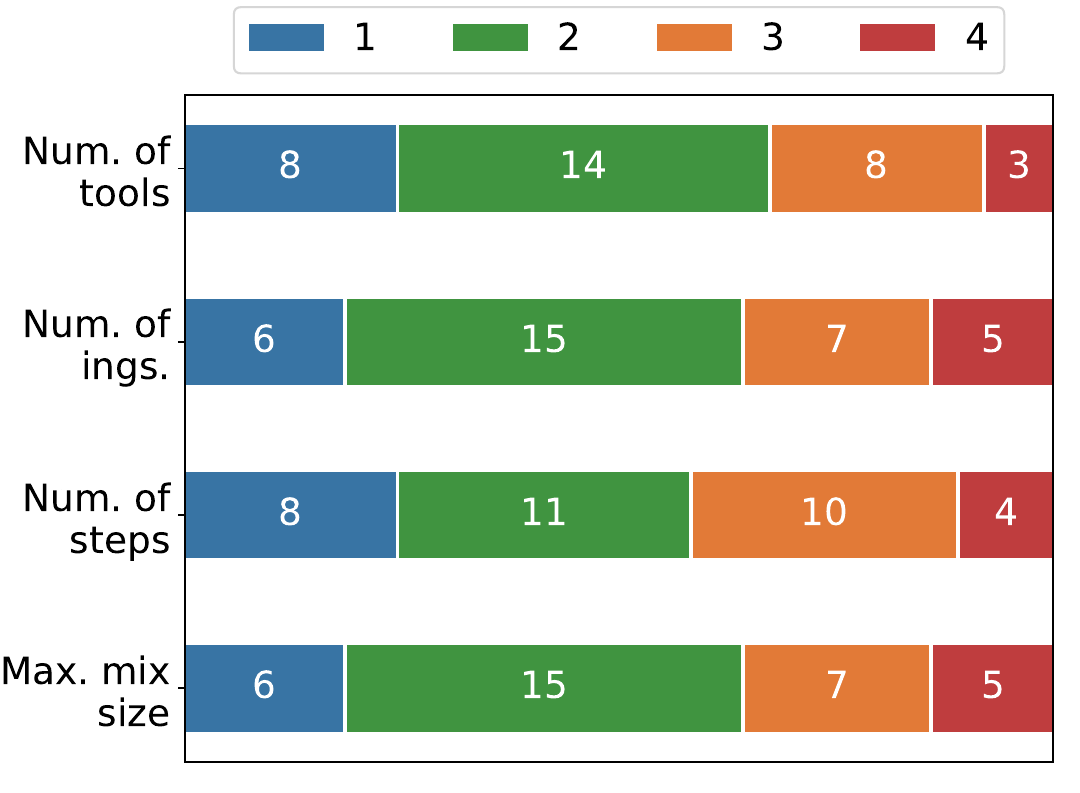}
        \captionof{figure}{\label{fig:sub2} Dish distribution over the number of tools and ingredients (ings.) involved, cooking steps, and maximum mixture size as in the recipe.}
\end{minipage}
\end{table}

\section{The New Gaming \overcook Design and Benchmark}\label{sec:benchmark}

% overview
We introduce \overcook as a novel and flexible game for multi-agent scheduling and coordination in a \textit{virtual kitchen} environment. In this game, a multi-agent system  needs to overlook multiple agents and coordinate them, with the goal of completing as many dish orders as possible. It is equipped with a textual interface since our focus is evaluating LLM-based planning agents. Our modularized design separates tasks and game engines, allowing more tasks (type of dishes) and domains (how to implement the ``kitchen'': text-based engine, Unity, Minecraft, \etc) to be included.

\subsection{Task Definition}

We follow prior works ~\citep{yao2023react,liu2023agentbench,deng2023mind2web} to \textbf{interactively evaluate LLMs as planning agents}. Overall, the interactive evaluation can be formulated as a \textit{Markov Decision Process} $(\mathcal{S}, \mathcal{A}, \mathcal{T}, \mathcal{R}, \mathcal{G})$, with state space $\mathcal{S}$, action space $\mathcal{A}$, (effectively indicating all the possible schedules that can be made at a single time step), transition dynamics $\mathcal{T}$, reward function $\mathcal{R}$ and task instruction space $\mathcal{G}$. Note that, although there are multiple agents inside \overcook that can be coordinated, as we mentioned above, we adopt a centralized planning scheme and thereby formulate our game as a single-agent and fully-observable decision-making problem. An illustration of the state \& action space and the possible tasks of our game can be found in \autoref{fig:game_overview}.
% Nonetheless, beating the game still requires extensive knowledge and skill in scheduling and coordinating multiple agents.

\textbf{State Space $\mathcal{S}$.}~~In \overcook virtual kitchen, there are two types of entity: \loc and \agent. For each entity, the game will provide a set of descriptions, the aggregated descriptions of all entities will be the state returned by our game.
A \loc can be \textit{storage}, where you could obtain ingredients and dispense waste, a \textit{serving table}, where you should put the completed dish on, or a cooking tool, \eg \textit{pan}, \textit{blender}. We offer up to two descriptions for each location: \texttt{inside(location, items)}, indicating what items (some ingredients, completed dishes, \etc) are now inside the location; and \texttt{occupy(location)}, suggesting \loc is now being used and cannot be touched, \eg an activated blender. A \agent is an entity that can be dispatched to complete the task, and we provide up to three descriptions for each agent: \texttt{at(location, agent)}, indicating now \agent is at \loc; \texttt{hold(agent, items)}, suggesting what items \agent is holding; and finally \texttt{occupy(agent)}, implying \agent is now operating a tool, \eg chopping some fruits, and will not respond to any dispatching command. 

\textbf{Action Space $\mathcal{A}$.}~~An action in \overcook is a list of dispatching commands. Given $N$ \agent entities, a total of $N$ commands need to be generated. The agent provides the following commands (also illustrated in \autoref{tab:action_space}): 1) \texttt{goto(agent, location)}, to let \agent move to \loc; 2) \texttt{get(agent, location, item)}, to let \agent get a specific item from \loc; 3) \texttt{put(agent, location)}, to put whatever \agent is holding into \loc; 4) \texttt{activate(agent, location)}, to let \agent turn on \loc if it is a cooking tool, \eg \textit{blender}; 5) \texttt{noop(\agent)}, to have \agent perform no actions in this round of dispatching. We will provide more detailed illustrations and rules about the action space in \supp. Note that, to avoid the possible confusion of multiple agents being dispatched to operate with the same \loc, the dispatcher also needs to properly order the dispatching commands as they will be executed sequentially.

\textbf{Tasks and Reward.}~~A task in \overcook is a dish order, ranging from the most basic \texttt{tunaSashimi}, which can be made by simply chopping some tuna meat, to sophisticated dishes like \texttt{porkPasta} that requires various cooking tools. In a game episode with maximum steps of $T$, every \interval steps (we name this \textit{task interval}), a new task or dish order will be added to the active task list. A task will be viewed as \textit{completed} and removed from the active task list when a matched dish has been put on the serving table. On the contrary, a task will be deemed to have \textit{failed} and removed from the list when it reaches its \textit{lifetime} \lifetime. Lifetime depends on the complexity of the dish and details can be found in \supp. 
Along with the tasks, the game provides rewards \& penalties or feedback on certain occasions, \eg when a task is just completed, some infeasible commands are dispatched, \etc. Due to the space limit, we defer details on tasks to Appendix~B..

\subsection{Implementing \overcook}
The implementation of \overcook mostly follows the spirit of \textit{Overcooked!}, a renowned video game. Therefore we refer to many of its game mechanisms while simplifying some of them, \eg we skip low-level control and assume all \agent have access to all \loc at any time (detailed comparisons between \overcook and the original video game can be found in \supp). Specifically, we crawled the rules and recipes from the community-contributed wiki\footnote{\href{https://steamcommunity.com/sharedfiles/filedetails/?id=1769729191}{https://steamcommunity.com/sharedfiles/filedetails/?id=1769729191}}, streamlined them and made necessary modifications, ending up with the basic version of \overcook comprising \textbf{10} types of \loc (\textit{serving table}, \textit{storage}, and 8 different cooking tools), \textbf{27} types of ingredients, and \textbf{33} unique dishes. We group the dishes based on their difficulty to make (primarily the number of cooking tools involved) and design \textbf{12} game levels, which are further categorized into 4 classes: \textit{entry}, \textit{simple}, \textit{intermediate} and \textit{advanced}, with 3 levels each. Note that the recipes, dishes, and levels can be easily extended to allow more challenging tasks.

\subsection{Evaluation Metric}
\paragraph{Collaboration Score (CoS)} We would like to evaluate to which extent the dispatcher (played by an LLM) can coordinate multiple agents into completing dish orders, across different scenarios. Similar to the original \textit{Overcooked!} game, we are particularly interested in this question: Can the dispatcher still coordinate the agents into efficient collaborations with smaller \interval, \ie more dish orders are flooding in? Our hypothesis is, an ideal dispatcher should be capable of coordinating agents until there are way more tasks than the system can handle. Therefore, we introduce \textit{collaboration score} \colab, defined as below:
\begin{align}
    \text{\textbf{CoS}} = \frac{1}{M}\sum^M_{i=1} \frac{\#\text{\textit{completed task}}\left[\tau_{\text{int},(i)}\right]}{\#\text{\textit{completed task}}\left[\tau_{\text{int},(i)}\right] + \#\text{\textit{failed task}}\left[\tau_{\text{int},(i)}\right]},
\end{align}
where $M$ is the total amount of \interval we evaluate. Effectively, \colab is the average task completion rate across different \interval conditions. In our default setting, we use $M=5$. While the actual values of \interval depend on the game level, we ensure they elicit a wide range of difficulty including both extremely relaxed and intense scenarios.

In a word, CuisineWorld is a game that emulates a virtual kitchen, where several robots are commanded to use various cooking tools and ingredients to prepare as many dish orders as possible in a limited period of time. To facilitate collaboration, new orders will keep flooding in while the existing ones should be completed before expiration. Therefore, LLMs need to properly coordinate these robots to maximize overall productivity. \overcook also offers game levels with a wide range of planning difficulty: dishes with different complexity (number of ingredients and tools involved), number of agents, order frequency and lifetime, etc, making it an ideal test bed for LLM-based multi-agent planning.

% \subsection{Human demonstration dataset}
\section{\mindagent: Infrastructure for Gaming AI}\label{sec:prompt}

\begin{figure}
    \centering    \includegraphics[width=0.99\textwidth]
    {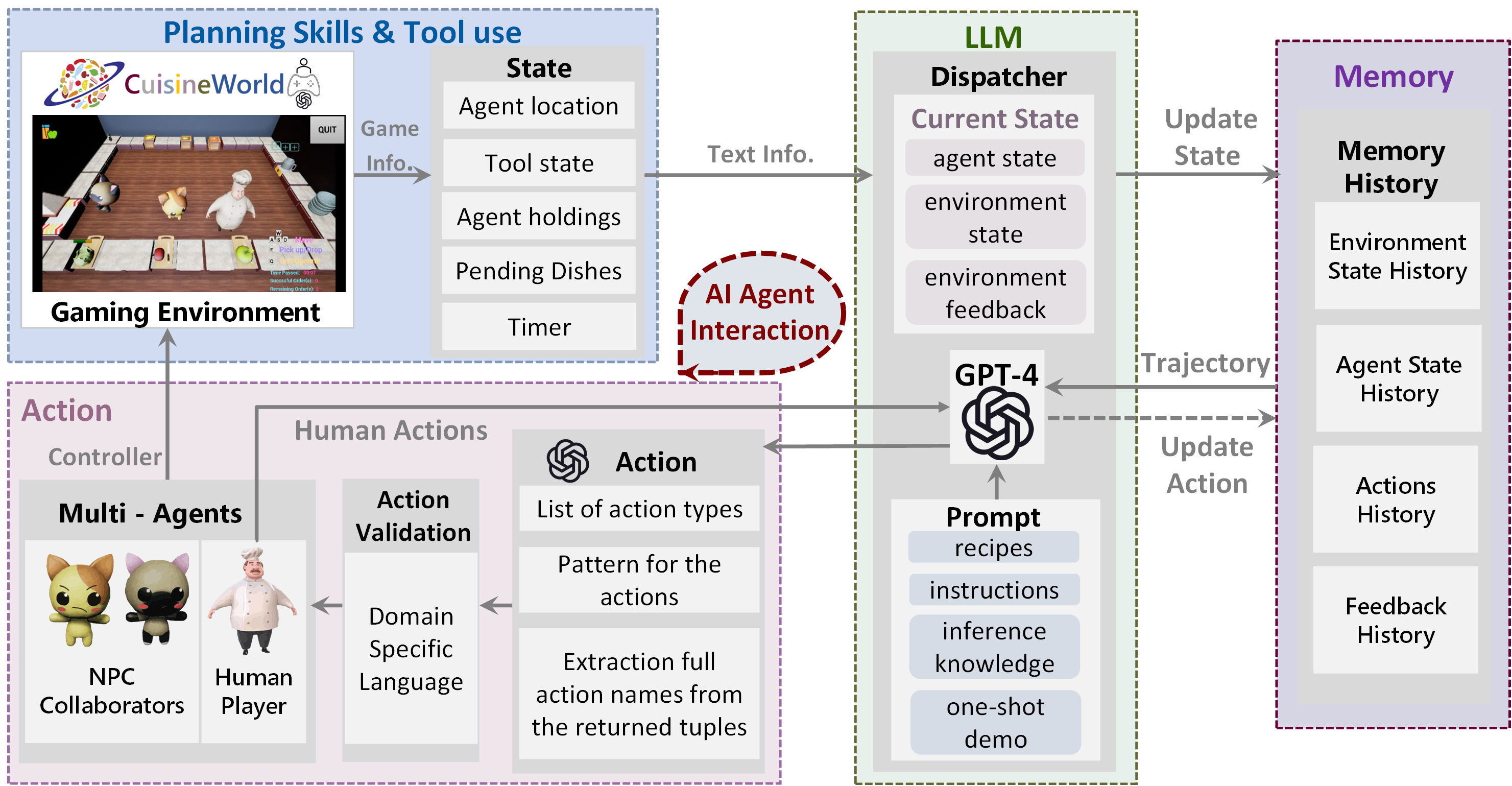}
    \vspace{5mm}
    \caption{Our overview of our \mindagent architecture. \textbf{Planning Skill \& Tool Use}: The game environment requires diverse planning skills and tool use to complete tasks. It emits related game information. This module also converts relevant game data into a structured text format so the LLMs can process it. \textbf{LLM}: The main workhorse of our infrastructure makes decisions, which is a dispatcher for the multi-agent system. \textbf{Memory History}: A storage utility that stores relevant information. \textbf{Action Module}, extract actions from text inputs and convert them into domain-specific language. Validate DSLs so they don't cause errors when executing.  }
    \label{fig:model}
\end{figure}

\subsection{Infrastructure}
Our first foray into the challenging \overcook benchmark is an interactive multi-agent planning framework for LLMs: \mindagent It adopts a minimalist design for the purpose of demonstrating the emergent capacity of LLMs in scheduling and coordination, while also bringing in exploratory prompting techniques that facilitate better planning and shed some light on future approaches. Our infrastructure follows in-context learning. We will outline the key techniques below:

To facilitate in-context learning, our \mindagent infrastructure is composed of three primary components: the prompt, current state, and memory. 

Within the prompt component, there are four distinct sub-components: recipes, general instructions, inference knowledge, and a one-shot demo.

\textbf{Recipes.} outline the hierarchical procedure for preparing various dishes at the given level. They specify the necessary ingredients for each intermediate or final product, the appropriate tools required, and the expected outcome post-cooking.

\textbf{Instructions.} detail the foundational rules of \overcook. These instructions delineate the array of actions agents can undertake within the game and enumerate the characteristics of every tool available in the current kitchen scenario. Moreover, they inform agents about the base ingredients retrievable from storage, as well as all potential intermediate products they can procure. Agents are also explicitly advised to remain cautious about feedback from the environment.

\textbf{Inference Knowledge.} houses insights and helpful hints for the agent. When utilized appropriately, these hints can guide agents to sidestep potential errors and enhance their collaborative efficiency.

\textbf{One-shot Demo.} presents a step-by-step demonstration of the preparation of a distinct dish, different from other dishes at the current level. This demonstration spans several time steps, each of which is incorporated as part of the prompt. The demonstration illustrates the major procedures for cooking one dish in \overcook, including obtaining ingredients, putting ingredients into different tools, transporting intermediate ingredients, and delivering the final dish to the serving table.

\paragraph{Current State} provides a snapshot of the prevailing observations from the environment. It encompasses information such as the agents' locations, the objects currently in the agents' possession, the tools that are accessible within the environment, the ingredients present within each tool, and the tools that are actively in use. Moreover, it includes optional feedback from the environment, triggered when the agents' actions contravene the environment rules— for instance, when assigning two distinct actions to the same agent.  

\textbf{Memory History.} archives the interaction history with the environment. Specifically, it chronicles the state of the environment and the state of the agents at every time step.

In addition to the prompt modules, additional modules are implemented to help interface between LLMs and \overcook. 

\textbf{Action Extraction.} employs a regular expression matching procedure to distill agent actions from the LLM's textual output. This module is indispensable because, on occasion, the LLM's output is not clean. The output contains information reflecting its internal thought processes. At times, the LLM might even issue apologies for prior missteps in reaction to environment feedback.

\textbf{Action Validation.} utilizes a look-ahead checking mechanism. This module parses the proposed actions, assessing their feasibility. Should an action be deemed inexecutable, an error message is promptly returned.
%\textbf{Reading the rules and examples.}~~

%\textbf{Feedback and re-planning.}~~
%Feedback vs. reflexion -- where do retrospective come from?

\subsection{Infrastructure Mechanism}

Assuming a multi-agent system with a total of $N$ agents, the system must complete a sequence of $P$ different tasks. Each task has $M_p$ different sub-tasks. Furthermore, the number and types of tasks are unknown at the beginning of the episode. The environment will sample a task for the agents to finish for a given interval. Then the agents need to complete the designated task along with other tasks in the task queue. In addition, each task has an expiration time. After the expiration time, the task will be marked as a failure. The objective of the multi-agent system is to finish as many tasks as possible and fail as fewer tasks as possible within a given time frame. 

We aim to find valid and optimal task planning, scheduling, and allocations. We define $q_{pim}$ and $c_{pim}$ as quality and cost, respectively, for allocating agent $i$ to work on the sub-task $m$ for the $p$ th task in the episode. Then the combined utility for the sub-task is:
$$
 u_{pim}= \begin{cases}q_{pim}-c_{pim}, & \text{ if agent } i \text{ can execute sub-task m for the } p \text{ th task in the episode} \\ -\infty. & \text { otherwise }\end{cases}
 $$
 We define the assignment of sub-task $m$ to agent $i$ as 
 $$
 v_{pim}= \begin{cases}1, & \text{agent } i \text{ is assigned to sub-task m for the } p \text{ th task in the episode} \\ 0. & \text { otherwise }\end{cases}
$$
% with $\gamma_m = i$ being an assignment variable for each sub-task $m$, indicating that sub-task $m$ is assigned to agent $i$. 

The goal is to maximize the utility of the episode under a time constraint. Define the execution time for task $m$ by agent $i$ for the $p$ th task in the episode as $\tau_{p i m}$, and the maximum time allowed to execute the task as $T_{max}$, we can express the task decomposition and assignment problem as follows:

\begin{equation}
\label{equ:obj}
\argmax_{v} \sum_{p=1}^{P} \sum_{i=1}^{N} \sum_{m=1}^{M_p} u_{p i m} v_{p i m}
\end{equation}

Subject to:
$$
\begin{array}{rlrl}
\sum_p \sum_i\sum_m  \tau_{p i m} v_{p i m} & \leq T_{max} \\
\sum_{i} v_{p i m} & \leq 1 & \forall m \in M, \forall p \in P \\
v_{p i m} & \in\{0,1\} & \forall i \in N, \forall m \in M, \forall p \in P
\end{array}
$$

As pointed out by \citep{korsah2013comprehensive}, this problem cannot be solved in polynomial time. In this work, we tackle this problem by using large-language models. 

Our prompt design choices try to help LLM system solve \autoref{equ:obj}. In practice, we reformulate \autoref{equ:obj} with qualities or rewards expressed in natural languages as environment feedback. For example, when the agent successfully collects an item, the environment emits a signal ``collect finish." When the dispatcher assigns a different task to the same agent, the environment will emit a signal ``agent ids cannot be the same."   As rewards are not immediately observable, we borrow sprites from temporal difference learning. We accumulate state-action history into memory history. Due to context length limits, it's infeasible to fit the entire history into the context window. We select a fixed horizon history as a part of the prompt to guide the model performance. We further express the constraints of the system in natural language formats and repeat important constraints multiple times if necessary.

\section{Experiments and Results}\label{sec:exp}

\paragraph{Overview}
We conduct extensive experiments in \overcook. We first introduce the experiment settings and present an analysis of empirical results in \overcook. Our experiments focus on addressing the following research questions: 

\textbf{Q1: } How efficiently can the model dispatch multiple agents?

\textbf{Q2: } Can the model dispatch agents for dynamic, on-the-fly goals across different tasks?

\textbf{Q3: } How do various components of the input prompt influence the model’s performance?

\textbf{Q4: } How do other LLMs perform compared to GPT-4?

\textbf{Q5: } To what extent can the existing methods collaborate with human users? 

\textbf{Q6: } What's the human perception of collaborating with numerous intelligent agents?

\subsection{LLM Settings}
We perform experiments on \overcook through OpenAI APIs and anthropic APIs. All GPT-4 experiments are using gpt-4-0613 model, and all chat-GPT experiments are using gpt-3.5-turbo-0613. For Llama 2 experiments, we use hugging face inference endpoints Llama-2-70b-chat-hf. We set the temperature for all experiments to 0.1 following \citep{wang2023voyager}. We report the average results over three episodes. 

\subsection{EXPERIMENT SETTING I: LLMs Dispatch Multi-Agents (NPC)}

\begin{figure}
    \centering
    \includegraphics[width=0.99\textwidth]{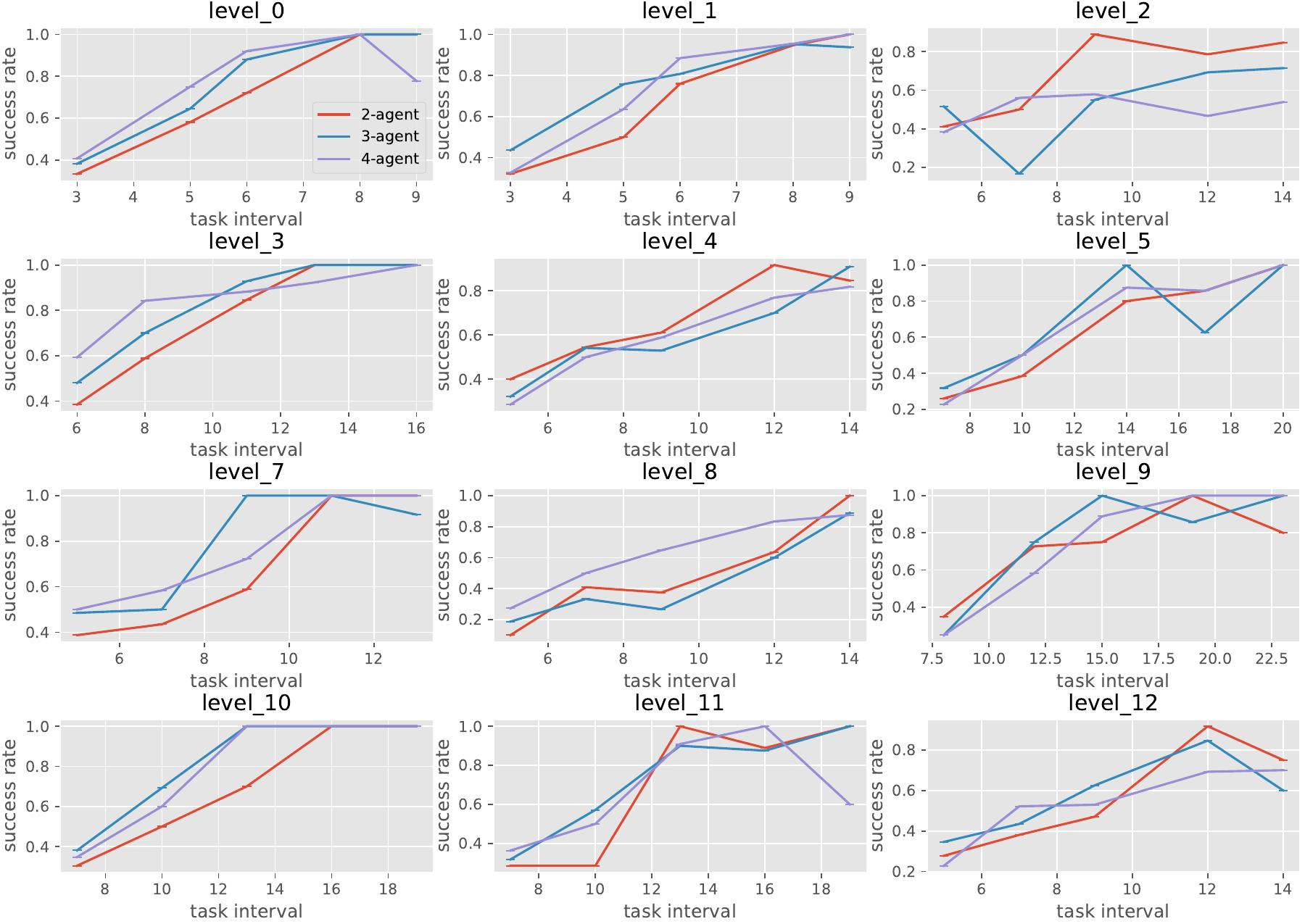}
    \caption{Collaboration Results on Different Tasks}
    \label{fig:main_result}
\end{figure}

\paragraph{Collaboration Efficiency (Q1, Q2)} \autoref{fig:main_result} and \autoref{tab:main_result_2agents}, \autoref{tab:main_result_3agents} and \autoref{tab:main_result_4agents} reports the system performance under different settings. In particular, \autoref{tab:main_result_2agents} reports the multi-agent collaboration results among two agents. \autoref{tab:main_result_3agents} reports the multi-agent collaboration results among three agents, and \autoref{tab:main_result_4agents} reports the multi-agent collaboration results among four agents. \autoref{fig:main_result} displays the collaboration efficiency curve.

As shown in \autoref{fig:main_result}, across different task levels, more agents generally lead to better collaboration efficiencies. As the collaboration efficiency curve is generally higher with more agents. 

Computing \colab by levels also reveals that more agents will lead to better collaboration efficiencies. As shown in the tables, the \colab score is the highest when there are two agents in two cases. The \colab score is the highest when there are three agents in seven cases.  The \colab score is the highest when there are four agents in three cases. The results also confirm that more agents will lead to higher collaboration efficiencies.

\paragraph{Findings} First, we observe that the system performance is generally better when there are more agents, indicating that LLM dispatcher can coordinate more agents to execute tasks more efficiently. Second, we observe that the system performance degrades with more agents in less demanding conditions, indicating that LLM  dispatcher struggles when there are fewer tasks.

\subsection{EXPERIMENT SETTING II: Human and Multi-NPCs with LLMs}

\subsubsection{Human Data Collection} 

\paragraph{Human Testing of Study Protocol} Before starting the experiment, a webpage introduction to the game is handed to the players. It contains rules and the basic controls of the game. Then we randomly assign the playing order. Participants can drop out of the testing at any time as they wise; in that case, their data will be discarded. The human evaluation interface is shown in Appendix~D.

\paragraph{Measurement} In the background, we collect the number of failed and successful tasks during the participant's interaction with the game system. In addition, we record the entire action history of players and intelligent agents. Therefore, we can replay action histories for further analysis. After each episode, the participants must complete a survey about their engagement with the system on a 5-point likert chart.

Our objective measure is intended to evaluate the human AI teaming performance, and the subjective measure is designed to evaluate users’ perceptions of the system.

\begin{table*}[h!]
\resizebox{\textwidth}{!}{

\begin{tabular}{c|ccc|ccc|ccc|ccc|c}
\toprule
\multirow{2}{*}{2-agent} & \multicolumn{3}{c|}{ very simple } & \multicolumn{3}{c|}{ simple } & \multicolumn{3}{c|}{ intermediate } & \multicolumn{3}{c|}{ advanced } & \multirow{2}{*}{Avg.} \\ \cmidrule{2-13}
% \midrule
& level 0 & level 1 & level 7 & level 2 & level 4 & level 8 & level 3 & level 9 & level 10 & level 5 & level 11 & level 12 & \\
\midrule GPT4 $\tau_{\text{int},(1)}$ & $18 / 54$ & $18 / 56$ & $12 / 31$ & $14 / 34$ & $12 / 30$ & $3 / 30$ & $10 / 26$ & $7 / 20$ & $7 / 23$ & $6 / 23$ & $6 / 21$ & $10 / 36$ & 0.318\\
GPT4 $\tau_{\text{int},(2)}$ & $18 / 31$ & $17 / 34$ & $10 / 23$ & 13/26 & $12 / 22$ & $9 / 22$ & 10/17 & $8 / 11$ & $6 / 12$ & $5 / 13$ & $4 / 14$ & $8 / 21$ & 0.486 \\
 GPT4 $\tau_{\text{int},(3)}$ & $18 / 25$ & $19 / 25$ & $10 / 17$ & 16/18 & 11/18 & $6 / 16$ & $11 / 13$ & $6 / 8$ & $7 / 10$ & $8 / 10$ & $9 / 9$ & $8 / 17$ & 0.709 \\
 GPT4 $\tau_{\text{int},(4)}$ & 18/18 & $18 / 19$ & $12 / 12$ & 11/14 & 11/12 & 7/11 & $12 / 12$ & $8 / 8$ & 9/9 & $6 / 7$ & $8 / 9$ & $11 / 12$ & \textbf{0.912} \\
GPT4 $\tau_{\text{int},(5)}$ & 18/18 & $17 / 17$ & $12 / 12$ & 11/13 & $11 / 13$ & 9/9 & $11 / 11$ & $4 / 5$ & $7 / 7$ & $8 / 8$ & $8 / 8$ & $9 / 12$ & \textbf{0.937}\\
 \colab & 0.727 & 0.706 & 0.682 & \textbf{0.687} & \textbf{0.664} & 0.504 & 0.764 & 0.725 & 0.701 & 0.661 & 0.692 &  0.559  & 0.673\\
\bottomrule
\end{tabular}
}

\caption{2 agents performance on different tasks}
\label{tab:main_result_2agents}

\end{table*}

% \begin{table*}[h!]
%     \centering
%    \resizebox{\textwidth}{!}{\begin{tabular}{|c|c|c|c|c|c|c|c|c|c|c|c|c|c|}
% \hline \multirow[t]{2}{*}{3 agent } & \multicolumn{3}{|c|}{ very simple } & \multicolumn{3}{|c|}{ simple } & \multicolumn{3}{|c|}{ intermediate } & \multicolumn{3}{|c|}{ advanced } & \\
% \hline & level 0 & level 1 & level 7 & level 2 & level 4 & level 8 & level 3 & level 9 & level 10 & level 5 & level 11 & level 12 & Average \\
% \hline GPT4 $\tau_{\text{int},(1)}$ & $21 / 55$ & $24 / 55$ & $16 / 33$ & $17 / 33$ & $9 / 28$ & $6 / 32$ & $12 / 25$ & $5 / 20$ & $8 / 21$ & 7/22 & 7/22 & 9/26 & \textbf{0.368}\\
% \hline GPT4 $\tau_{\text{int},(2)}$ & 20/31 & $25 / 33$ & $11 / 22$ & $4 / 24$ & $13 / 24$ & $7 / 21$ & $14 / 20$ & 9/12 & 9/13 & 7/14 & 8/14 & 10/23 & 0.549 \\
% \hline GPT4 $\tau_{\text{int},(3)}$ & $22 / 25$ & $21 / 26$ & 17/17 & $11 / 20$ & 9/17 & $4 / 15$ & 13/14 & 8/8 & 12/12 & 7/7 & 9/10 & 10/16 & \textbf{0.791}\\
% \hline GPT4 $\tau_{\text{int},(4)}$ & $22 / 22$ & 20/21 & $14 / 14$ & $9 / 13$ & 7/10 & $6 / 10$ & 10/10 & $6 / 7$ & 10/10 & $5 / 8$ & $7 / 8$ & 11/13 & 0.846\\
% \hline GPT4 $\tau_{\text{int},(5)}$ & $20 / 20$ & $15 / 16$ & 11/12 & 10/14 & 10/11 & $8 / 9$ & 12/12 & $6 /6 $ & $8 / 8$ & $5 / 5$ & 8/8 & $6 / 10$ & 0.914\\
% \hline \colab & \textbf{0.781} & \textbf{0.778} & \textbf{0.780} & 0.528 & 0.600 & 0.455 & 0.822 & \textbf{0.771} & \textbf{0.815} & 0.689 & \textbf{0.733} & \textbf{0.570} & \textbf{0.694}\\
% \hline
% \end{tabular}
% }
%     \caption{3 agents performance on different tasks}
%     \label{tab:main_result_3agents}
% \end{table*}

\begin{table*}[h!]
\centering
\resizebox{\textwidth}{!}{
\begin{tabular}{c|ccc|ccc|ccc|ccc|c}
\toprule
\multirow{2}{*}{3-agent} & \multicolumn{3}{c|}{ very simple } & \multicolumn{3}{c|}{ simple } & \multicolumn{3}{c|}{ intermediate } & \multicolumn{3}{c|}{ advanced } & \multirow{2}{*}{Average} \\
\cmidrule{2-13}
& level 0 & level 1 & level 7 & level 2 & level 4 & level 8 & level 3 & level 9 & level 10 & level 5 & level 11 & level 12 & \\
\midrule
GPT4 $\tau_{\text{int},(1)}$ & $21 / 55$ & $24 / 55$ & $16 / 33$ & $17 / 33$ & $9 / 28$ & $6 / 32$ & $12 / 25$ & $5 / 20$ & $8 / 21$ & 7/22 & 7/22 & 9/26 & \textbf{0.368} \\
GPT4 $\tau_{\text{int},(2)}$ & 20/31 & $25 / 33$ & $11 / 22$ & $4 / 24$ & $13 / 24$ & $7 / 21$ & $14 / 20$ & 9/12 & 9/13 & 7/14 & 8/14 & 10/23 & 0.549 \\
GPT4 $\tau_{\text{int},(3)}$ & $22 / 25$ & $21 / 26$ & 17/17 & $11 / 20$ & 9/17 & $4 / 15$ & 13/14 & 8/8 & 12/12 & 7/7 & 9/10 & 10/16 & \textbf{0.791} \\
GPT4 $\tau_{\text{int},(4)}$ & $22 / 22$ & 20/21 & $14 / 14$ & $9 / 13$ & 7/10 & $6 / 10$ & 10/10 & $6 / 7$ & 10/10 & $5 / 8$ & $7 / 8$ & 11/13 & 0.846 \\
GPT4 $\tau_{\text{int},(5)}$ & $20 / 20$ & $15 / 16$ & 11/12 & 10/14 & 10/11 & $8 / 9$ & 12/12 & $6 /6 $ & $8 / 8$ & $5 / 5$ & 8/8 & $6 / 10$ & 0.914 \\
CoS & \textbf{0.781} & \textbf{0.778} & \textbf{0.780} & 0.528 & 0.600 & 0.455 & 0.822 & \textbf{0.771} & \textbf{0.815} & 0.689 & \textbf{0.733} & \textbf{0.570} & \textbf{0.694} \\
\bottomrule
\end{tabular}
}
\caption{3 agents performance on different tasks}
\label{tab:main_result_3agents}
\end{table*}

\begin{table*}[h!]
\centering
\resizebox{\textwidth}{!}{
\begin{tabular}{c|ccc|ccc|ccc|ccc|c}
\toprule
\multirow{2}{*}{4-agent} & \multicolumn{3}{c|}{very simple} & \multicolumn{3}{c|}{simple} & \multicolumn{3}{c|}{intermediate} & \multicolumn{3}{c|}{advanced} & \multirow{2}{*}{Average} \\
\cmidrule{2-13}
& level 0 & level 1 & level 7 & level 2 & level 4 & level 8 & level 3 & level 9 & level 10 & level 5 & level 11 & level 12 \\
\midrule
GPT4 $\tau_{\text{int},(1)}$ & $22 / 54$ & $18 / 55$ & $17 / 34$ & $13 / 34$ & $8 / 28$ & $9 / 33$ & $16 / 27$ & $5 / 20$ & $8 / 23$ & $5 / 22$ & $8 / 22$ & $8 / 35$ & 0.349 \\
GPT4 $\tau_{\text{int},(2)}$ & $24 / 32$ & $21 / 33$ & $14 / 24$ & $14 / 25$ & $12 / 24$ & $11 / 22$ & $16 / 19$ & 7/12 & 9/15 & $7 / 14$ & $6 / 12$ & $12 / 23$ & \textbf{0.590} \\
GPT4 $\tau_{\text{int},(3)}$ & $23 / 25$ & $23 / 26$ & $13 / 18$ & $11 / 19$ & 10/17 & $11 / 17$ & $15 / 17$ & $8 / 9$ & $11 / 11$ & $7 / 8$ & 10/11 & 9/17 & 0.785 \\
GPT4 $\tau_{\text{int},(4)}$ & $22 / 22$ & $21 / 22$ & $14 / 14$ & $7 / 15$ & $10 / 13$ & 10/12 & $12 / 13$ & 9/9 & $10 / 10$ & $6 / 7$ & $8 / 8$ & 9/13 & 0.875 \\
GPT4 $\tau_{\text{int},(5)}$ & 14/18 & $20 / 20$ & 14/14 & $7 / 13$ & 9/11 & $7 / 8$ & $12 / 12$ & $5 / 5$ & $7 / 7$ & $6 / 6$ & $3 / 5$ & $7 / 10$ & 0.859 \\
CoS & 0.771 & 0.761 & 0.761 & 0.505 & 0.592 & \textbf{0.626} & \textbf{0.848} & 0.744 & 0.790 & \textbf{0.692} & 0.675 & 0.534 & 0.692 \\
\bottomrule
\end{tabular}
}
\caption{4 agents performance on different tasks}
\label{tab:main_result_4agents}
\end{table*}

\subsubsection{Experiment II Setting} 
We conducted a  user study in our gaming environment that tries to answer \textbf{Q5, Q6}. 

The user study evaluates the LLM dispatcher's capabilities of collaborating with humans, where participants are collaborating with 1,2,3 agents or working alone on the virtual cooking tasks. We consider the most general setting, where the LLM works on the unseen task, level\_3. 

\subsubsection{Experiment II Design}

\textbf{Hypotheses}. The user study tests the following hypotheses:
\begin{itemize}
    \item \textbf{H1: Task productivity}. Participants have higher productivity if collaborating with AI agents.
    \item \textbf{H2: Task productivity with more agents}. Participants have higher productivity if collaborating with more AI agents.
    \item \textbf{H3: Perception of the robot.} Participants would have higher perceived task efficiency and have more fun playing the game due to collaboration. 
\end{itemize}

\textbf{Manipulated Variables}. We use a within-subject design for our experiment. In particular, every user tries to finish the task by himself or collaborates with different numbers of robots with varying degrees of competency. We randomize the order of the treatment to mitigate practice effects, fatigue effects, and carryover effects.
\begin{itemize}
    \item \textbf{Single agent: } Participants work on the task by themselves. 
    \item \textbf{LLM powered multi-agent system: } Participants collaborate with the multi-agent system powered by LLM. 
    \item \textbf{Random agent: } Random agents execute random actions from a pool of valid actions. Participants collaborate with random agents.
\end{itemize}

\begin{figure}[h!]
    \centering
    
        \subcaptionbox{ 
        \textbf{Collaboration score} 
        We can tell that the collaboration score is higher if more agents are collaborating with human players, even though the difference is not significant.
        }{\includegraphics[width=0.31\linewidth]{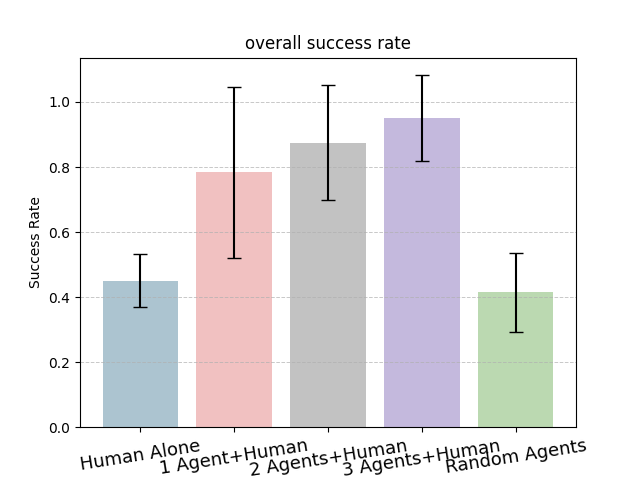}}
    ~ %add desired spacing between images, e.g. ~, \quad, \qquad, \hfill etc.
        \subcaptionbox{
        \textbf{Perceived Enjoyment} 
        Humans enjoy the game more if they collaborate with the right number of agents
        }
        { 
        \includegraphics[width=0.31\linewidth]{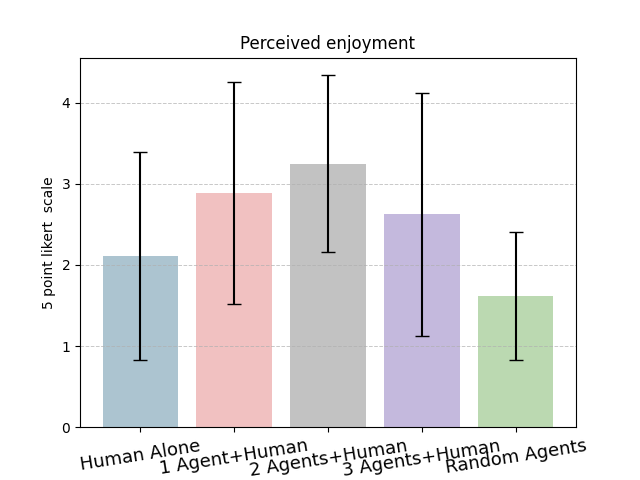} }
    ~ %add desired spacing between images, e.g. ~, \quad, \qquad, \hfill etc.
        \subcaptionbox{\textbf{Perceived more fun} 
        due to collaboration. Players enjoy the game more because of collaborating with competent agents.
        }{
        \includegraphics[width=0.31\linewidth]{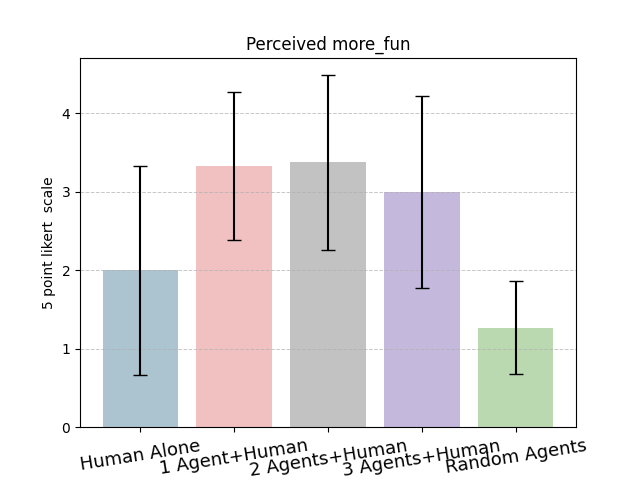}
        }
        \subcaptionbox{
        \textbf{Perceived Assisting}. There is no significant difference in terms of human perceptions of helpfulness when collaborating with more agents, even though the task success rate is higher.
        }{ \includegraphics[width=0.31\linewidth]{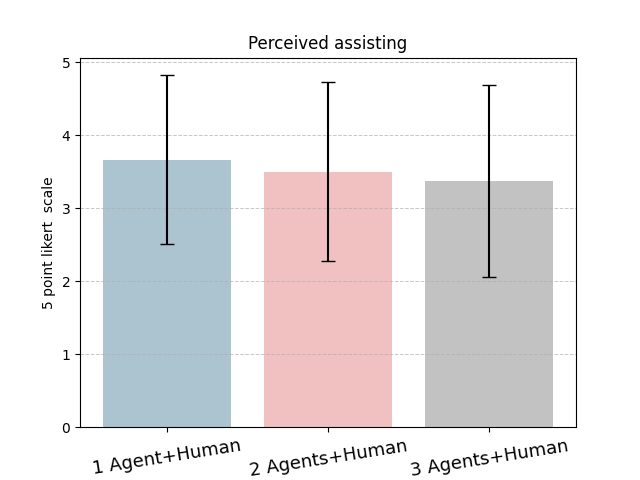}}
    ~ %add desired spacing between images, e.g. ~, \quad, \qquad, \hfill etc.
        \subcaptionbox{ 
        \textbf{Perceived dependability}. When collaborating with more agents,  players depend on the agents more.
        }{\includegraphics[width=0.31\linewidth]{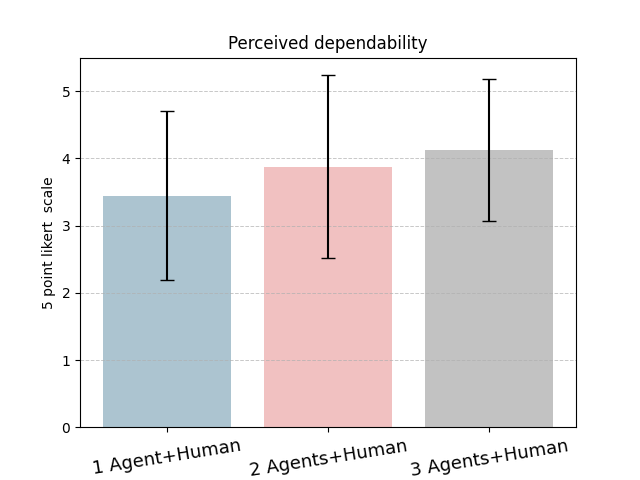}}
    ~ %add desired spacing between images, e.g. ~, \quad, \qquad, \hfill etc.
        \subcaptionbox{\textbf{Perceived Predictability}. There is no difference in terms of the predictability of agents' behaviors when collaborating with more agents. }{ \includegraphics[width=0.32\linewidth]{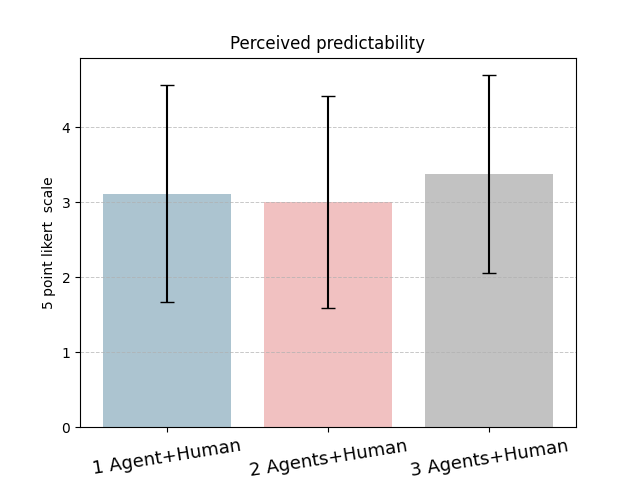}}
        \subcaptionbox{ \textbf{Perceived productivity}. Players think collaborating with AI agents will improve productivity. }{ \includegraphics[width=0.31\linewidth]{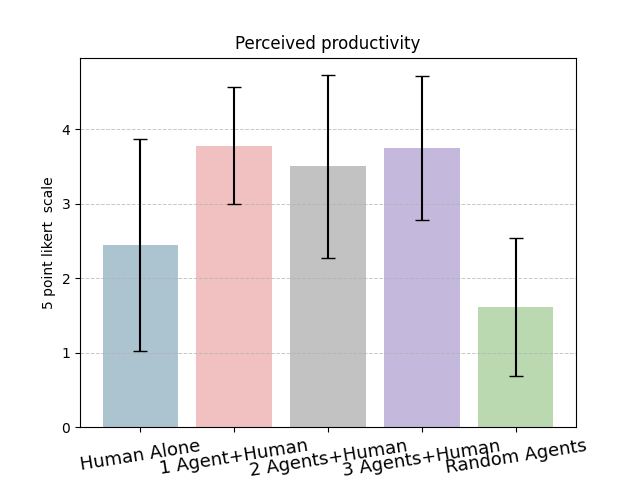}
        }
        \subcaptionbox{ \textbf{Perceived Trust}. There is no difference in terms of trust when collaborating with more agents. }{
         \includegraphics[width=0.31\linewidth]{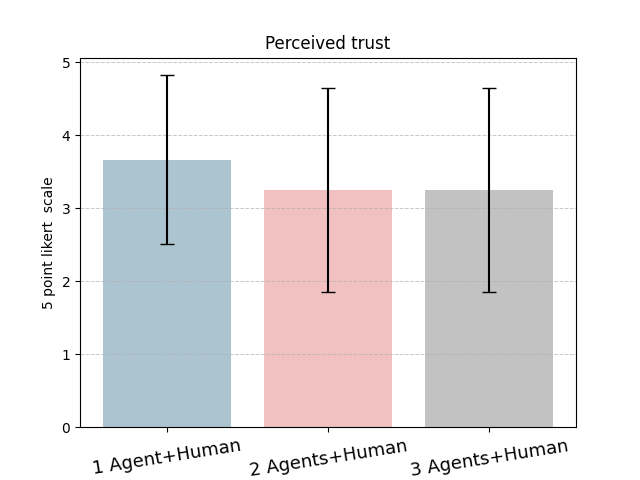}
        }
    
    \caption{Human Evaluations}
    \label{fig:three_images}
\end{figure}
\paragraph{Main Results}
We recruited 12 subjects for our study. Among them, there are two females and 10 males. 

We use ANOVA to test the effects of different experimental conditions on collaboration performance and subjective perception of the AI agents.  Tukey HSD tests are conducted on all possible pairs of experimental conditions.

\paragraph{Findings} We find significant effects on team collaboration success rate $F(4, 55) = 28.11, p < 0.001 $.  Post-hoc comparisons using the Tukey HSD tests revealed that  the team of the player with LLM agents achieves a higher success rate than a human working alone, $p < 0.001 $ across different numbers of agents, \textbf{confirming H1}. Even though the success rate is generally higher when collaborating with more agents, there is no significant effect compared with collaborating with one agent, collaborating with two agents $p=0.774$, or collaborating with three agents $p=0.231$.  We observe that human players have more fun playing the game when collaborating with LLM-powered intelligent agents than playing alone, $p=0.0126$. Players feel that collaboration with intelligent agents leads to higher productivity, $p=0.0104$, thus \textbf{confirming H3}.

In addition, when playing with intelligent agents, human players will take their actions based on other players' actions $p=0.00266$. Human players also found that intelligent agents are more predictable compared with random agents $p < 0.001$. 

Further insights from player feedback highlighted an intriguing trade-off: while more agents improved overall task success rates, it reduced the game's enjoyment. Often, players felt sidelined and less involved. Thus, game developers should adjust AI performance to maintain player engagement and fun. As indicated by \citet{yuan2022situ}, aligning human values with AIs might be a promising way to solve this problem.

\subsection{Visualing "CuisineWorld"}
\begin{figure}[h!]
    \centering
    \setlength{\tabcolsep}{2pt}  % Adjust this to control the space between columns
    \begin{tabular}{m{1cm}@{\hspace{-10pt}} m{6cm} m{6cm}}
        \rotatebox[origin=l]{90}{\textbf{Multi-agent}} & \includegraphics[width=6cm]{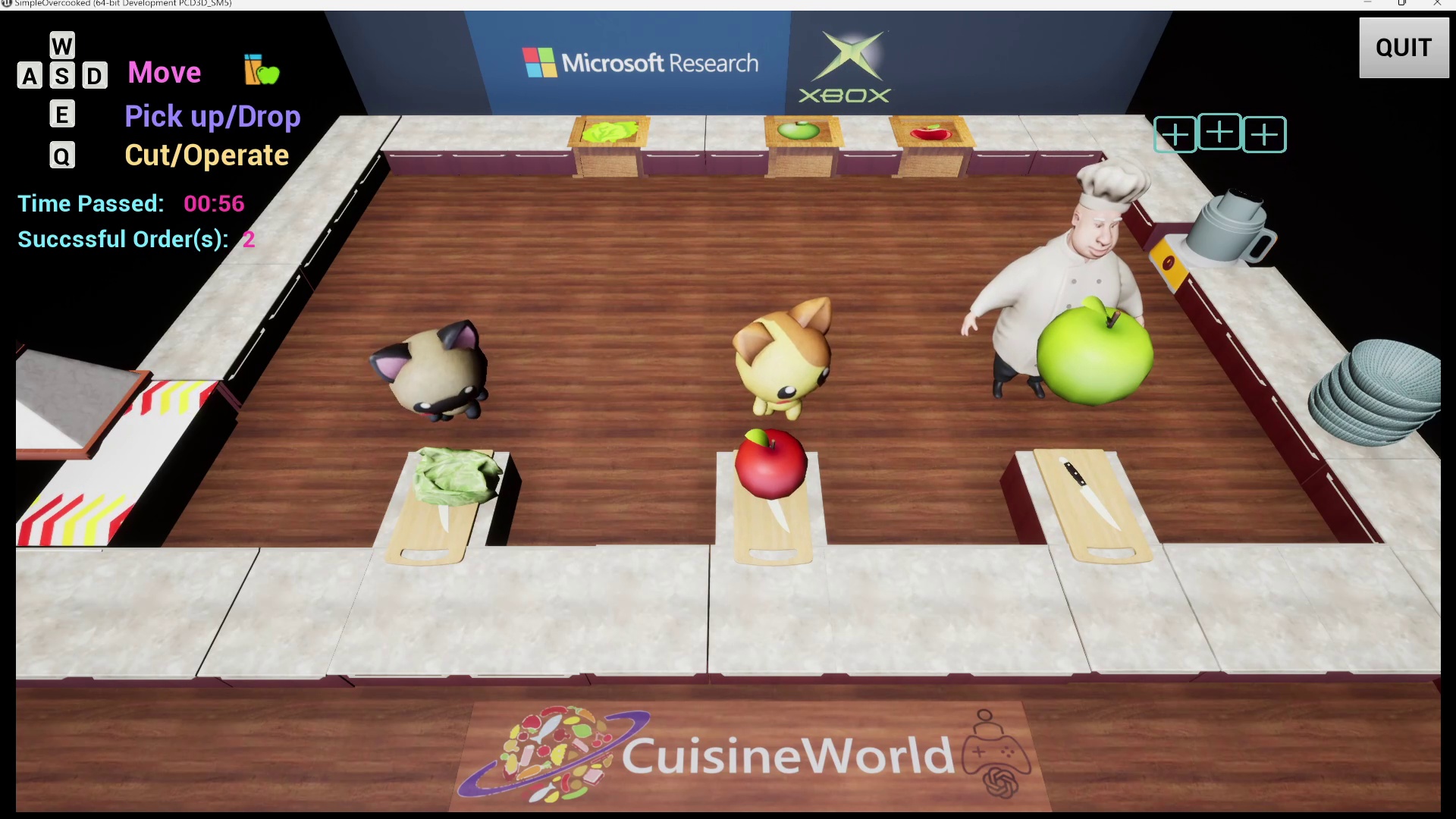} & \includegraphics[width=6cm]{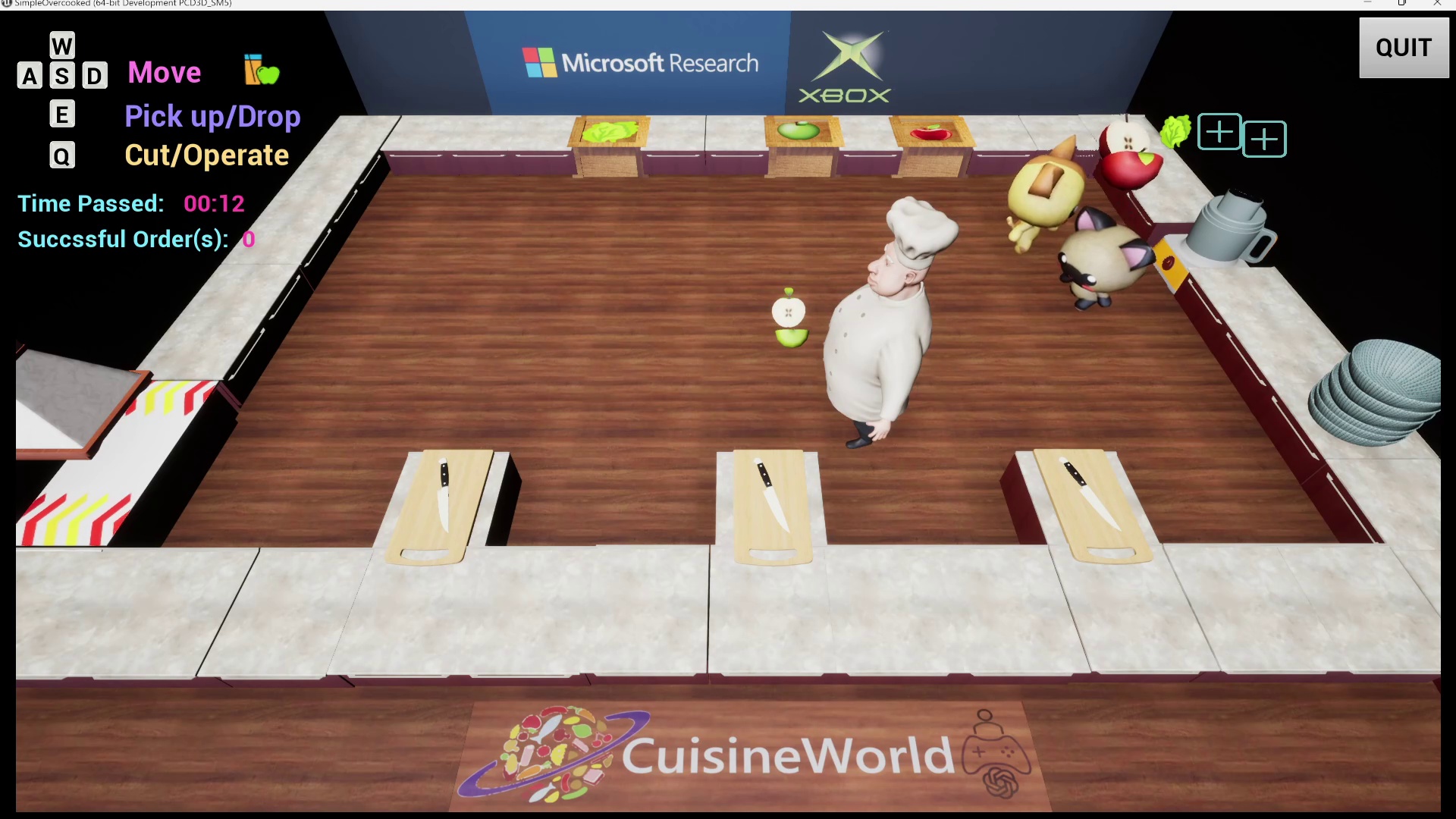} \\ 
        \rotatebox[origin=l]{90}{\textbf{Human-agent}} & \includegraphics[width=6cm]{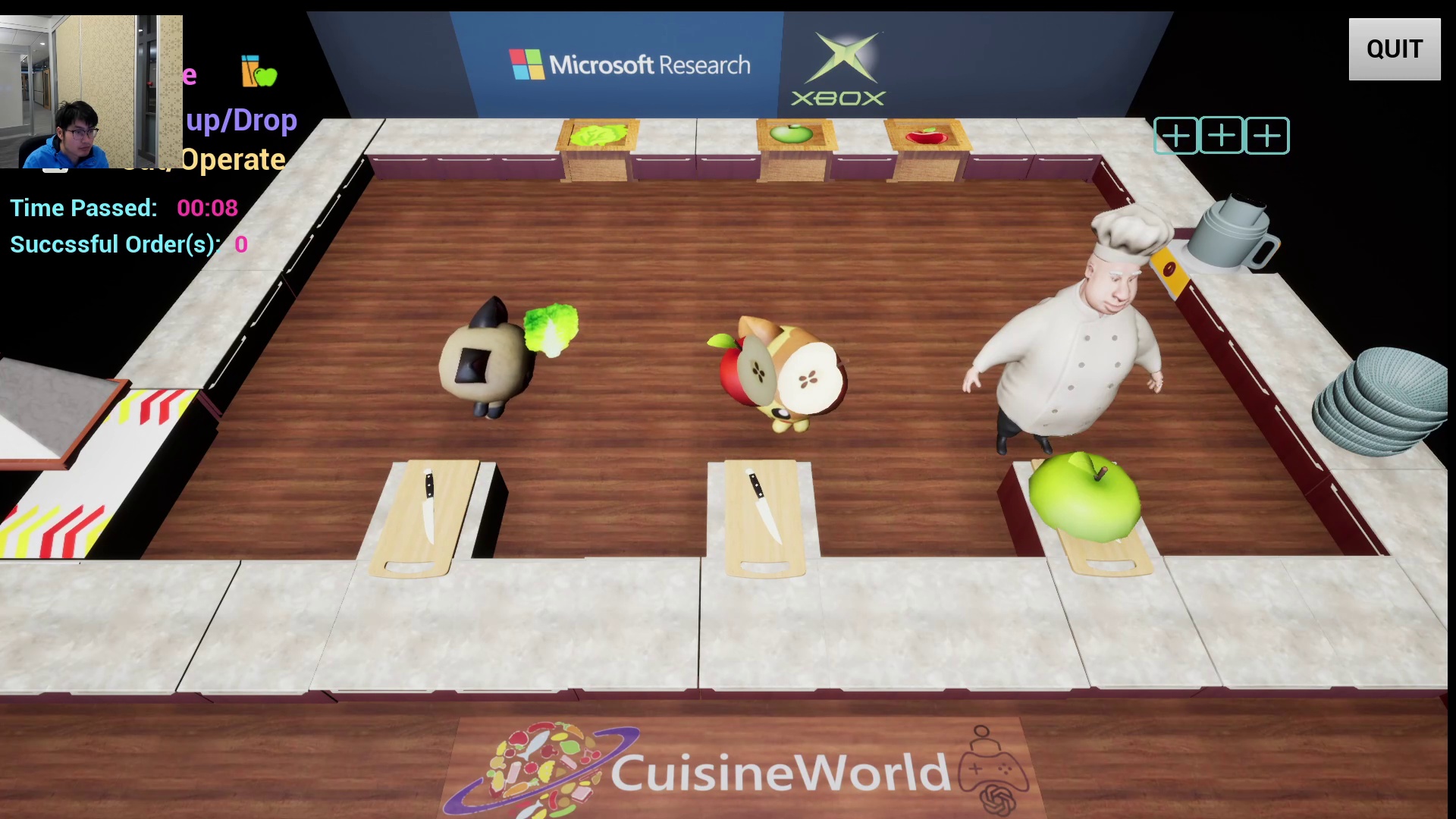} & \includegraphics[width=6cm]{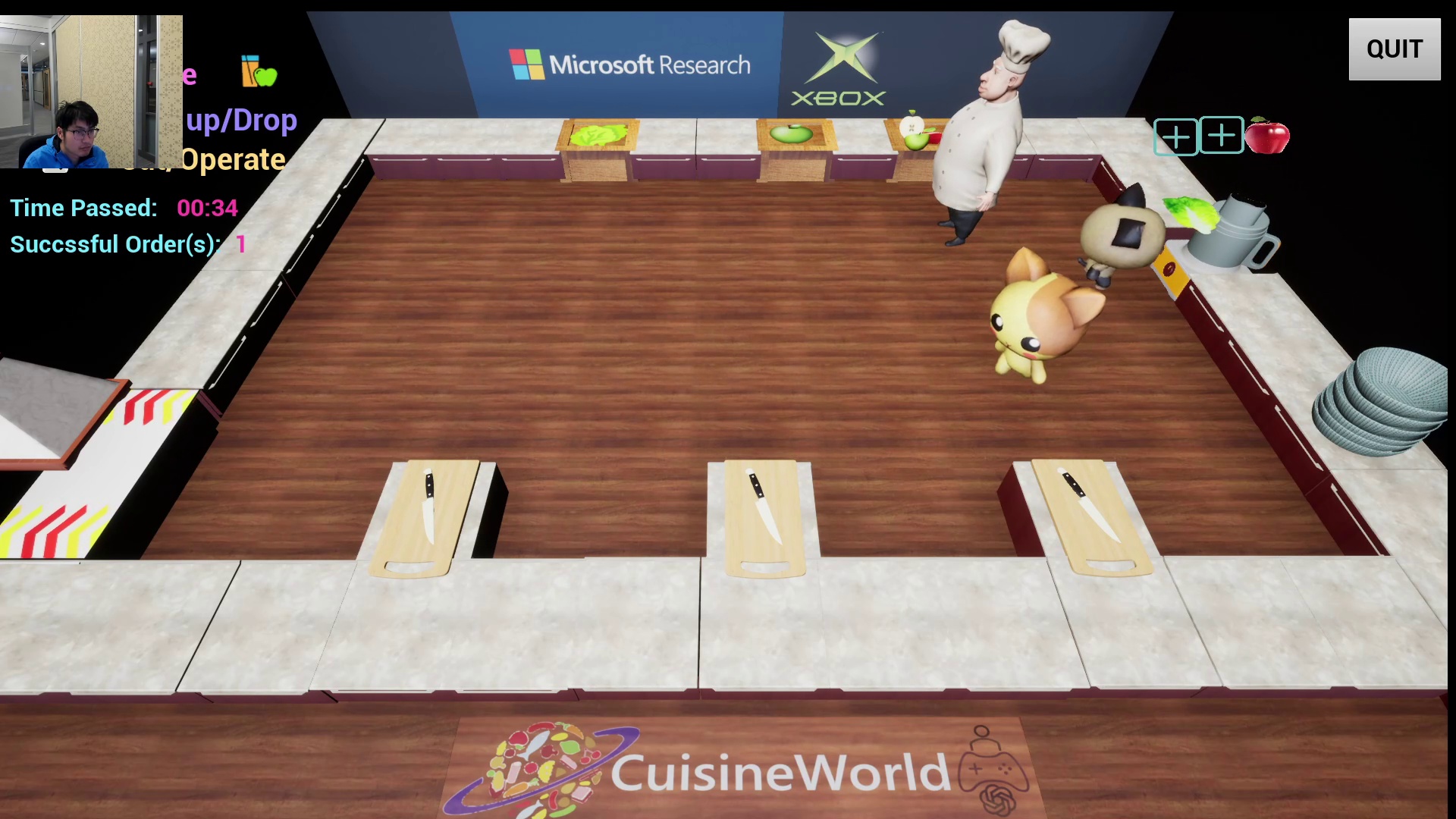} \\ 
        \rotatebox[origin=l]{90}{\textbf{VR Interaction}} & \includegraphics[width=6cm]{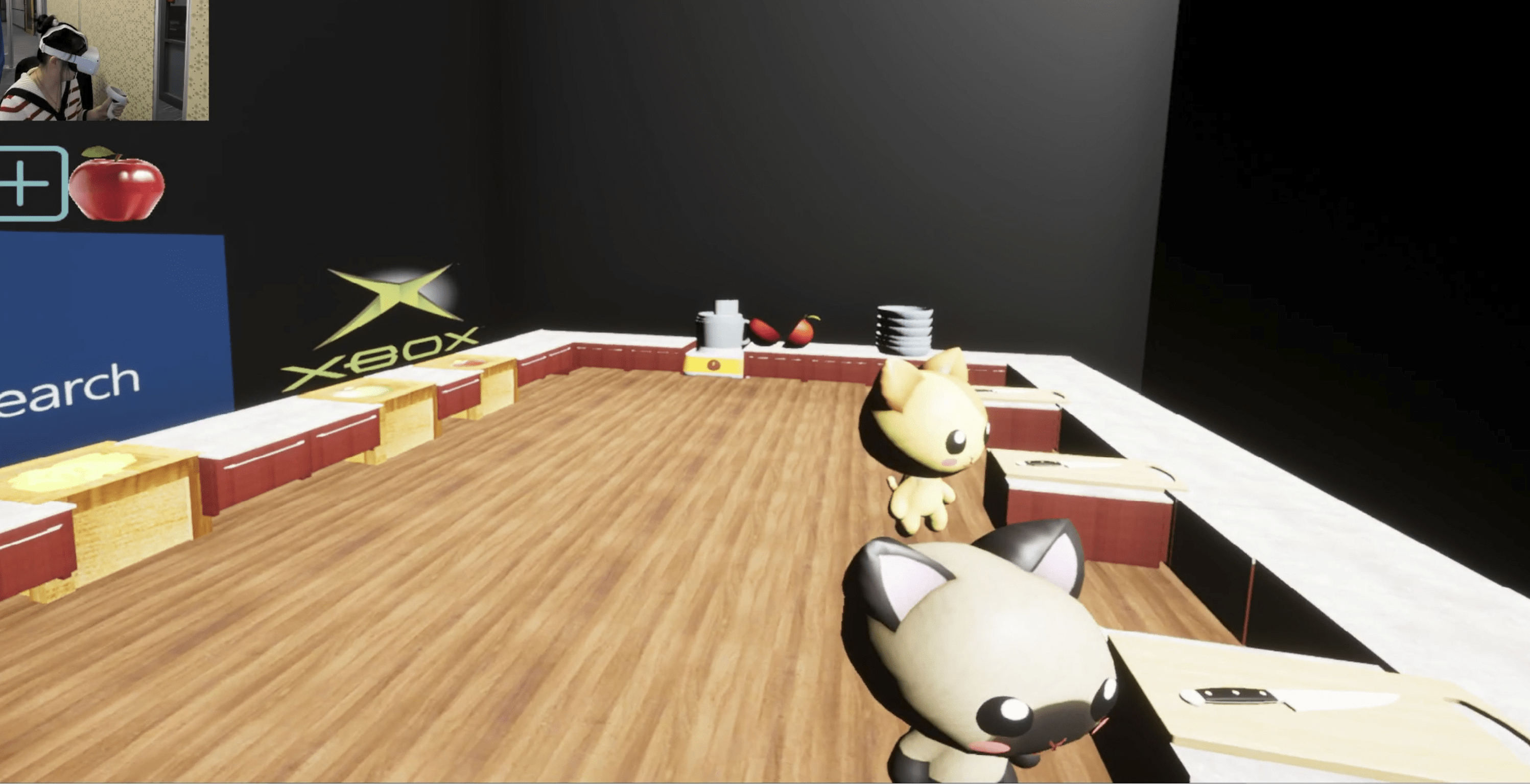} & \includegraphics[width=6cm]{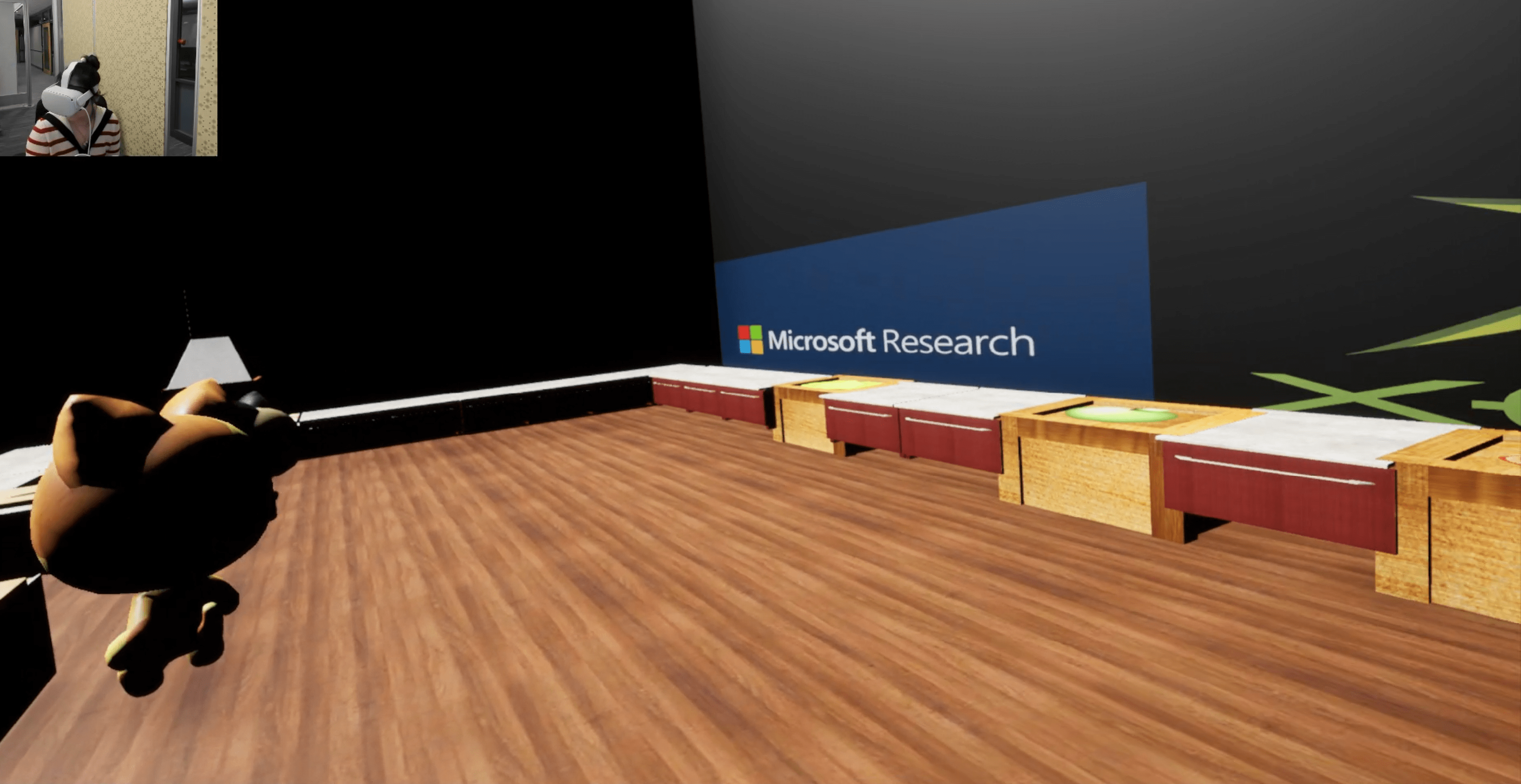} \\ 
    \end{tabular}
    \caption{The top two images show a multi-agent collaboration example in CuisineWorld, the three agents are preparing a mixed juice together. The middle two images show a human player as the head chef instructing the agents to cook mixed juice. The bottom two images  show a human player collaborating with collaborative agents in VR.}
    \label{fig:cusineworldMulti}
\end{figure}

To implement \overcook into a real game system,  we built on top of \citet{gao2020joint}.  In our game, as visually depicted in  \autoref{fig:cusineworldMulti}, players are given the opportunity to engage in collaborative interactions with NPCs. In this game, human players' actions can be obtained from an inverse dynamic model by checking preconditions and post-effects. This introduces a unique dynamic to the gameplay, enabling users to experience a more immersive cooperative environment. Additionally, the game's interface is versatile, allowing players multiple ways to interact within the game world. They can either use a standard keyboard setup, which is more conventional and likely familiar to most PC gamers, or they can immerse themselves even further using a Virtual Reality (VR) device. This VR functionality ensures a more tactile and realistic interaction, as players can physically move, gesture, and engage with the NPCs and other in-game elements in a 3D environment.

\section{Analysis and Emergent Gaming Abilities}

\subsection{Ablation Study for Multi-Agents}

\paragraph{Study on the Prompt Components Q3} In \autoref{tab:gpt-4_ablation}, we elucidate the performance of LLM dispatchers with certain components of the prompt omitted. Details about prompt can be found in Appendix \autoref{fig:one-shot} and \autoref{fig:prompt}. Specifically, for these tests, we excluded individual components like inference knowledge, reduced the prompt example to a mere two steps instead of the complete demonstration, and evaluated the model without environment feedback. For context, our principal experiments, varying in the number of agents, incorporate a one-shot example for the corresponding number of agents. Our ablation studies further probe how varying the number of agents can influence model performance, with details in \autoref{tab:agent_demo}.

\textbf{Findings: } From \autoref{tab:gpt-4_ablation}, a significant drop in performance is observed when environment feedback is excluded, underscoring its pivotal role in the efficacy of the LLM dispatcher. Replaying action sequences reveals that, without feedback, the LLM dispatcher tends to repeat mistakes and gets stuck in specific states for prolonged durations. Another key takeaway is that a succinct two-step demonstration of input and output format can still achieve commendable performance for unseen tasks with dynamic objectives. Notably, in these two-step instances, there's no explicit guide to finish any tasks. Yet, the model doesn't merely complete the task but continually performs additional tasks within the same episode. Furthermore, we also observe that integrating human-crafted inference knowledge bolsters the LLM dispatcher's performance. Lastly, even with few-shot demonstrations involving fewer agents, the LLM dispatcher retains satisfactory performance as shown in \autoref{tab:agent_demo}.

\textbf{Study on Other LLMs' Performance Q4}. To study how other LLMs perform on our tasks,   we tested the collaboration performance of GPT-3.5, Claude-2 and LLaMA in \autoref{tab:LLM_ablation}. For a fair comparison, all tests employed identical prompt inputs.

\textbf{Findings: } We observe that while other LLMs tend to underperform, models such as Claude-2 still manage to complete the task to a considerable extent.

\subsection{Emerging Capabilities}
Across our experiments, we observe the following emergent properties under our \mindagent framework.

\paragraph{Emergent Collaboration Tasks Understanding}
As shown in Table 7, especially in the few-step ablation entries, GPT-4 exhibits its proficiency even when not provided with a full demonstration for specific tasks. To clarify, a "full few-shot demo" typically refers to a comprehensive demonstration of a task, detailing each step and procedure involved. In contrast, we use provide GPT-4 with only a partial demonstration or a glimpse of the task only executing two steps.

Yet, despite this limited input, GPT-4's performance is remarkable. This underscores GPT-4's impressive \textbf{emergent zero-shot multi-agent planning} capabilities. Beyond simply completing unseen tasks, GPT-4 also demonstrates adaptability by dynamically prioritizing multiple different tasks as they arise, emphasizing its \textbf{emergent multi-task, on-the-fly planning} skills.

\paragraph{Emergent Multi-agent Reasoning Capabilities}
Referencing \autoref{tab:agent_demo}, GPT-4 has the capability to deploy more agents based on demonstrations of fewer agents. For instance, GPT-4 can effectively dispatch four agents having only seen demonstrations involving two agents. Moreover, the efficiency of collaboration is higher as the number of agents increases, spotlighting its \textbf{emergent collaboration} prowess.

% \begin{table*}[h!]
%     \centering
%     \resizebox{\textwidth}{!}{%
%     \begin{tabular}{|l|c|c|c|c|c|c|c|c|c|c|c|c|}
%     \hline
%     & \multicolumn{4}{c|}{2 agent} & \multicolumn{4}{c|}{3 agent} & \multicolumn{4}{c|}{4 agent} \\
%     \cline{2-13}
%     & GPT-4 & Claude-2 & LLaMA & ChatGPT & GPT-4 & Claude-2 & LLaMA & ChatGPT & GPT-4 & Claude-2 & LLaMA & ChatGPT \\
%     \hline
%     $\tau_{\text{int},(1)}$ & $10 / 26$ & $3 / 24$ & $0$ & $0 / 24$ & $12 / 25$ & $5 / 26$ & $0$ & $0 / 24$ & $16 / 27$ & $9 / 25$ & $0$ & $0 / 24$ \\
%     \hline
%     $\tau_{\text{int},(2)}$ & $10 / 17$ & $3 / 16$ & $0$ & $0 / 15$ & $14 / 20$ & $4 / 16$ & $0$ & $0 / 15$ & $16 / 19$ & $4 / 15$ & $0$ & $0 / 15$ \\
%     \hline
%     $\tau_{\text{int},(3)}$ & $11 / 18$ & $3 / 12$ & $0$ & $0 / 12$ & $13 / 14$ & $3 / 12$ & $0$ & $0 / 12$ & $15 / 17$ & $4 / 12$ & $0$ & $0 / 12$ \\
%     \hline
%     $\tau_{\text{int},(4)}$ & $11 / 13$ & $3 / 9$ & $0$ & $0 / 9$ & $10 / 10$ & $5 / 11$ & $0$ & $0 / 9$ & $12 / 13$ & $6 / 11$ & $0$ & $0 / 9$ \\
%     \hline
%     $\tau_{\text{int},(5)}$ & $11 / 11$ & $4 / 6$ & $0$ & $0 / 6$ & $12 / 12$ & $5 / 7$ & $0$ & $0 / 6$ & $12 / 12$ & $6 / 7$ & $0$ & $0 / 6$ \\
%     \hline
%     \colab & 0.686 &  0.3125  & 0 & 0 & 0.822 & 0.372 & 0 & 0 & 0.848 & 0.473 & 0 & 0 \\ 
%     \hline
%     \end{tabular}%
%     }
%     \caption{Performance of other LLMs on level 3}
%     \label{tab:LLM_ablation}
% \end{table*}

\begin{table*}[h!]
\centering
\resizebox{\textwidth}{!}{%
\begin{tabular}{l|c|c|c|c|c|c|c|c|c|c|c|c}
\toprule
& \multicolumn{4}{c}{2 agent} & \multicolumn{4}{c}{3 agent} & \multicolumn{4}{c}{4 agent} \\
\cmidrule(r){2-5} \cmidrule(lr){6-9} \cmidrule(l){10-13}
& GPT-4 & Claude-2 & LLaMA & ChatGPT & GPT-4 & Claude-2 & LLaMA & ChatGPT & GPT-4 & Claude-2 & LLaMA & ChatGPT \\
\midrule
$\tau_{\text{int},(1)}$ & $10 / 26$ & $3 / 24$ & $0$ & $0 / 24$ & $12 / 25$ & $5 / 26$ & $0$ & $0 / 24$ & $16 / 27$ & $9 / 25$ & $0$ & $0 / 24$ \\
$\tau_{\text{int},(2)}$ & $10 / 17$ & $3 / 16$ & $0$ & $0 / 15$ & $14 / 20$ & $4 / 16$ & $0$ & $0 / 15$ & $16 / 19$ & $4 / 15$ & $0$ & $0 / 15$ \\
$\tau_{\text{int},(3)}$ & $11 / 18$ & $3 / 12$ & $0$ & $0 / 12$ & $13 / 14$ & $3 / 12$ & $0$ & $0 / 12$ & $15 / 17$ & $4 / 12$ & $0$ & $0 / 12$ \\
$\tau_{\text{int},(4)}$ & $11 / 13$ & $3 / 9$ & $0$ & $0 / 9$ & $10 / 10$ & $5 / 11$ & $0$ & $0 / 9$ & $12 / 13$ & $6 / 11$ & $0$ & $0 / 9$ \\
$\tau_{\text{int},(5)}$ & $11 / 11$ & $4 / 6$ & $0$ & $0 / 6$ & $12 / 12$ & $5 / 7$ & $0$ & $0 / 6$ & $12 / 12$ & $6 / 7$ & $0$ & $0 / 6$ \\
CoS & 0.686 & 0.3125 & 0 & 0 & 0.822 & 0.372 & 0 & 0 & 0.848 & 0.473 & 0 & 0 \\
\bottomrule
\end{tabular}%
}
\caption{Performance of Other LLMs on Level 3}
\label{tab:LLM_ablation}
\end{table*}

%  \begin{table*}[h]
 
%  \centering
%      \resizebox{\textwidth}{!}{\begin{tabular}{|l|l|l|l|l|}
% \hline 2 agent & GPT-4 & GPT-4 w/ just few-step & GPT-4 w/o inference knowledge & GPT-4 w/o feedback\\
% \hline $\tau_{\text{int},(1)}$ & $10 / 26$ & $8 / 26$  & 8/25 & 4/25\\
% \hline $\tau_{\text{int},(2)}$ & $10 / 17$ & $11 / 19$ & 9/17 & 4/17\\
% \hline $\tau_{\text{int},(3)}$ & $11 / 13$ & $11 / 13$ & 10/12& 4/12 \\
% \hline $\tau_{\text{int},(4)}$ & $12 / 12$ & $9 / 11$  & 8/9  & 1/9\\
% \hline $\tau_{\text{int},(5)}$ & $11 / 11$ & $10 / 10$ & 9/9  & 5/7\\
% \hline \colab & 0.764 & 0.710     & 0.714& 0.311\\
% \hline
% \end{tabular}}

% \caption{Additional ablation}
% \label{tab:gpt-4_ablation}
%  \end{table*}

\begin{table*}[h]
\centering
\resizebox{\textwidth}{!}{
\begin{tabular}{l|l|l|l|l}
\toprule
2 agent & GPT-4 & GPT-4 w/ few-step & GPT-4 w/o inference knowledge & GPT-4 w/o feedback \\
\midrule
$\tau_{\text{int},(1)}$ & $10 / 26$ & $8 / 26$  & 8/25 & 4/25 \\
$\tau_{\text{int},(2)}$ & $10 / 17$ & $11 / 19$ & 9/17 & 4/17 \\
$\tau_{\text{int},(3)}$ & $11 / 13$ & $11 / 13$ & 10/12& 4/12 \\
$\tau_{\text{int},(4)}$ & $12 / 12$ & $9 / 11$  & 8/9  & 1/9 \\
$\tau_{\text{int},(5)}$ & $11 / 11$ & $10 / 10$ & 9/9  & 5/7 \\
CoS & 0.764 & 0.710     & 0.714& 0.311 \\
\bottomrule
\end{tabular}
}
\caption{Additional Ablation}
\label{tab:gpt-4_ablation}
\end{table*}

% \begin{table*}[h]
%     \centering
%     \resizebox{\textwidth}{!}{\begin{tabular}{|l|l|l|l|l|}
% \hline level\_3 & 4agent using 4agent demo & 4agent using 2agent demo & 3agent using 3agent demo & 3agent using 2agent demo\\
% \hline GPT4 $\tau_{\text{int},(1)}$ & $16 / 27$ & $14 / 27$ & $12 / 25$ & 11/25 \\
% \hline GPT4 $\tau_{\text{int},(2)}$ & $16 / 19$ & $16 / 20$ & $14 / 20$ & 11/19 \\
% \hline GPT4 $\tau_{\text{int},(3)}$ & $15 / 17$ & $15 / 16$ & $13 / 14$ & 12/14 \\
% \hline GPT4 $\tau_{\text{int},(4)}$ & $12 / 13$ & $13 / 13$ & $10 / 10$ & 12/12 \\
% \hline GPT4 $\tau_{\text{int},(5)}$ & $12 / 12$ & $12 / 12$ & $12 / 12$ & 11/11 \\
% \hline \colab  & 0.848 & 0.851 & 0.822 & 0.775\\ 
% \hline
% \end{tabular}
% }
% \caption{Using different numbers of agent demos}
% \label{tab:agent_demo}
% \end{table*}

\begin{table*}[h]
\centering
\resizebox{\textwidth}{!}{
\begin{tabular}{l|l|l|l|l}
\toprule
level\_3 & 4agent using 4agent module & 4agent using 2agent module & 3agent using 3agent module & 3agent using 2agent module\\
\midrule
GPT4 $\tau_{\text{int},(1)}$ & $16 / 27$ & $14 / 27$ & $12 / 25$ & 11/25 \\
GPT4 $\tau_{\text{int},(2)}$ & $16 / 19$ & $16 / 20$ & $14 / 20$ & 11/19 \\
GPT4 $\tau_{\text{int},(3)}$ & $15 / 17$ & $15 / 16$ & $13 / 14$ & 12/14 \\
GPT4 $\tau_{\text{int},(4)}$ & $12 / 13$ & $13 / 13$ & $10 / 10$ & 12/12 \\
GPT4 $\tau_{\text{int},(5)}$ & $12 / 12$ & $12 / 12$ & $12 / 12$ & 11/11 \\
CoS  & 0.848 & 0.851 & 0.822 & 0.775 \\
\bottomrule
\end{tabular}
}
\caption{Using different numbers of agent demos}
\label{tab:agent_demo}
\end{table*}

\section{Novel Game Adaptation}
% In addition to our \overcook game, we adapted our game to a cooking task in Minecraft. Agents, Alex and Steve,  need to cook different types of meat according to the system generated task goals. Our results indicate that our infrastructure can generalize to multiple different game domains. 

% In addition, we connect Azure speech-to-text API with Minecraft so human players can collaborate with in-game NPC agents under our framework through voice chat.

In line with our ongoing efforts to create collaborative, in-game, multi-agent systems, we ventured beyond CuisineWorld and made strides in integrating our infrastructure into the widely popular sandbox game, Minecraft. In this new adaptation, we designed several unique cooking tasks where two in-game agents, Alex and Steve, are assigned the responsibility of cooking various types of meat as shown in \autoref{fig:minecraft4}. After cooking, agents need to deposit the items into a chest. More details can be found in Appendix~C. The experiment results are presented in \autoref{tab:GPT4_minecraft}.

We define the following actions for the multi-agent system in our Minecraft game: 1) \texttt{goto(agent, location)}; 2) \texttt{killMob(agent, mobType)}; 3) \texttt{mineBlock(agent, blockType)}; 4) \texttt{putFuelFurnace(agent, fuelType)}, to put the item from agent's inventory to the furnace's bottom slot. 5) \texttt{putItemFurnace(agent, itemType)}, to put the item from agent's inventory to the furnace's top slot; 6) \texttt{takeOutFurnace(agent)}, take out the cooked item from the furnace 7) \texttt{ putInChest(agent, itemType) };

The state space in Minecraft contains the following: 1) nearby blocks for each agent 2) nearby entities for each agent. 3) each agent's inventory 4) items inside the furnace 5) items inside the chest. 6) human player's inventory if a human player is involved.

To ensure reproducibility, we modify the game mechanism. A killed mob will respawn nearby, and a mined block will also respawn nearby. 

The empirical data we collected from these game sessions provided us with compelling evidence that the multi-agent collaboration infrastructure we've developed has the robustness to be extrapolated and adapted across multiple distinct games, paving the way for broader applications in the gaming industry.

Going a step further, we bridged the gap between human players and in-game (NPC) agents by integrating Microsoft's Azure speech-to-text API into the Minecraft environment. This addition allows human players to communicate and collaborate with in-game NPC agents using voice chat. Human players can express their intents and desired goals to NPCs in real-time through voice chat. This real-time vocal interaction enriches the gameplay experience, fostering a deeper level of immersion and synergy between human players and AI agents. Moreover, this integration opens the door for research into the efficacy of voice-assisted AI learning and how real-world human interactions can shape AI behavior in virtual domains.

In the case of the human player chatting with the multi-agent system, the prompt contains additional human instructions and human dialog history components.

In addition, by integrating Minecraft VR mode with our infrastructure, we can bring the player interactive experiences to the next level. 
\iffalse
\begin{figure}[h!]
    \centering
    \setlength{\tabcolsep}{2pt}  % Adjust this to control the space between columns
    \begin{tabular}{m{1cm}@{\hspace{-10pt}} m{6cm} m{6cm}}
        \rotatebox[origin=l]{90}{\textbf{Multi-agent}} & \includegraphics[width=6cm]{iclr2024/Figures/minecraft1.png} & \includegraphics[width=6cm]{iclr2024/Figures/minecraft2.png} \\ 
        \rotatebox[origin=l]{90}{\textbf{Human-agent}} & \includegraphics[width=6cm]{iclr2024/Figures/minecraftplay1.jpg} & \includegraphics[width=6cm]{iclr2024/Figures/minecraftplay2.jpg} \\ 
        \rotatebox[origin=l]{90}{\textbf{VR Interaction}} & \includegraphics[width=6cm]{iclr2024/Figures/minecraftVR1.png} & \includegraphics[width=6cm]{iclr2024/Figures/mineCraftVR2.png} \\ 
    \end{tabular}
    \caption{The top two images show a multi-agent collaboration example in Minecraft. In the left image, Alex and Steve are killing different animals, and in the right image, Alex and Steve are cooking meat in a furnace together. The middle two images show a human player instructing the agents to perform certain actions. The bottom two images show a human player collaborating with agents in VR.}
    \label{fig:minecraft4}
\end{figure}
\fi

\begin{figure}[h!]
    \centering
    \setlength{\tabcolsep}{2pt}  % Adjust this to control the space between columns
    \begin{tabular}{m{1cm}@{\hspace{-10pt}} m{6cm} m{6cm}}
        \rotatebox[origin=l]{90}{\textbf{Multi-agent}} & \includegraphics[width=6cm]{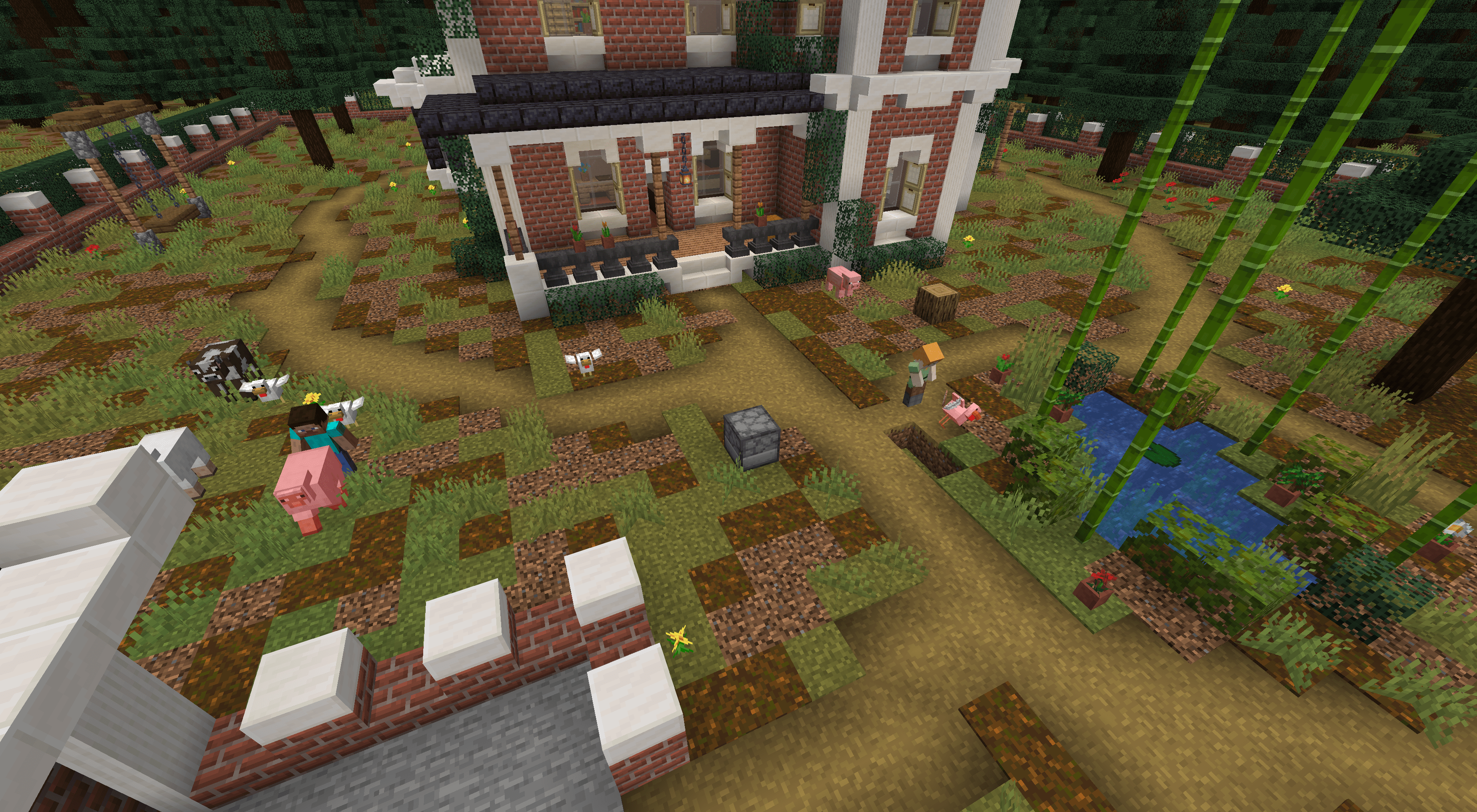} & \includegraphics[width=6cm]{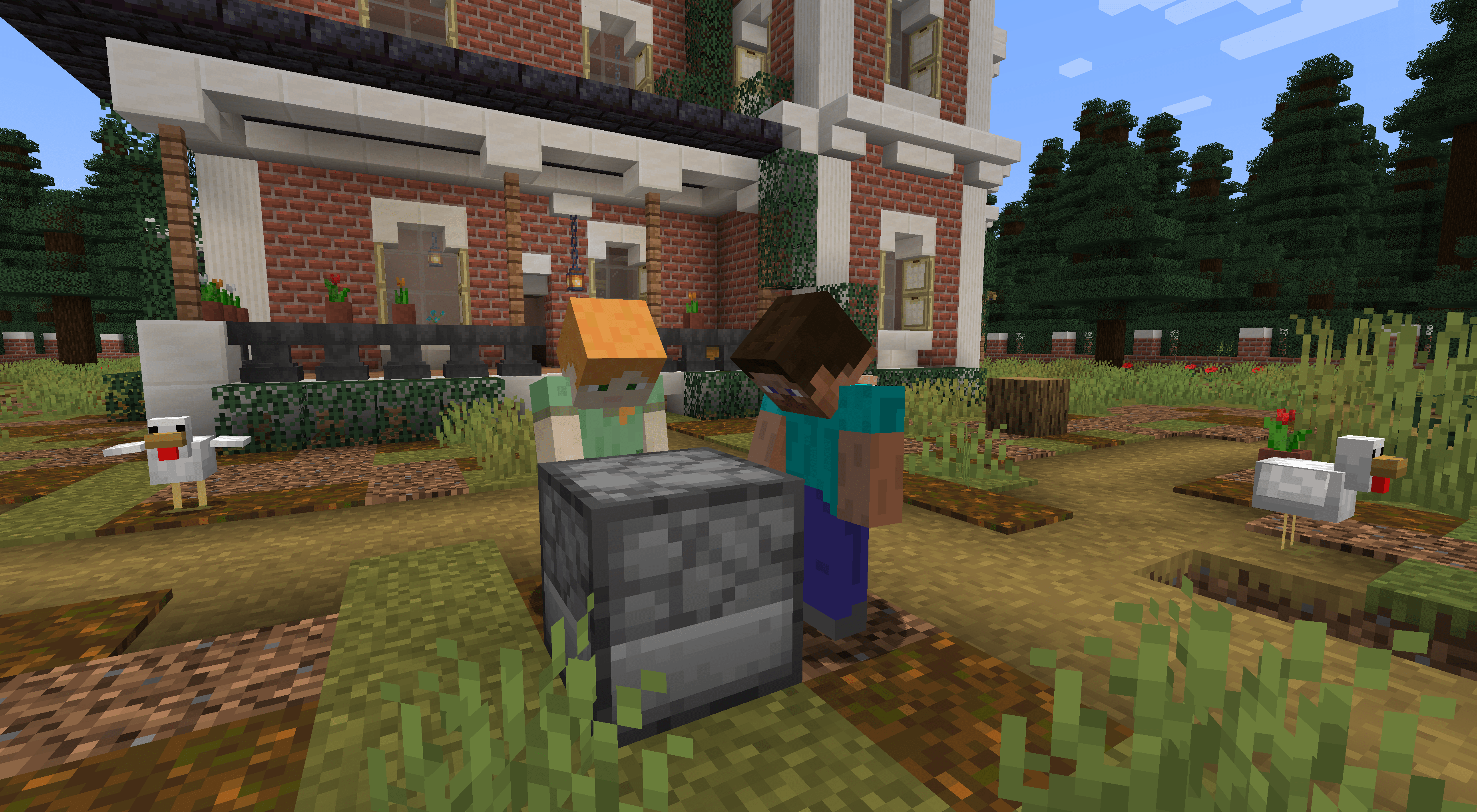} \\ 
        \rotatebox[origin=l]{90}{\textbf{Human-agent}} & \includegraphics[width=6cm]{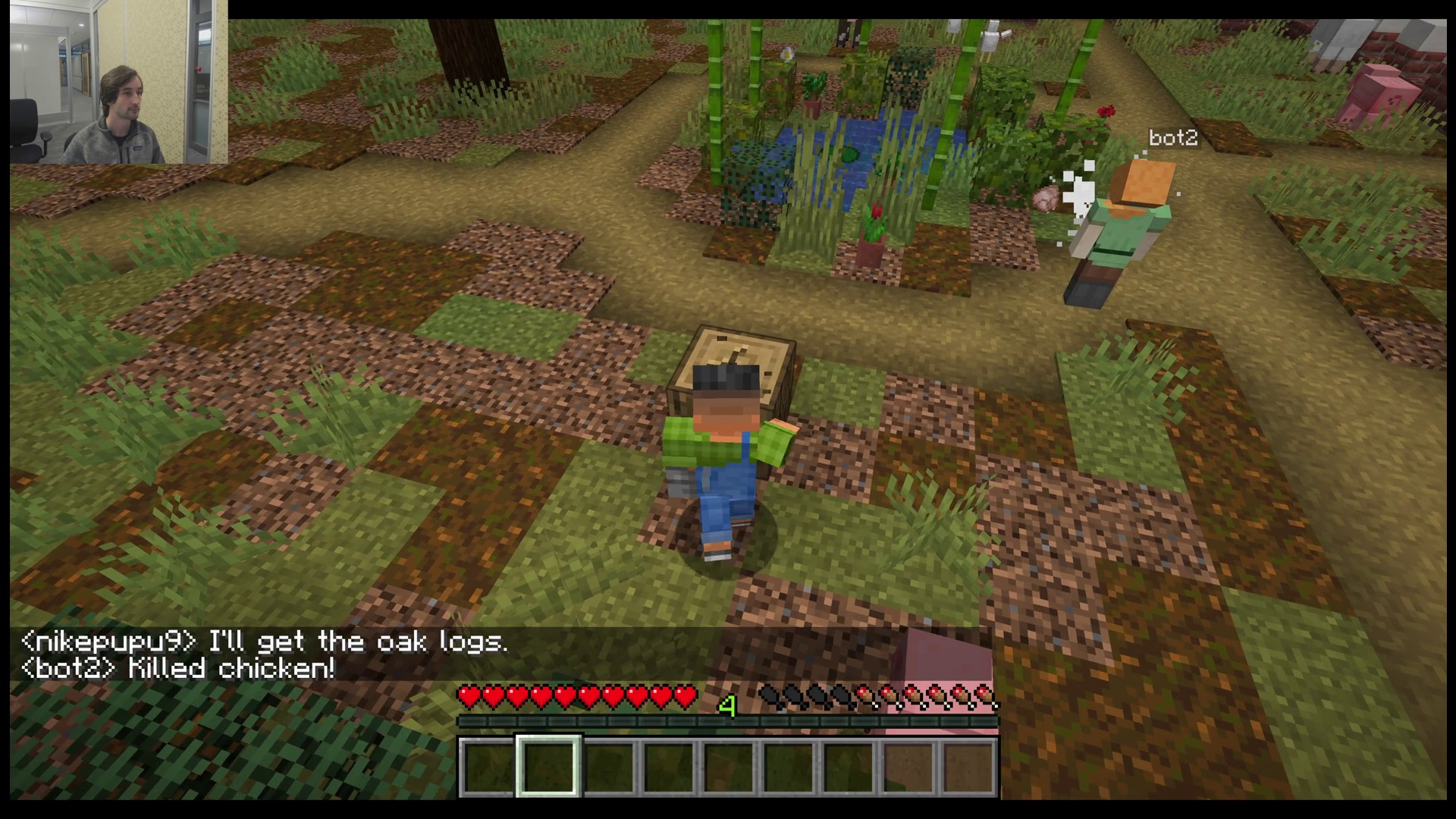} & \includegraphics[width=6cm]{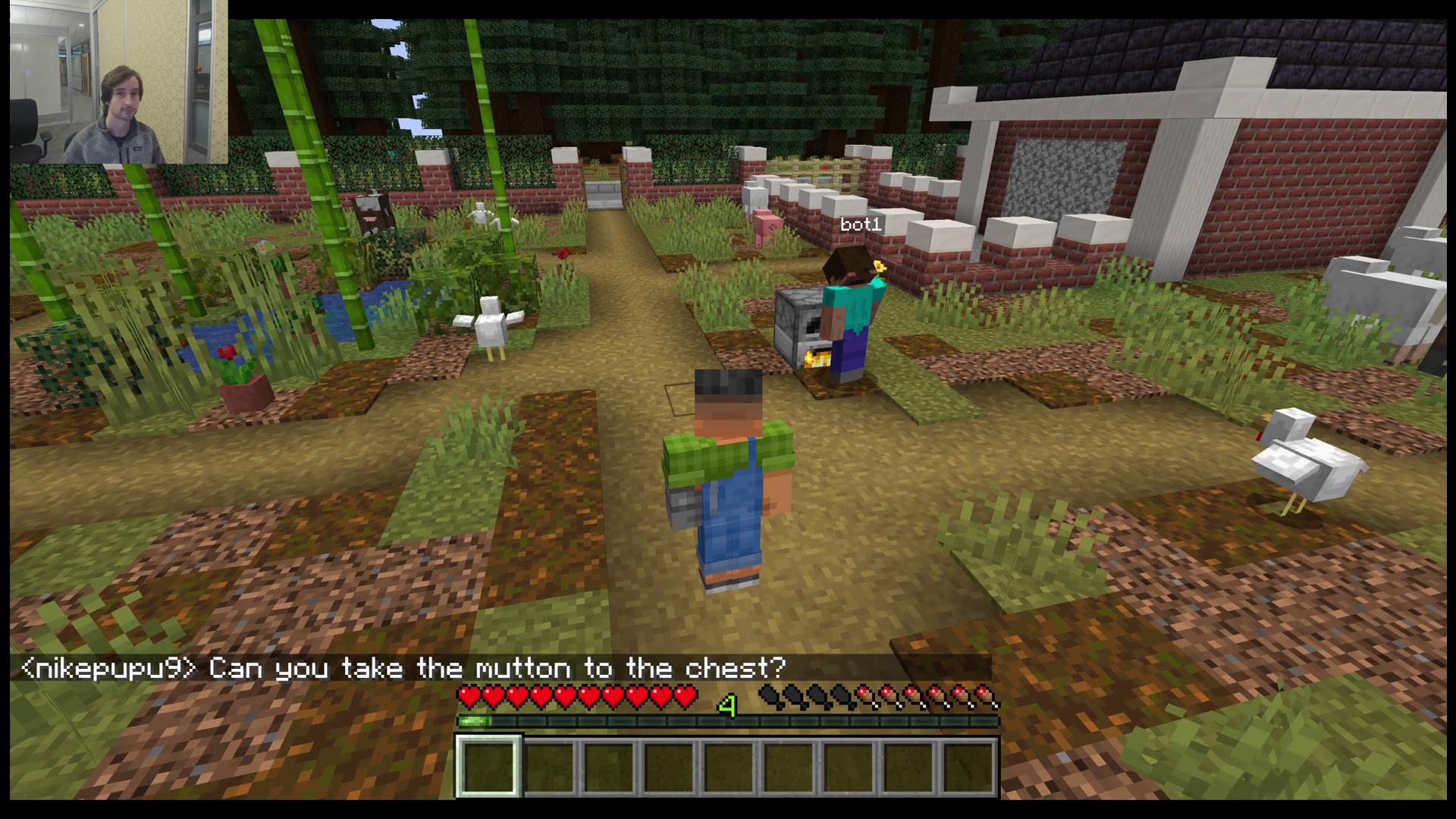} \\ 
        \rotatebox[origin=l]{90}{\textbf{VR Interaction}} & \includegraphics[width=6cm]{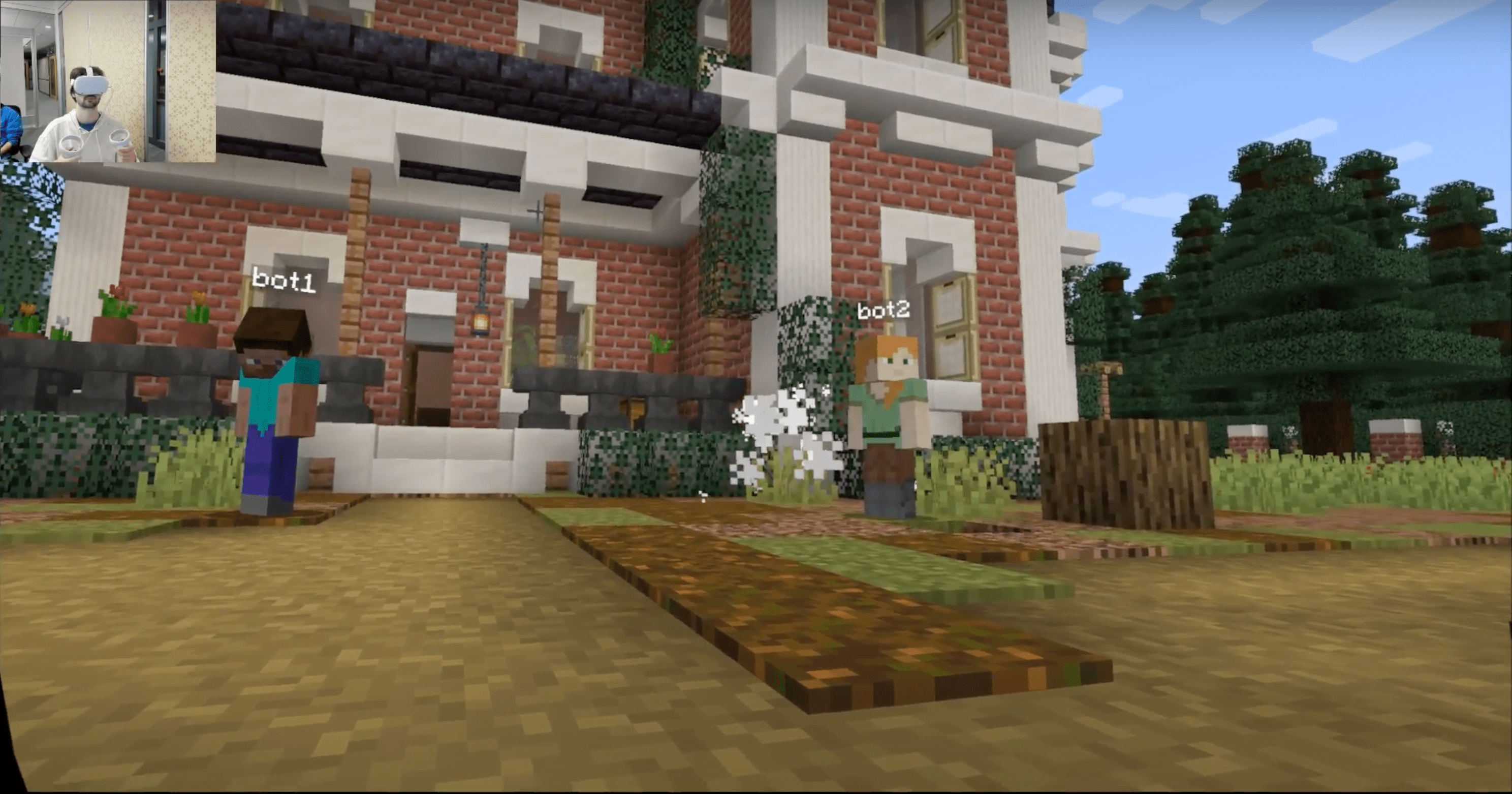} & \includegraphics[width=6cm]{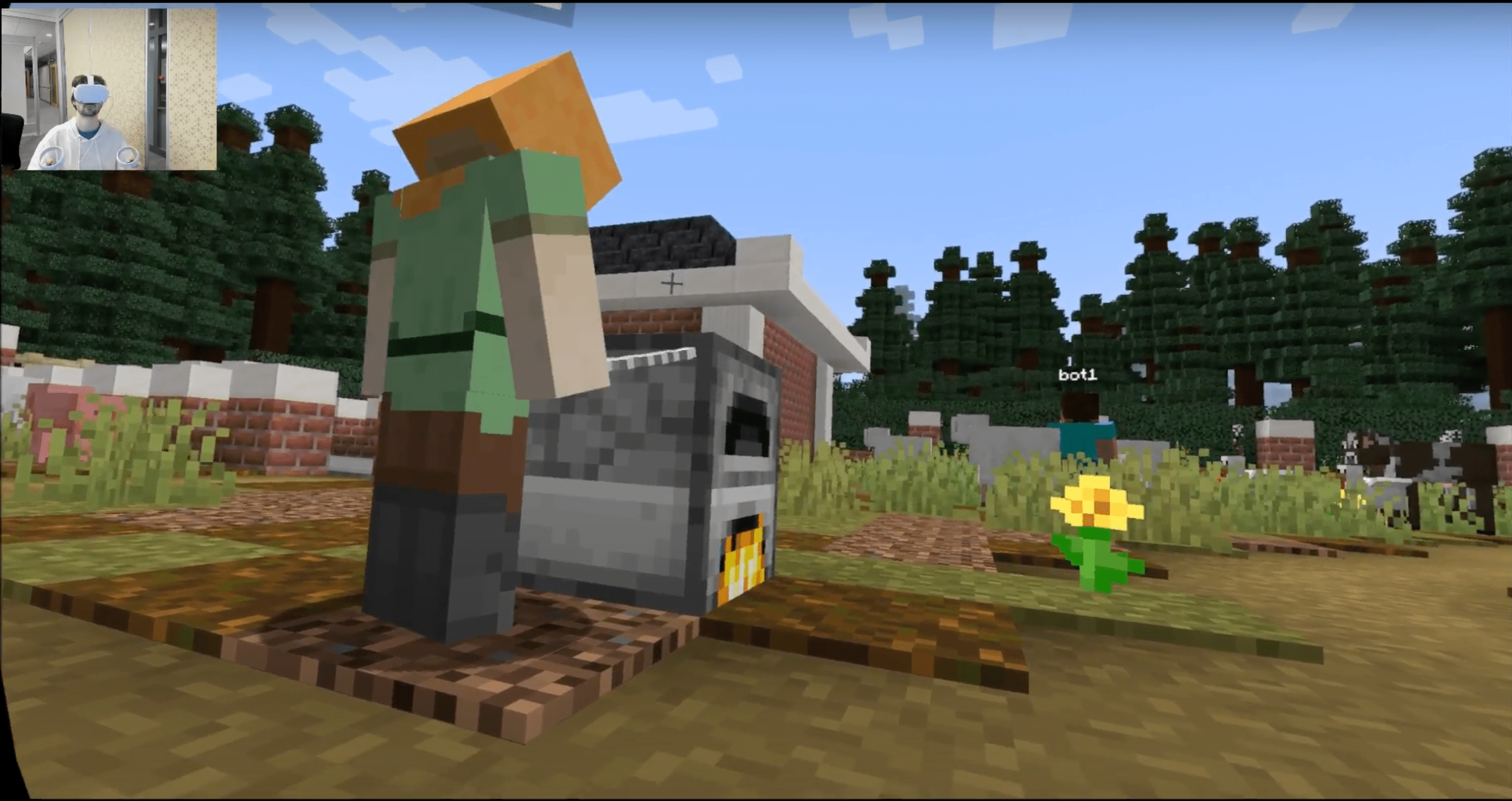} \\ 
    \end{tabular}
    \caption{The top two images show a multi-agent collaboration example in Minecraft. In the left image, Alex and Steve are killing different animals, and in the right image, Alex and Steve are cooking meat in a furnace together. The middle two images show a human player instructing the agents to perform certain actions. The bottom two images show a human player collaborating with agents in VR.}
    \label{fig:minecraft4}
\end{figure}

    %\begin{subfigure}[b]{0.49\textwidth}
    %    \centering
    %    \includegraphics[width=\textwidth]{iclr2024/Figures/minecraft1.png}
        
    %    \label{fig:minecraft1}
    %\end{subfigure}%
    ~ %add desired spacing between images, e.g. ~, \quad, \qquad, \hfill etc.
    %\begin{subfigure}[b]{0.49\textwidth}
    %    \centering
    %    \includegraphics[width=\textwidth]{iclr2024/Figures/minecraft2.png}
        
    %    \label{fig:minecraft2}
    %\end{subfigure}
    % \begin{subfigure}[b]{0.49\textwidth}
    %    \centering
    %    \includegraphics[width=\textwidth]{iclr2024/Figures/minecraftVR1.png}
        
    %    \label{fig:minecraft3}
    %\end{subfigure}
     %\begin{subfigure}[b]{0.49\textwidth}
     %   \centering
     %   \includegraphics[width=\textwidth]{iclr2024/Figures/mineCraftVR2.png}
        
     %   \label{fig:minecraft4}
    %\end{subfigure}
    %\caption{The top figures are a two-agent collaboration example in Minecraft. In the left image, Alex and Steve are killing different animals. In the right image, Alex and Steve are cooking meat in a furnace together. The bottom figure demonstrates human players collaborating with agents in Minecraft through VR.}
%\end{figure}   

% \begin{table*}[h!]
%     \centering
%     \begin{tabular}{|c|c|c|c|c|c|c|}
%     \hline
%     {GPT-4 minecraft} & $\tau_{\text{int},(1)}$ & $\tau_{\text{int},(2)}$ & $\tau_{\text{int},(3)}$  & $\tau_{\text{int},(4)}$ & $\tau_{\text{int},(5)}$ & CoS\\
%     \hline
%     Performance & $0.195$ & $ 0.381$ & $ 0.704 $ & $0.792$ & $0.833$ & $0.581$  \\
%     \hline
%     \end{tabular}%
%     \caption{Performance of our framework in Minecraft}
%     \label{tab:GPT4_minecraft}
% \end{table*}
\begin{table*}[h!]
\centering
\begin{tabular}{l|c|c|c|c|c|c}
\toprule
GPT-4 minecraft & $\tau_{\text{int},(1)}$ & $\tau_{\text{int},(2)}$ & $\tau_{\text{int},(3)}$ & $\tau_{\text{int},(4)}$ & $\tau_{\text{int},(5)}$ & \colab \\
\midrule
Performance & $0.195$ & $ 0.381$ & $ 0.704$ & $0.792$ & $0.833$ & $0.581$ \\
\bottomrule
\end{tabular}
\caption{Performance of our framework in Minecraft}
\label{tab:GPT4_minecraft}
\end{table*}

\section{Conclusion}
\label{sec:Conclusion}
In this paper, we presented \mindagent, an infrastructure for multi-agent collaboration through LLMs across multiple gaming domains. We investigated the multi-agent planning capabilities of \mindagent, and we deployed our infrastructure into real-world video games to demonstrate its effectiveness for multi-agent collaboration and human-AI collaboration. Beyond its practical applications, we hope that our endeavor serves as a beacon, guiding the development of future gaming systems where human-AI collaboration is seamless and intuitive. Furthermore, we are optimistic that our insights and findings might catalyze innovations in crafting games that are not only technologically advanced but also significantly more engaging and enjoyable for players.

\section*{Acknowledgments}
We are especially grateful to Johannes Gehrke, Ryen White, Haiyan Zhang, Kareem Choudhry for their enormous advice, support and encouragement of the work. We appreciate Katja Hofmann, Andrzej Banburski-Fahey, Jianwei Yang, Michel Galley, Nebojsa Jojic, Bill Dolan for the early insightful discussions, suggestions and comments. The authors gratefully acknowledge Adrian Brown from X-Box team for his discussion, feedback and pointers to the modeling generation and literature. We thank Rohan Taori, Janardhan Kulkarni, Ziheng Zhou, Yu Wang, Eloi Moliner Juanpere, Xiaofeng Gao, Collin Huang, Xiaodong Yu, and Shuwen Qiu for their help on the human experiment setup. 

\clearpage

\bibliography{iclr2024_conference}
\bibliographystyle{iclr2024_conference}

\vspace{30mm}

\pagebreak
\appendix
\section*{Appendix}

\section{Prompt Examples}
\begin{figure}[H]
    \centering
    \captionsetup{type=figure}
    \includegraphics[width=0.9\linewidth]{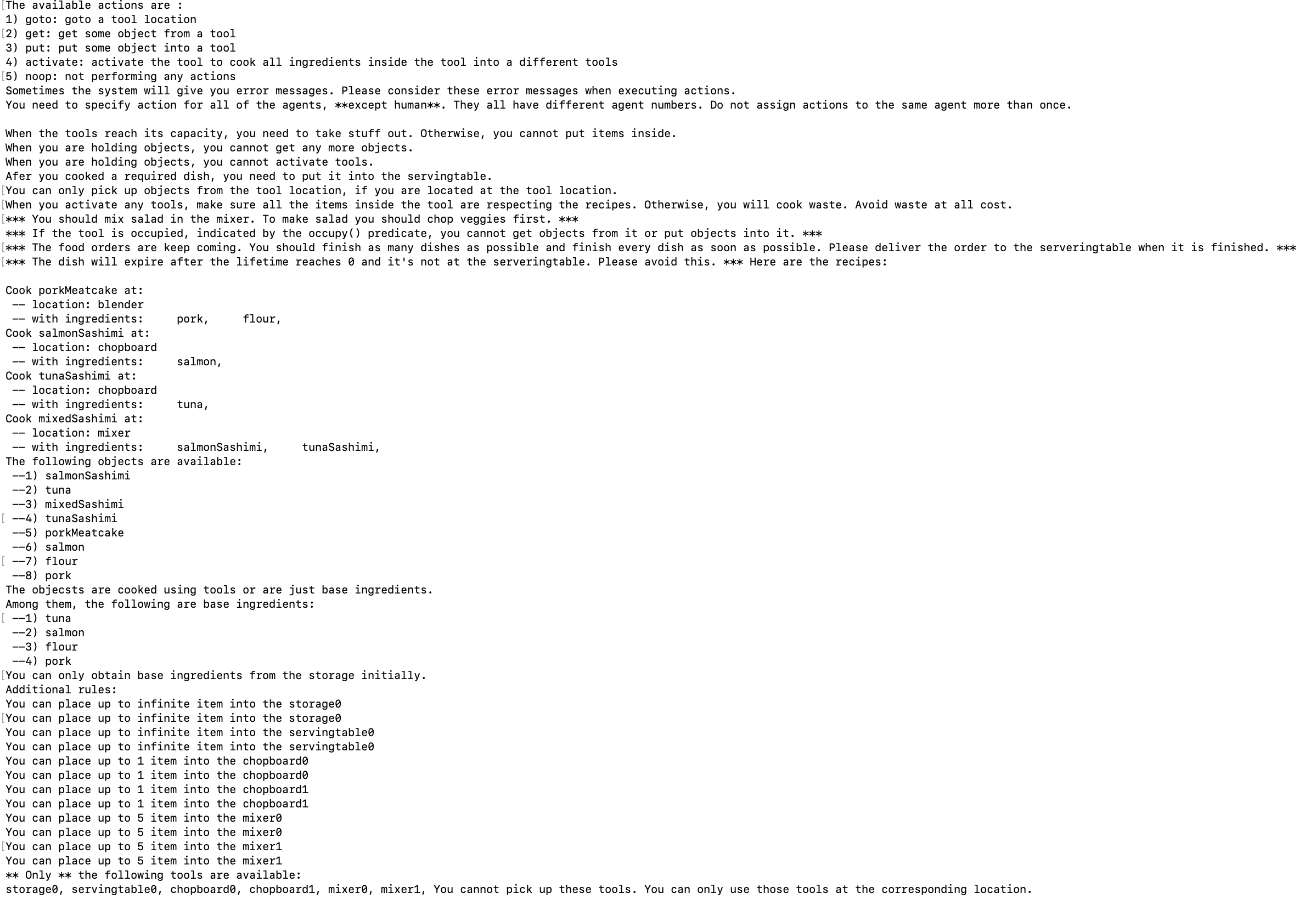}
    \captionof{figure}{ The \mindagent system prompt example.
    }
    \label{fig:prompt}
\end{figure}

\begin{figure}[H]
    \centering
    \captionsetup{type=figure}
    \includegraphics[height=0.4\textheight, keepaspectratio]{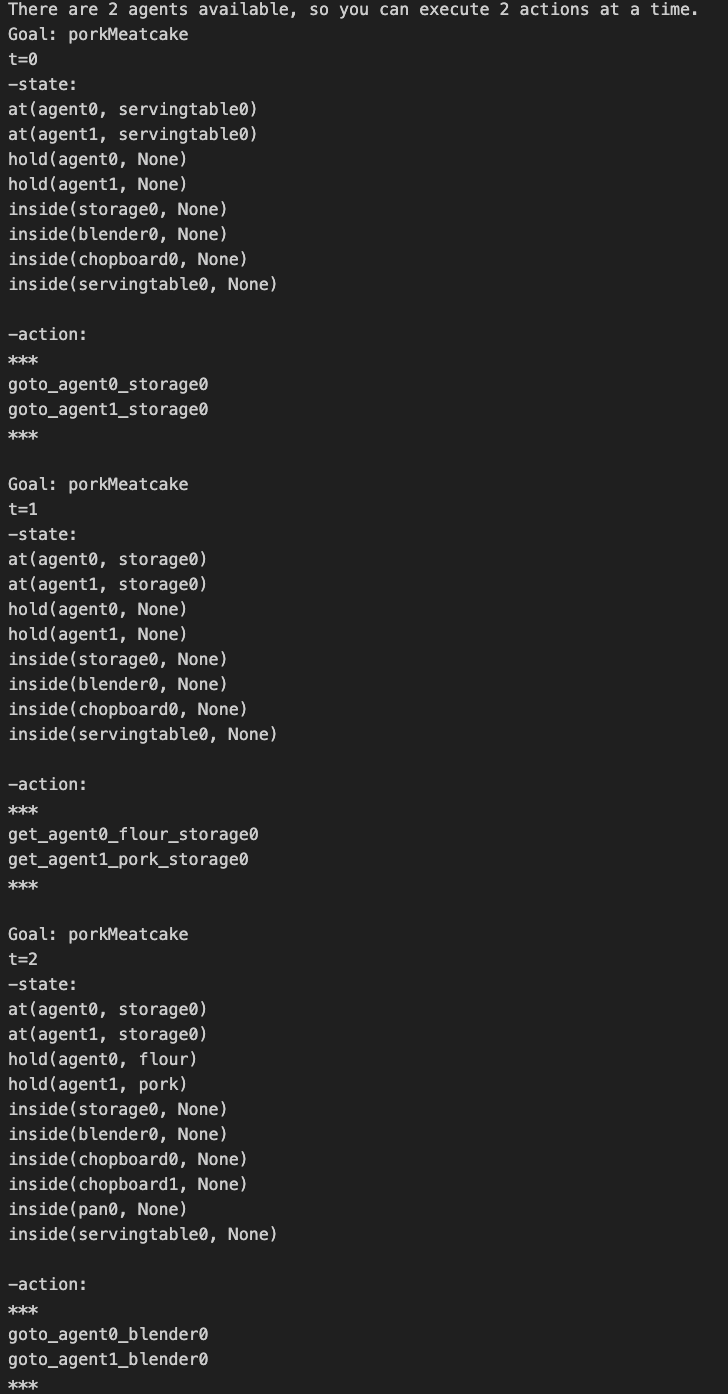}
    \captionof{figure}{ The \mindagent system partial one-shot demo example.
    }
    \label{fig:one-shot}
\end{figure}

We provide some prompt examples for CuisineWorld. \autoref{fig:prompt} shows an example of the system prompt info.  \autoref{fig:one-shot} shows an example of a partial demonstration. 

\pagebreak
%You may include other additional sections here.

\section{Task Details in CuisineWorld}
Here we visualize different task graphs in \overcook. In \overcook, we provide tasks of different complexities to holistically evaluate the multi-agent system's performance. In addition, the environment is highly customizable and extendable. Users only need to modify the JSON files to add more tasks or modify existing tasks. 
% Level 0
\subsection{Level 0}
\begin{figure}[H]
    \centering
    \includegraphics[width=0.3\linewidth]{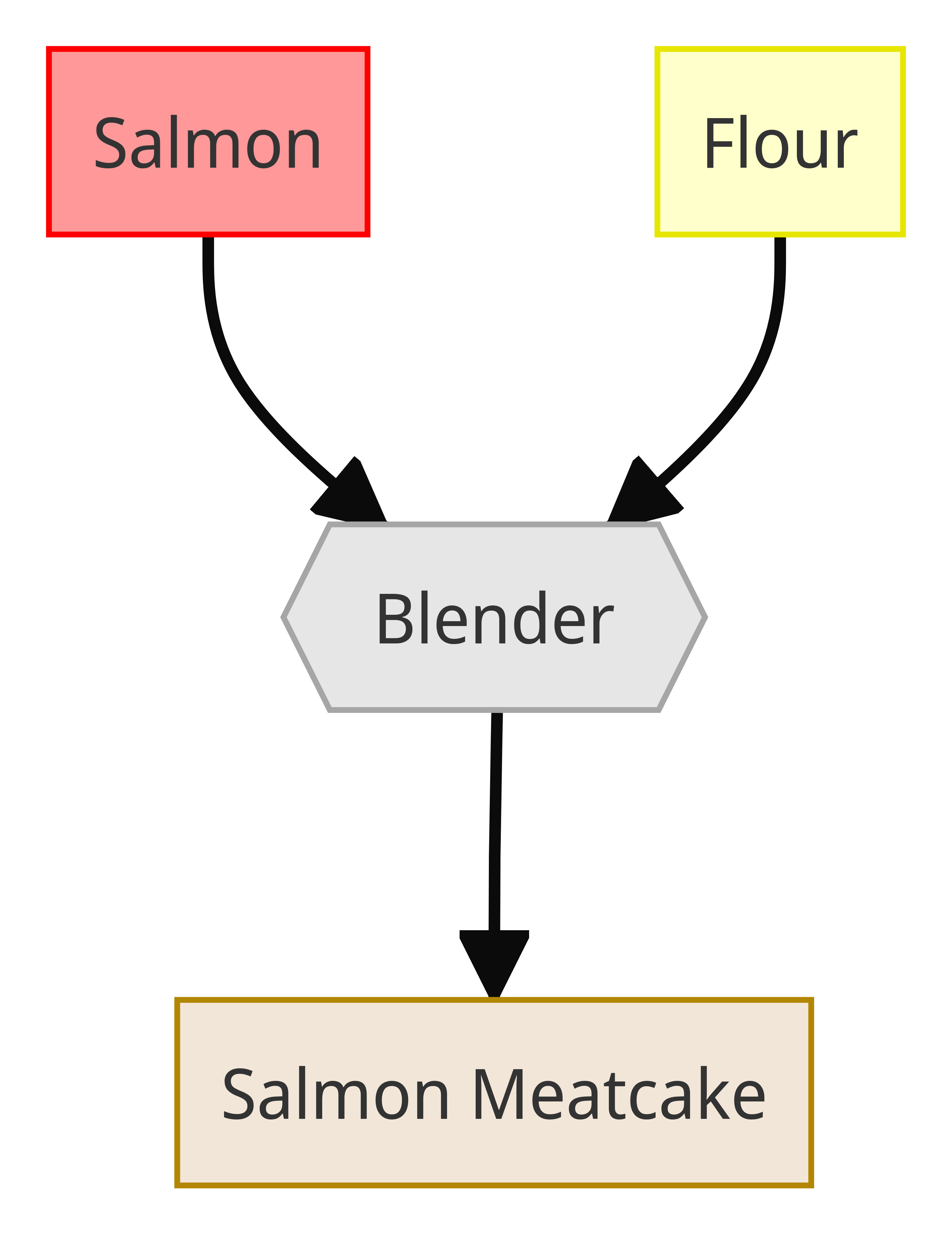}
    \caption{Salmon Meatcake}
\end{figure}

% Level 1
\subsection{Level 1}
\begin{figure}[H]
    \begin{subfigure}{0.3\textwidth}
        \includegraphics[width=\linewidth]{iclr2024/task_figures/salmonMeatcake.png}
        \caption{Salmon Meatcake}
    \end{subfigure}
    \hfill
    \begin{subfigure}{0.3\textwidth}
        \includegraphics[width=\linewidth]{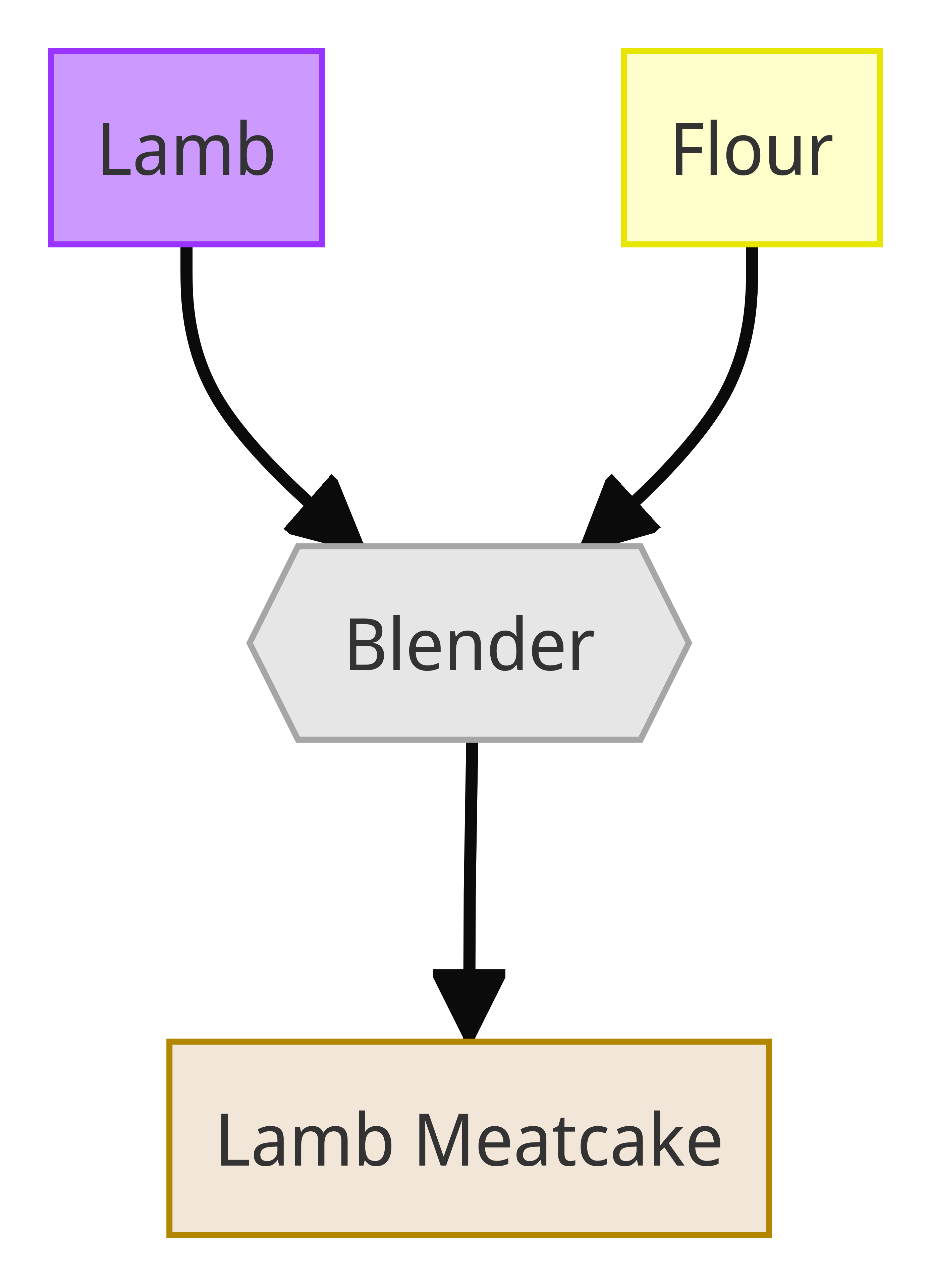}
        \caption{Lamb Meatcake}
    \end{subfigure}
    \hfill
    \begin{subfigure}{0.3\textwidth}
        \includegraphics[width=\linewidth]{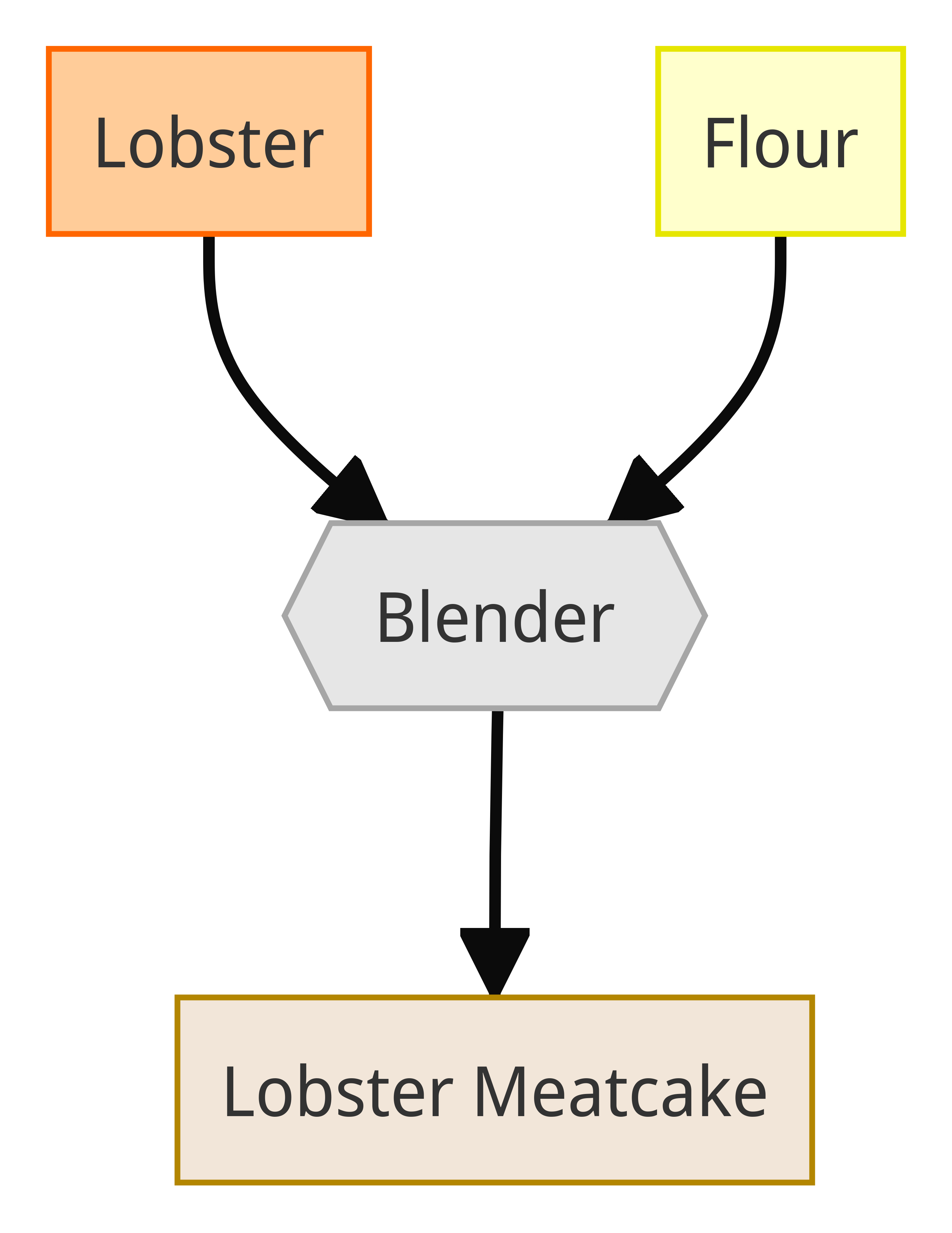}
        \caption{Lobster Meatcake}
    \end{subfigure}
\end{figure}

% Level 2
\subsection{Level 2}
\begin{figure}[H]
    \begin{subfigure}{0.3\textwidth}
        \includegraphics[width=\linewidth]{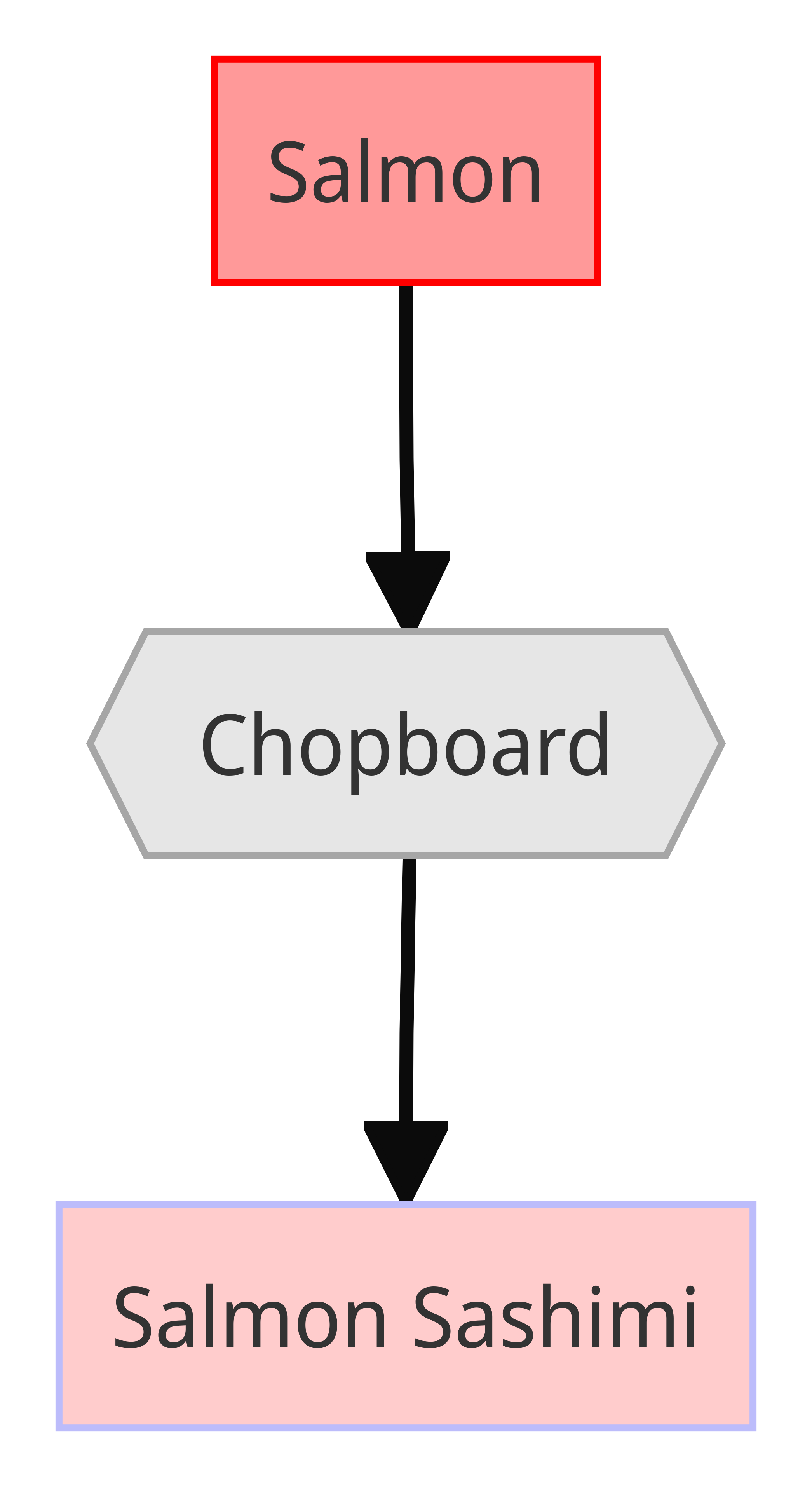}
        \caption{Salmon Sashimi}
    \end{subfigure}
    \hfill
    \begin{subfigure}{0.3\textwidth}
        \includegraphics[width=\linewidth]{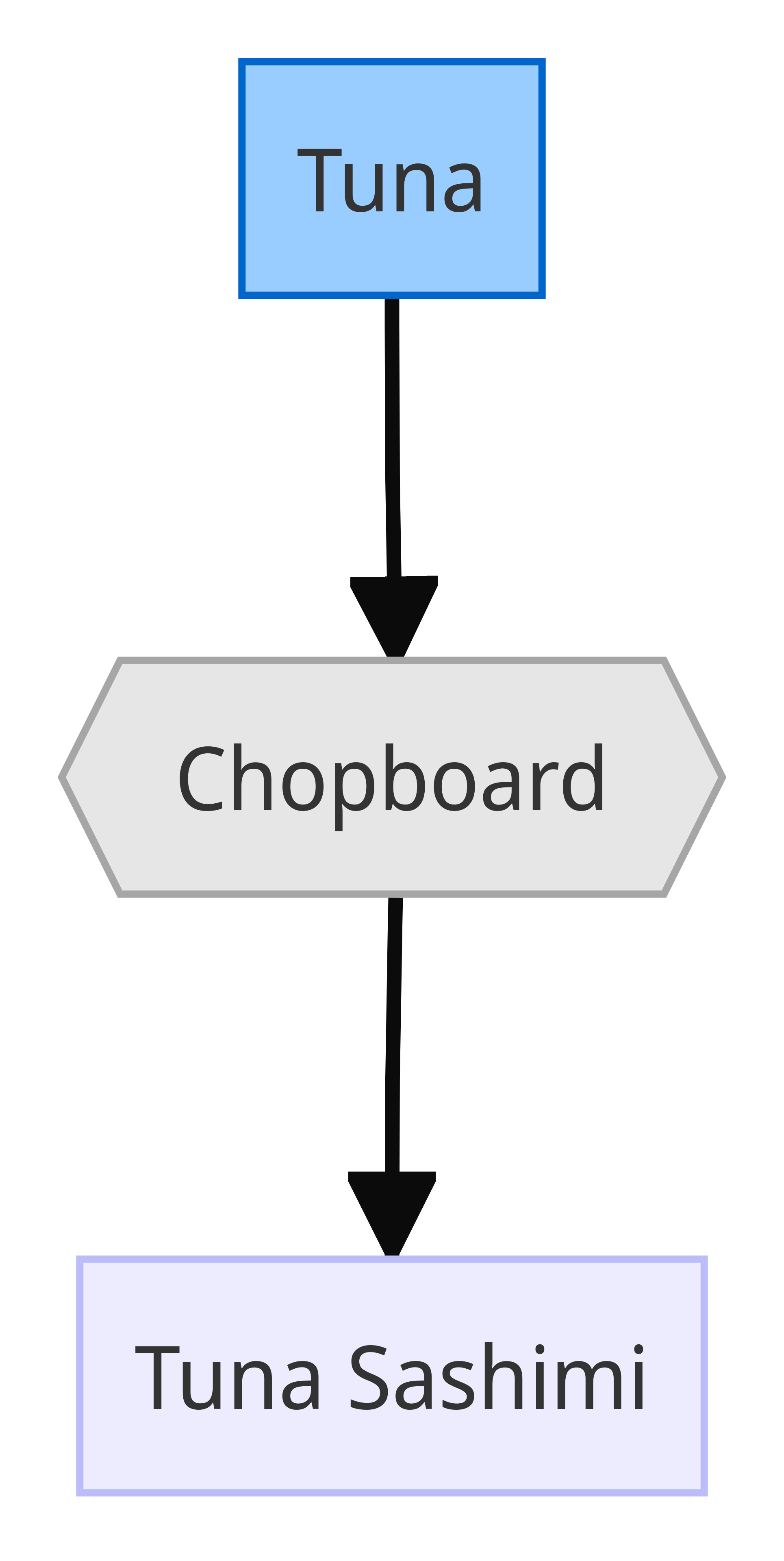}
        \caption{Tuna Sashimi}
    \end{subfigure}
    \hfill
    \begin{subfigure}{0.3\textwidth}
        \includegraphics[width=\linewidth]{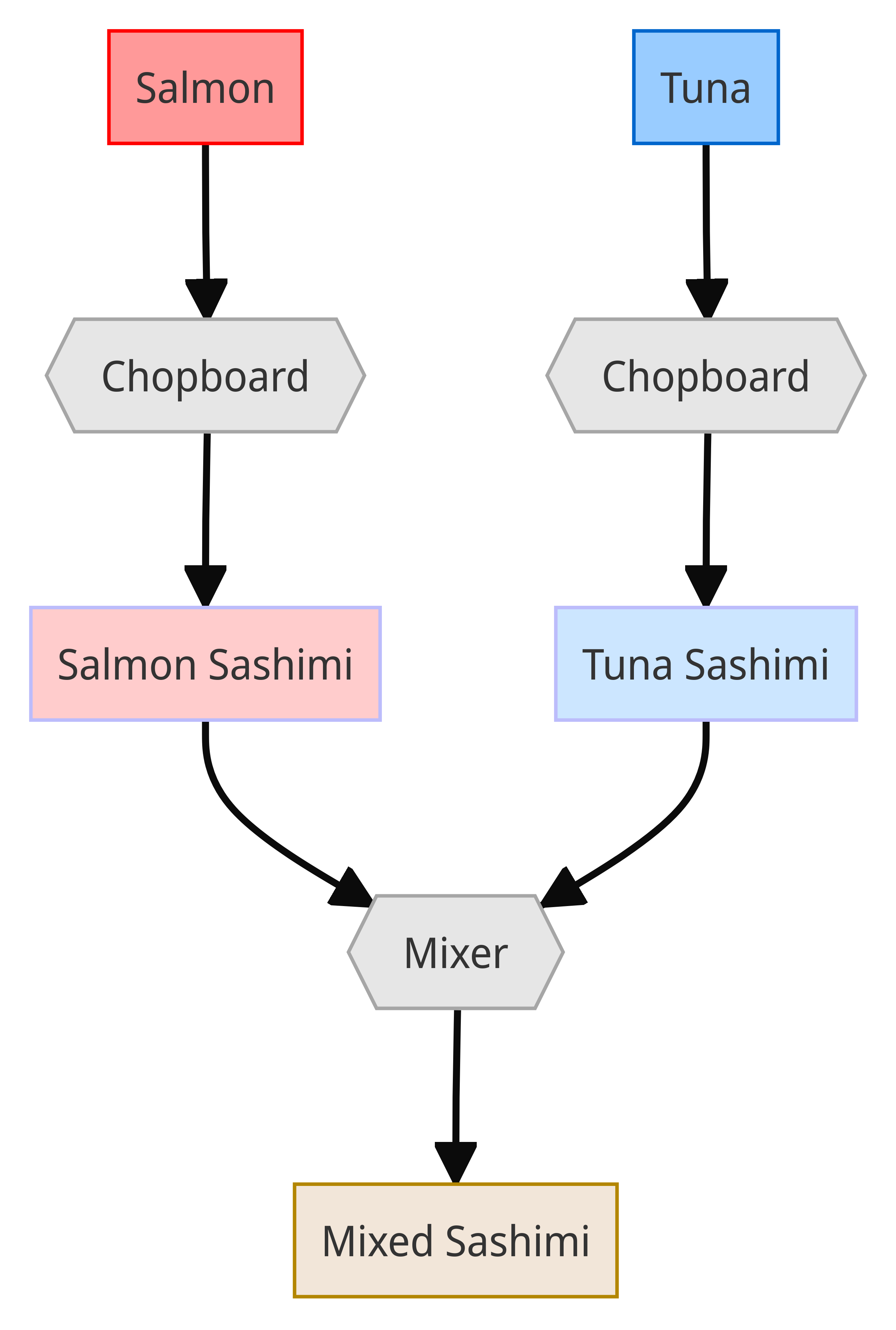}
        \caption{MixedSashimi}
    \end{subfigure}
\end{figure}

% Level 3
\subsection{Level 3}
\begin{figure}[H]
  
    \begin{subfigure}{0.45\textwidth}
        \includegraphics[width=\linewidth]{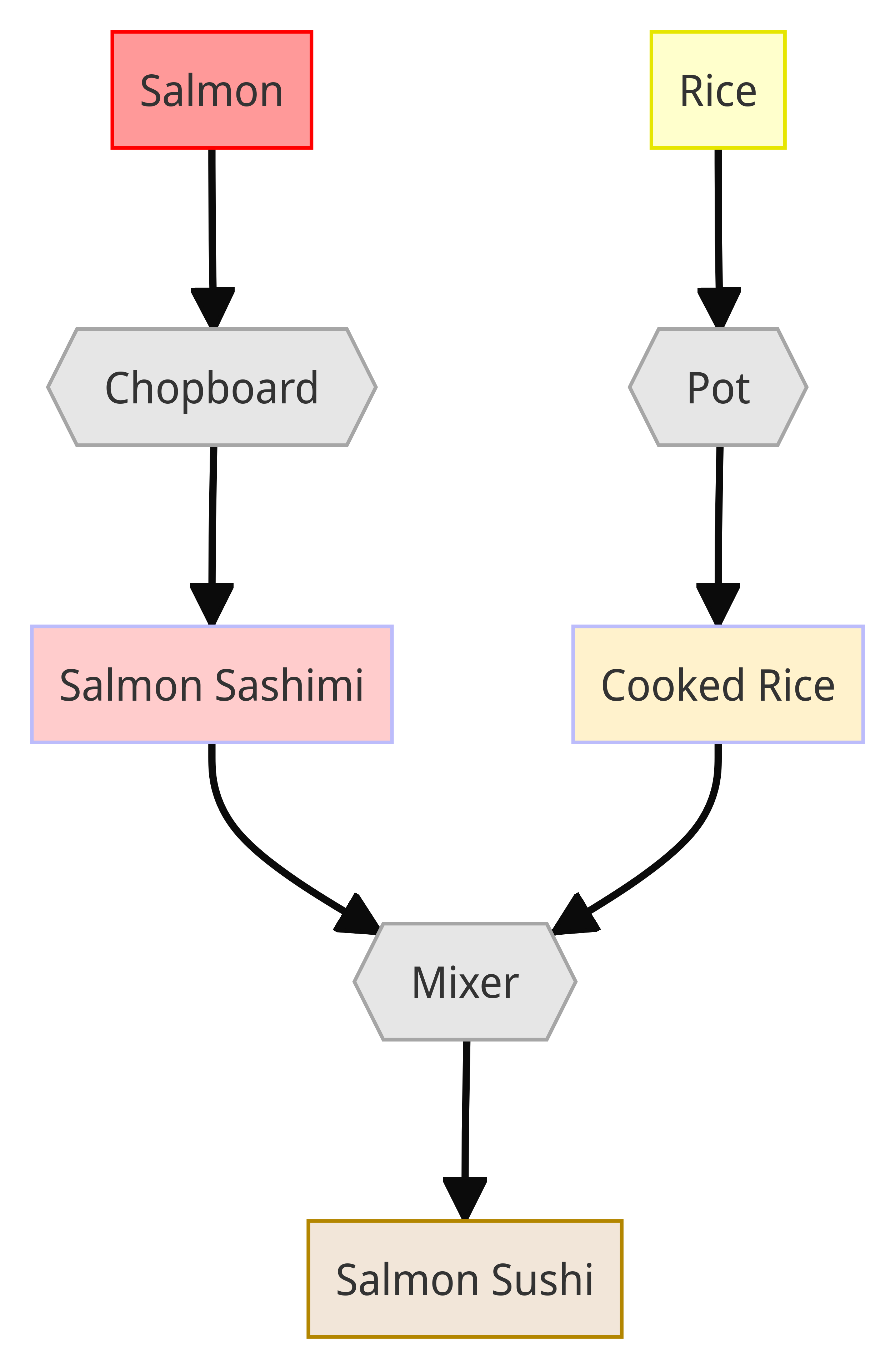}
        \caption{Salmon Sushi}
    \end{subfigure}
    \hfill
    \begin{subfigure}{0.45\textwidth}
        \includegraphics[width=\linewidth]{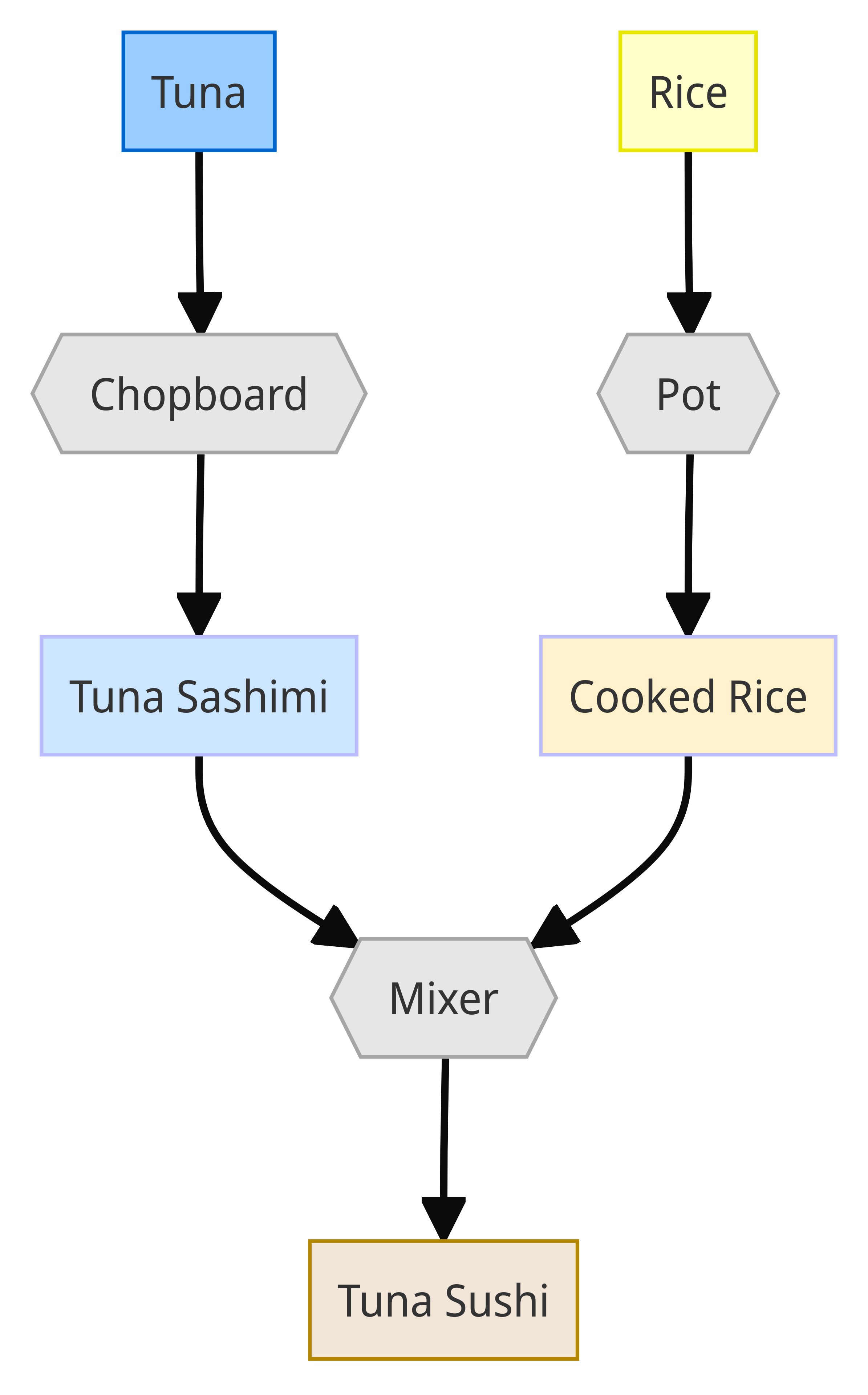}
        \caption{Tuna Sushi}
    \end{subfigure}
\end{figure}

% Level 4
\subsection{Level 4}
\begin{figure}[H]
    \begin{subfigure}{0.2\textwidth}
        \includegraphics[width=\linewidth]{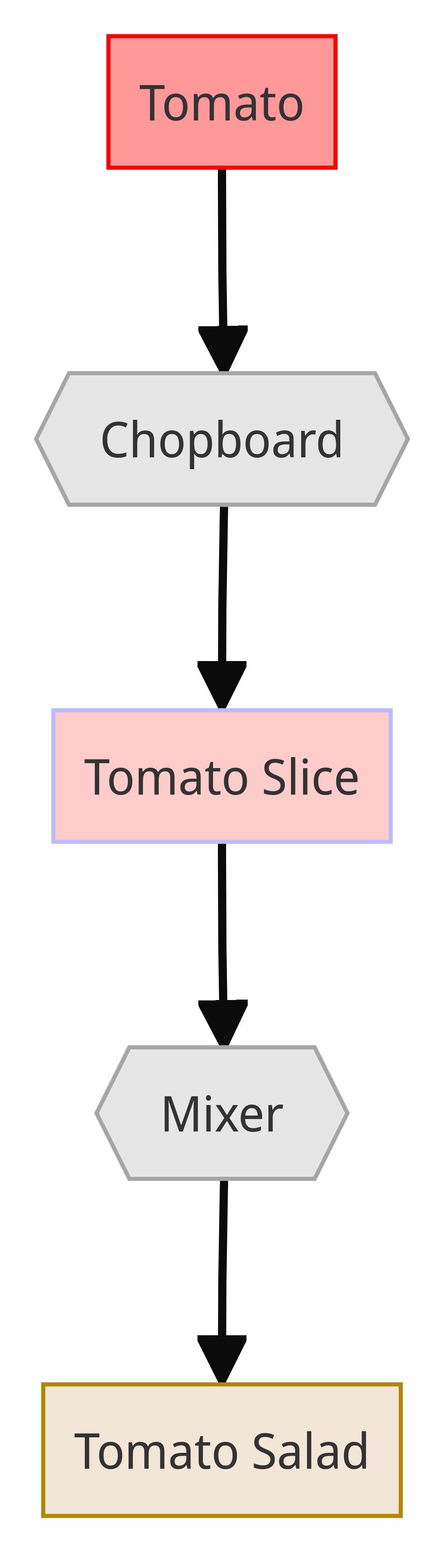}
        \caption{Tomato Salad}
    \end{subfigure}
    \hfill
    \begin{subfigure}{0.2\textwidth}
        \includegraphics[width=\linewidth]{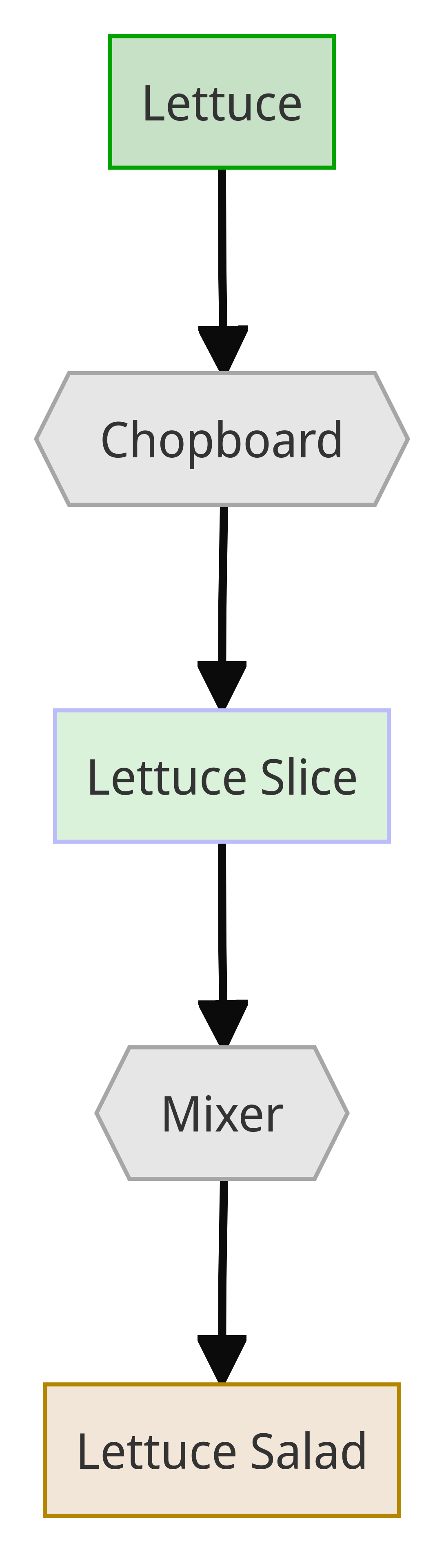}
        \caption{Lettuce Salad}
    \end{subfigure}
    \hfill
    \begin{subfigure}{0.23\textwidth}
        \includegraphics[width=\linewidth]{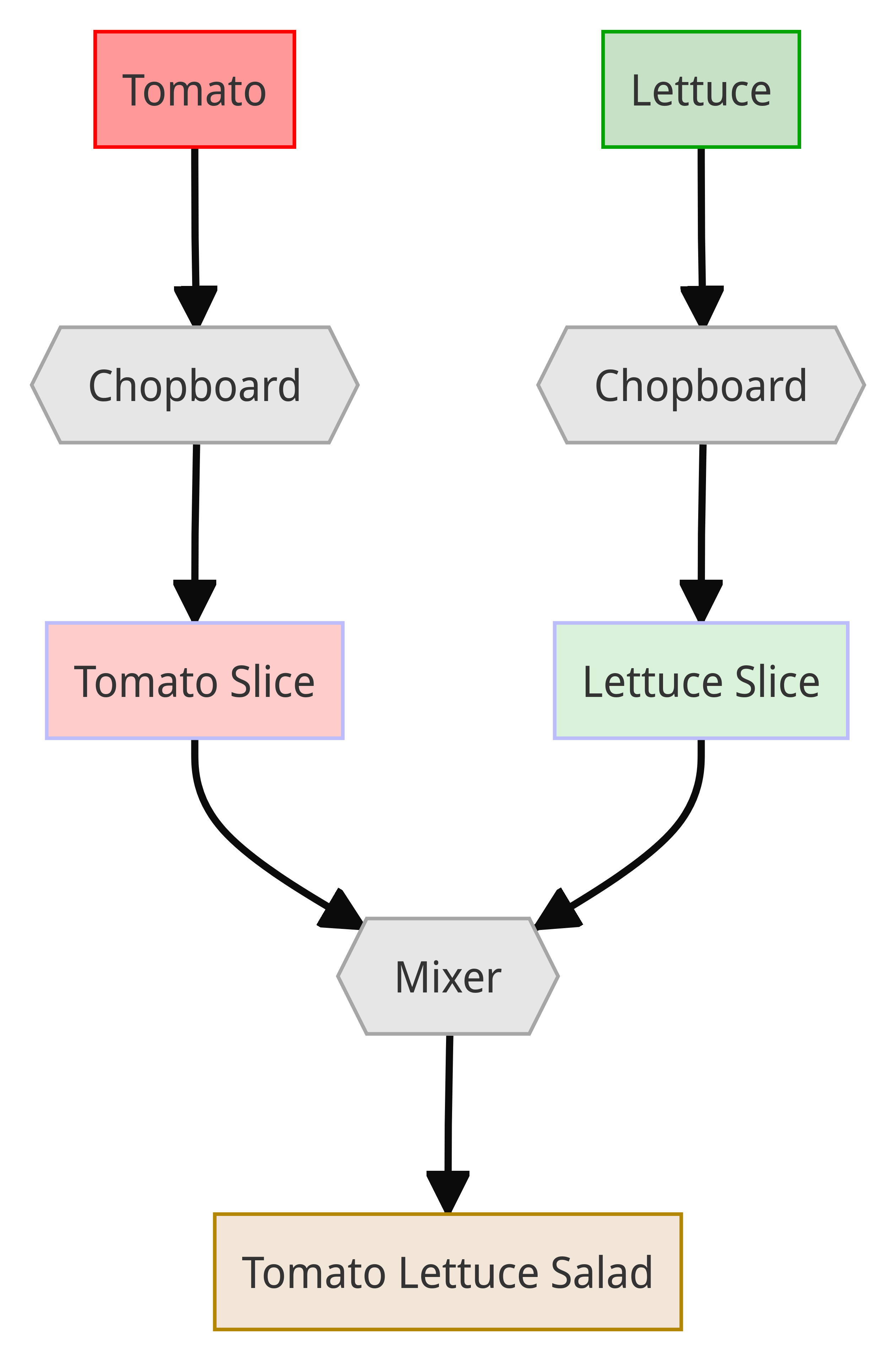}
        \caption{Tomato Lettuce Salad}
    \end{subfigure}
    \hfill
    \begin{subfigure}{0.23\textwidth}
        \includegraphics[width=\linewidth]{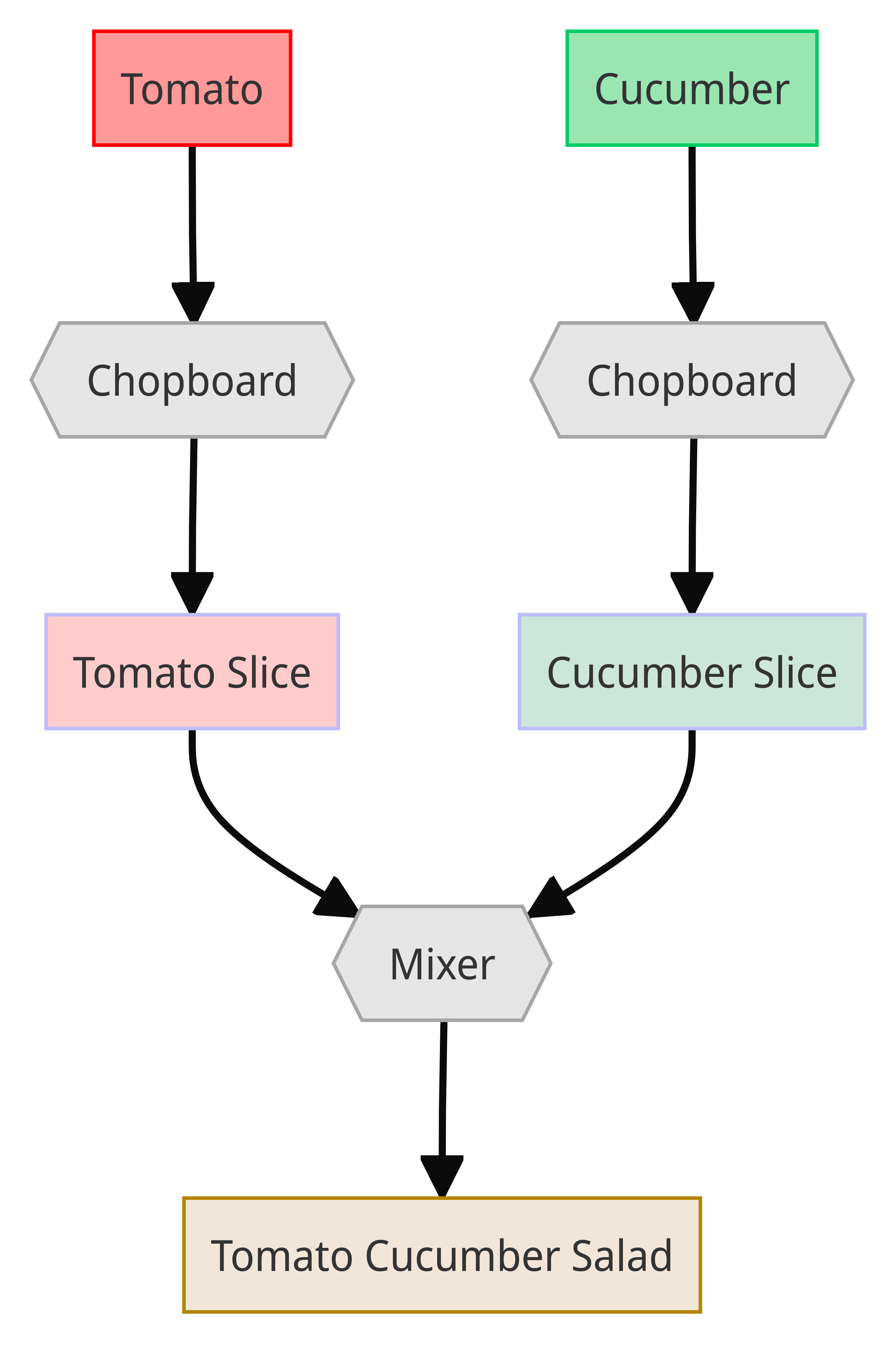}
        \caption{Tomato Cucumber Salad}
    \end{subfigure}
\end{figure}

% Level 5
\subsection{Level 5}
\begin{figure}[H]
    \begin{subfigure}{0.3\textwidth}
        \includegraphics[width=\linewidth]{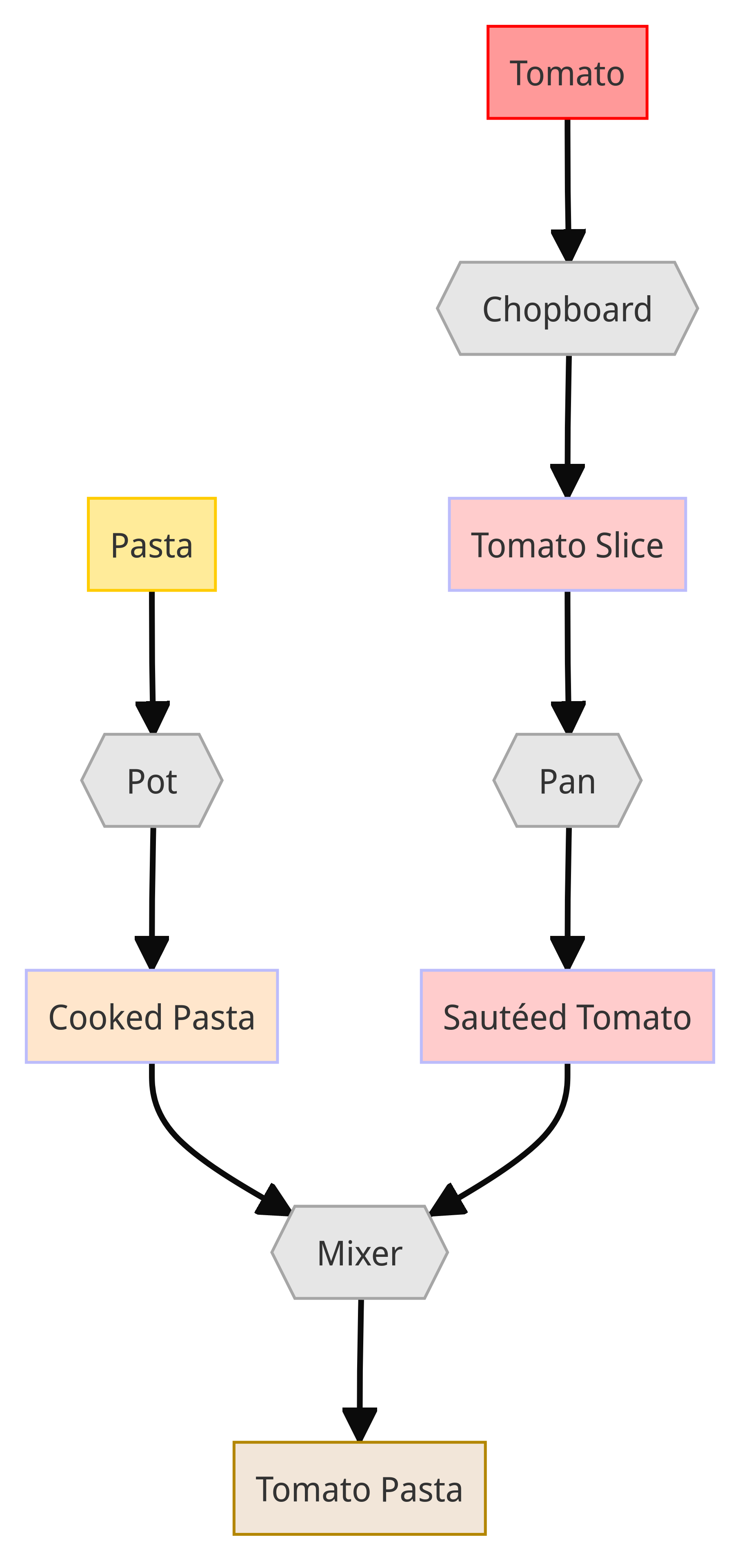}
        \caption{Tomato Pasta}
    \end{subfigure}
    \hfill
    \begin{subfigure}{0.3\textwidth}
        \includegraphics[width=\linewidth]{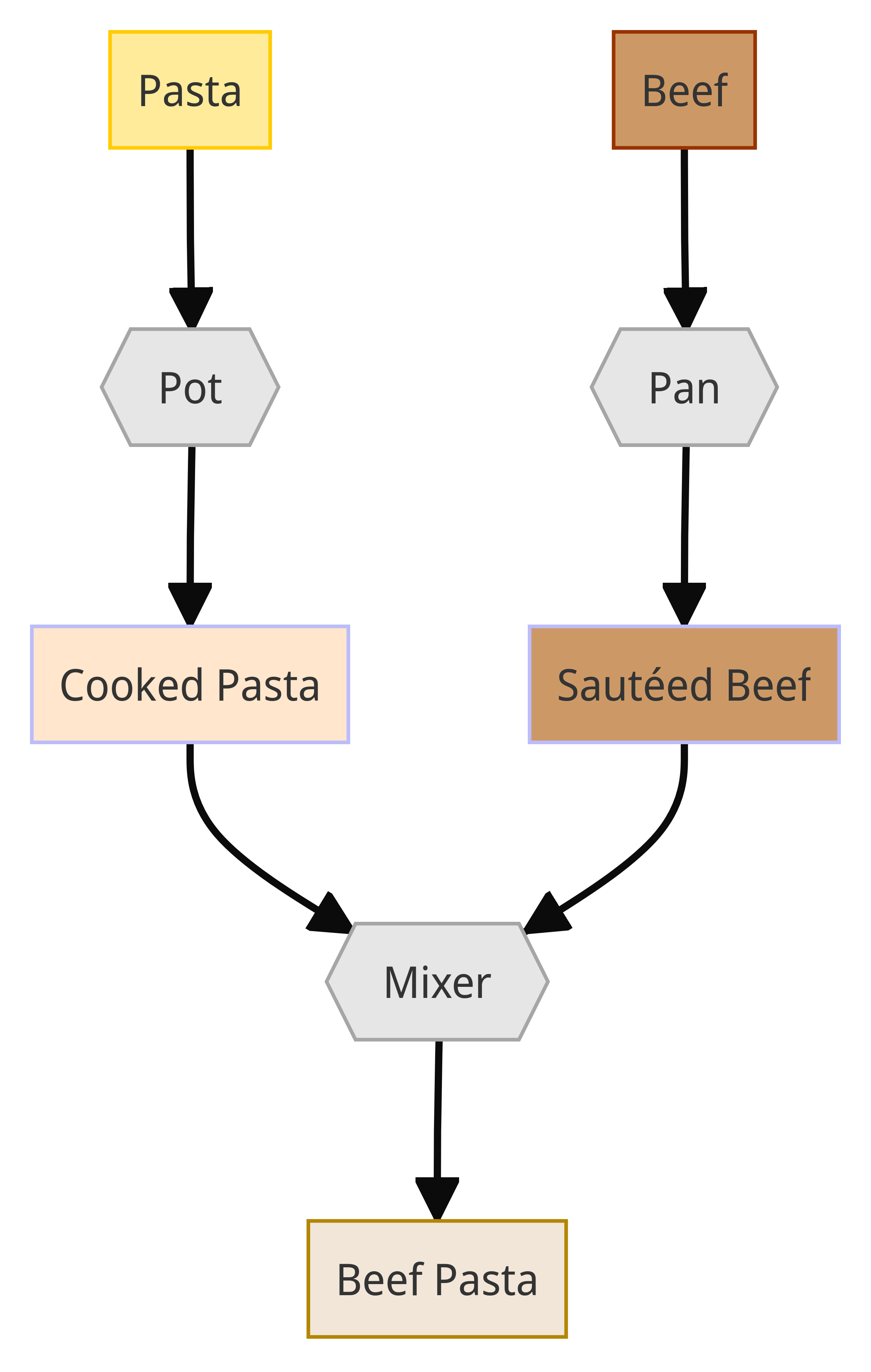}
        \caption{Beef Pasta}
    \end{subfigure}
    \hfill
    \begin{subfigure}{0.3\textwidth}
        \includegraphics[width=\linewidth]{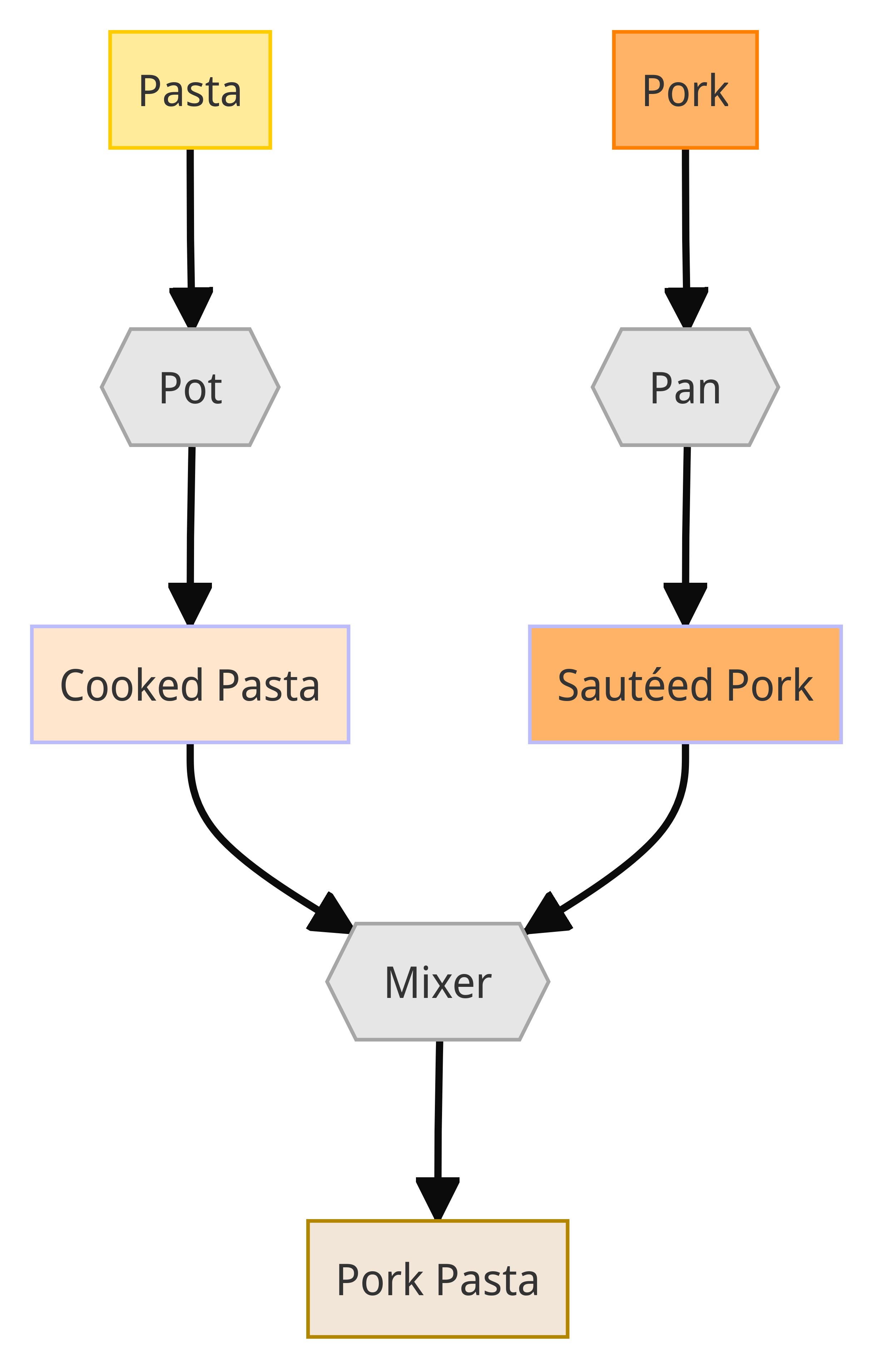}
        \caption{Pork Pasta}
    \end{subfigure}
\end{figure}

% Level 6
\subsection{Level 6}
\begin{figure}[H]
    \begin{subfigure}{0.32\textwidth}
        \includegraphics[width=\linewidth]{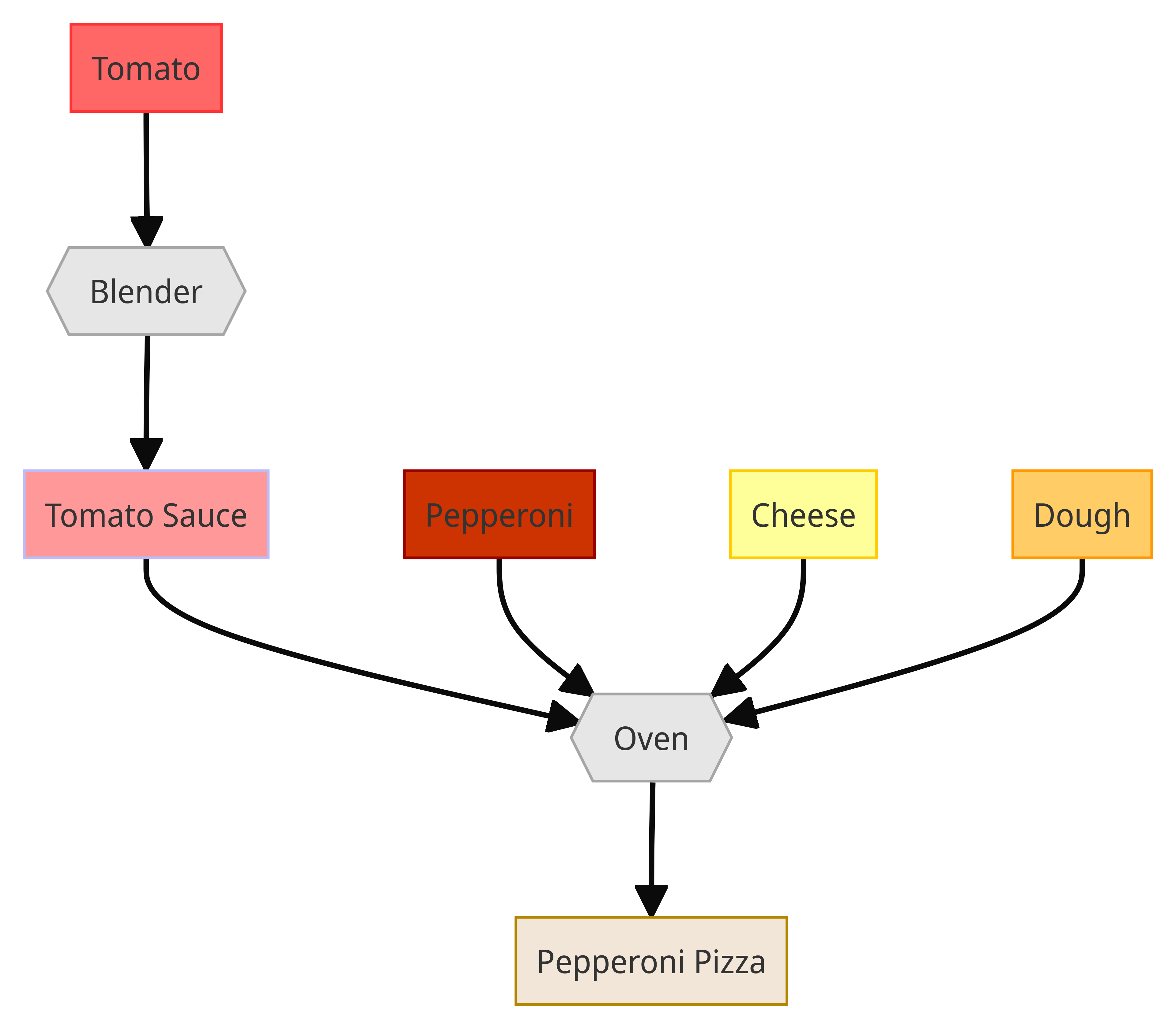}
        \caption{pepperoniPizza}
    \end{subfigure}
    \hfill
    \begin{subfigure}{0.32\textwidth}
        \includegraphics[width=\linewidth]{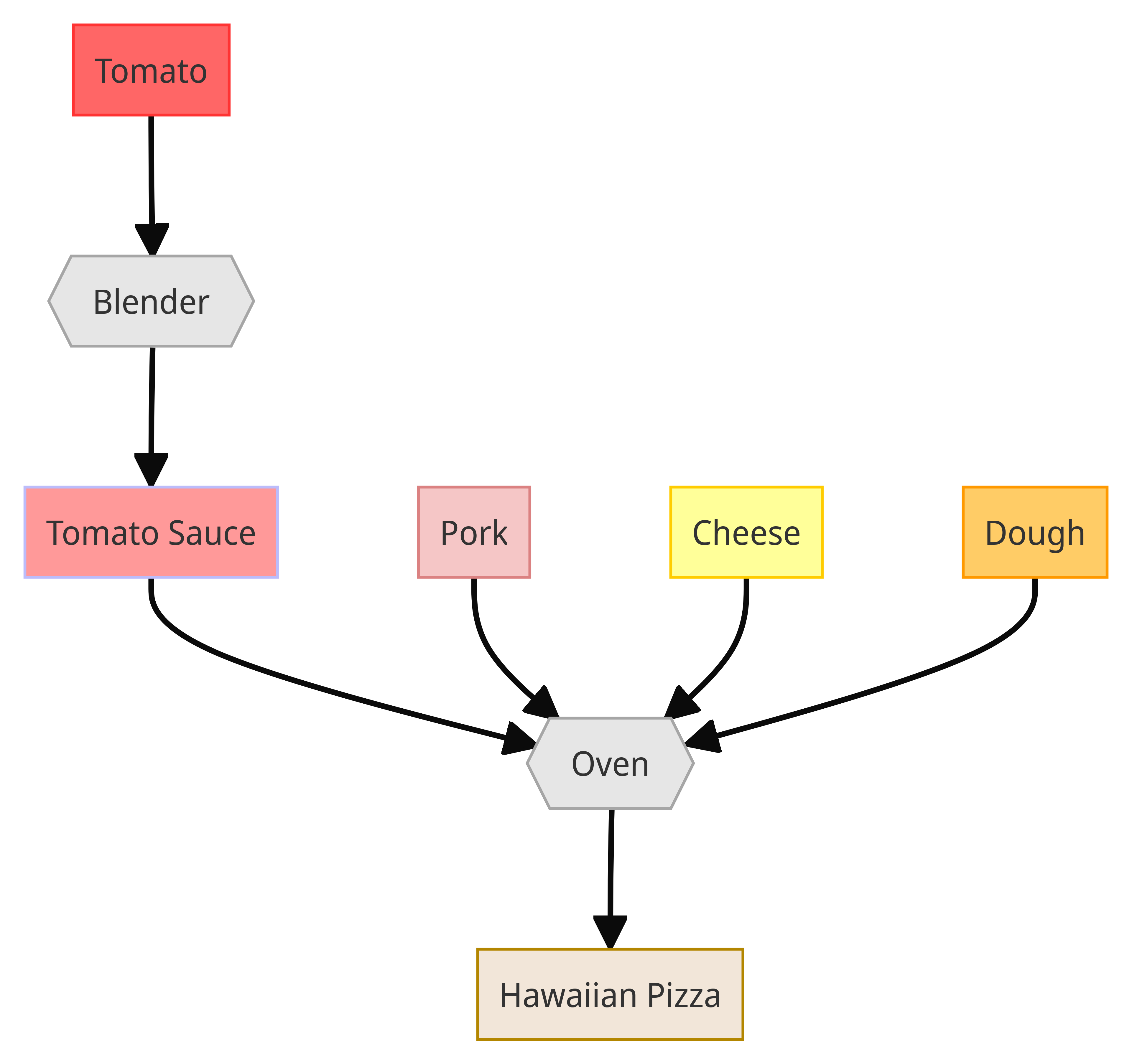}
        \caption{hawaiianPizza}
    \end{subfigure}
    \hfill
    \begin{subfigure}{0.32\textwidth}
        \includegraphics[width=\linewidth]{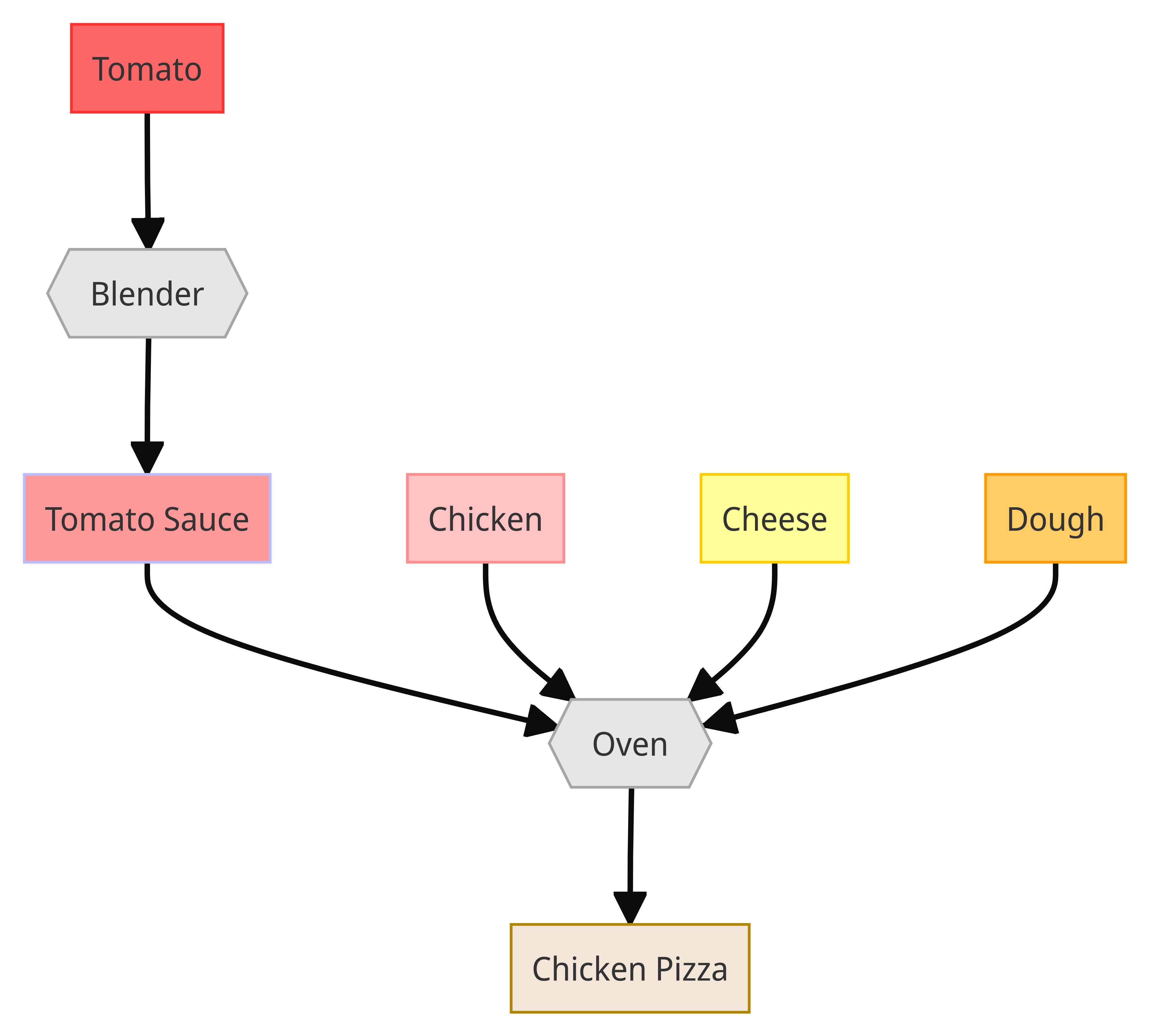}
        \caption{chickenPizza}
    \end{subfigure}
\end{figure}

% Level 7
\subsection{Level 7}
\begin{figure}[H]
    \begin{subfigure}{0.3\textwidth}
        \includegraphics[width=\linewidth]{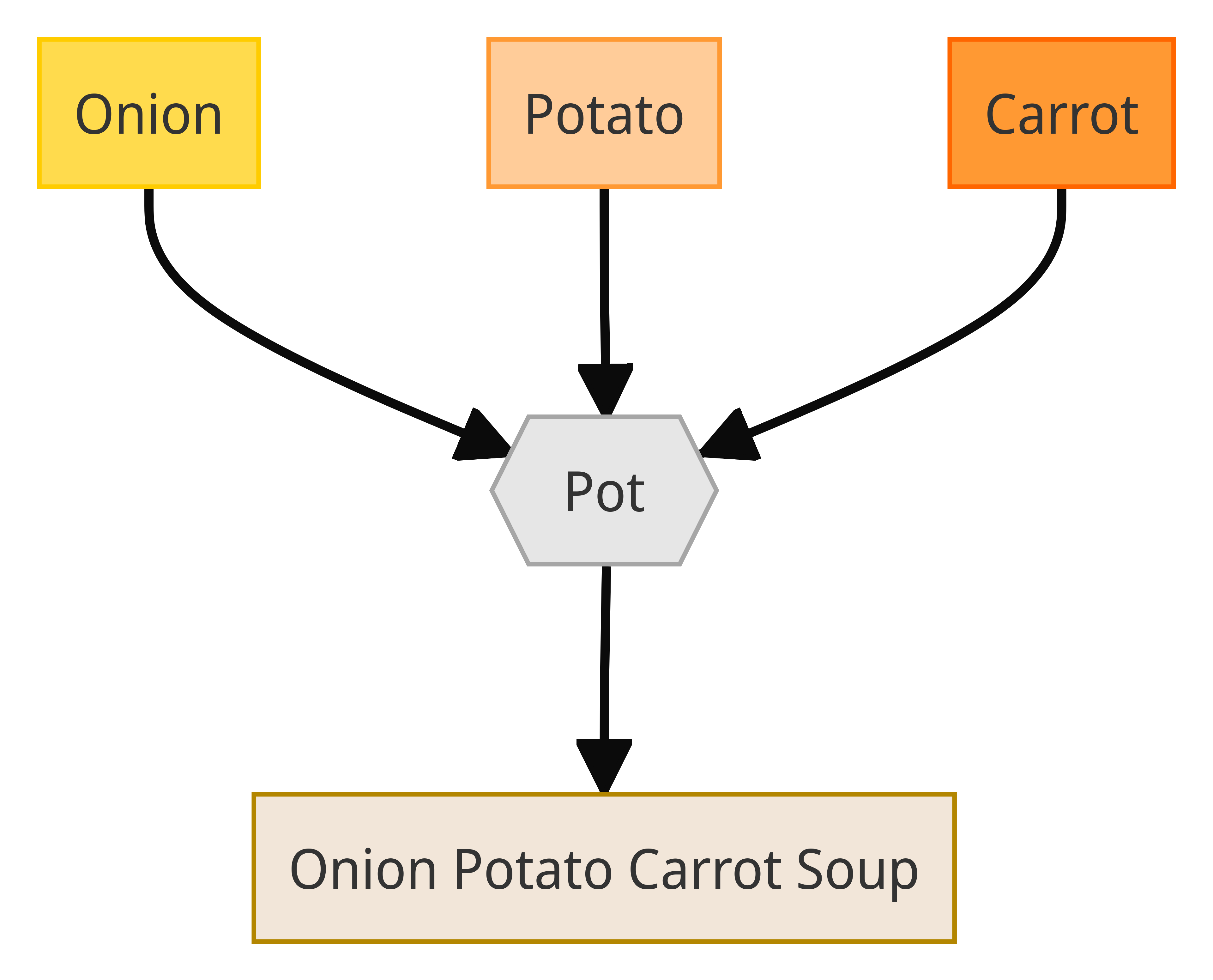}
        \caption{onionPotatoCarrotSoup}
    \end{subfigure}
    \hfill
    \begin{subfigure}{0.3\textwidth}
        \includegraphics[width=\linewidth]{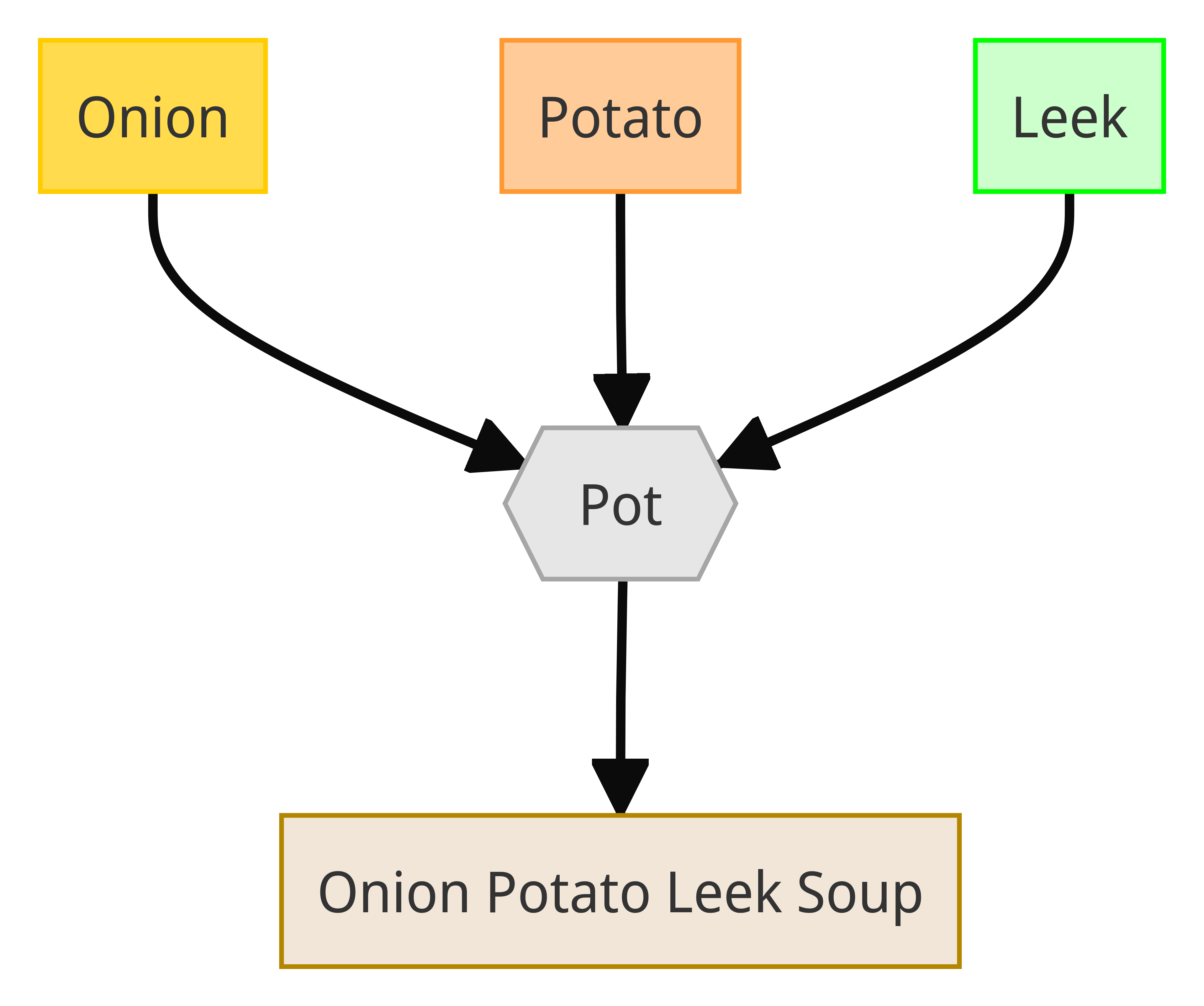}
        \caption{onionPotatoLeekSoup}
    \end{subfigure}
    \hfill
    \begin{subfigure}{0.3\textwidth}
        \includegraphics[width=\linewidth]{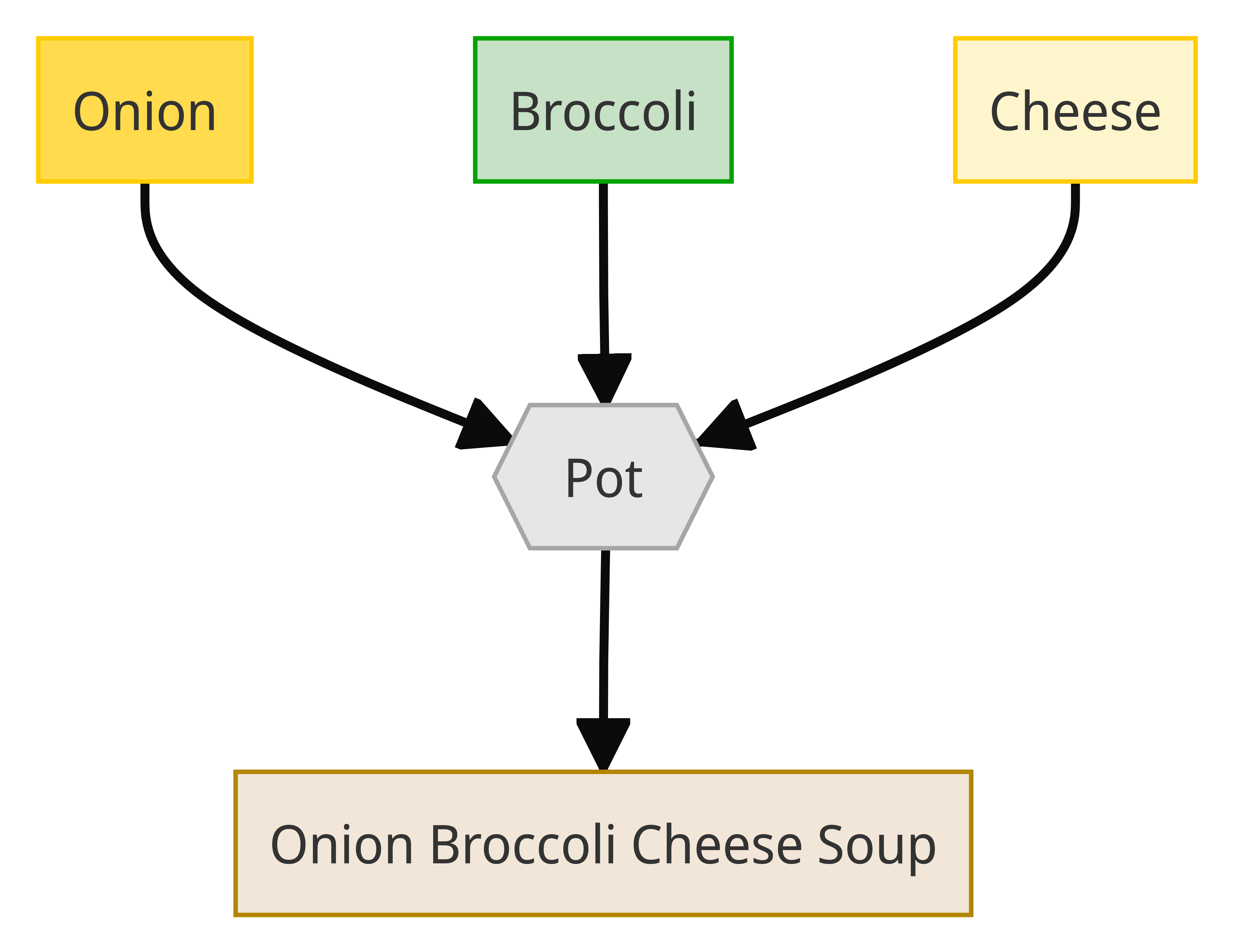}
        \caption{onionBroccoliCheeseSoup}
    \end{subfigure}
\end{figure}

% Level 8
\subsection{Level 8}
\begin{figure}[H]
    \begin{subfigure}{0.3\textwidth}
        \includegraphics[width=\linewidth]{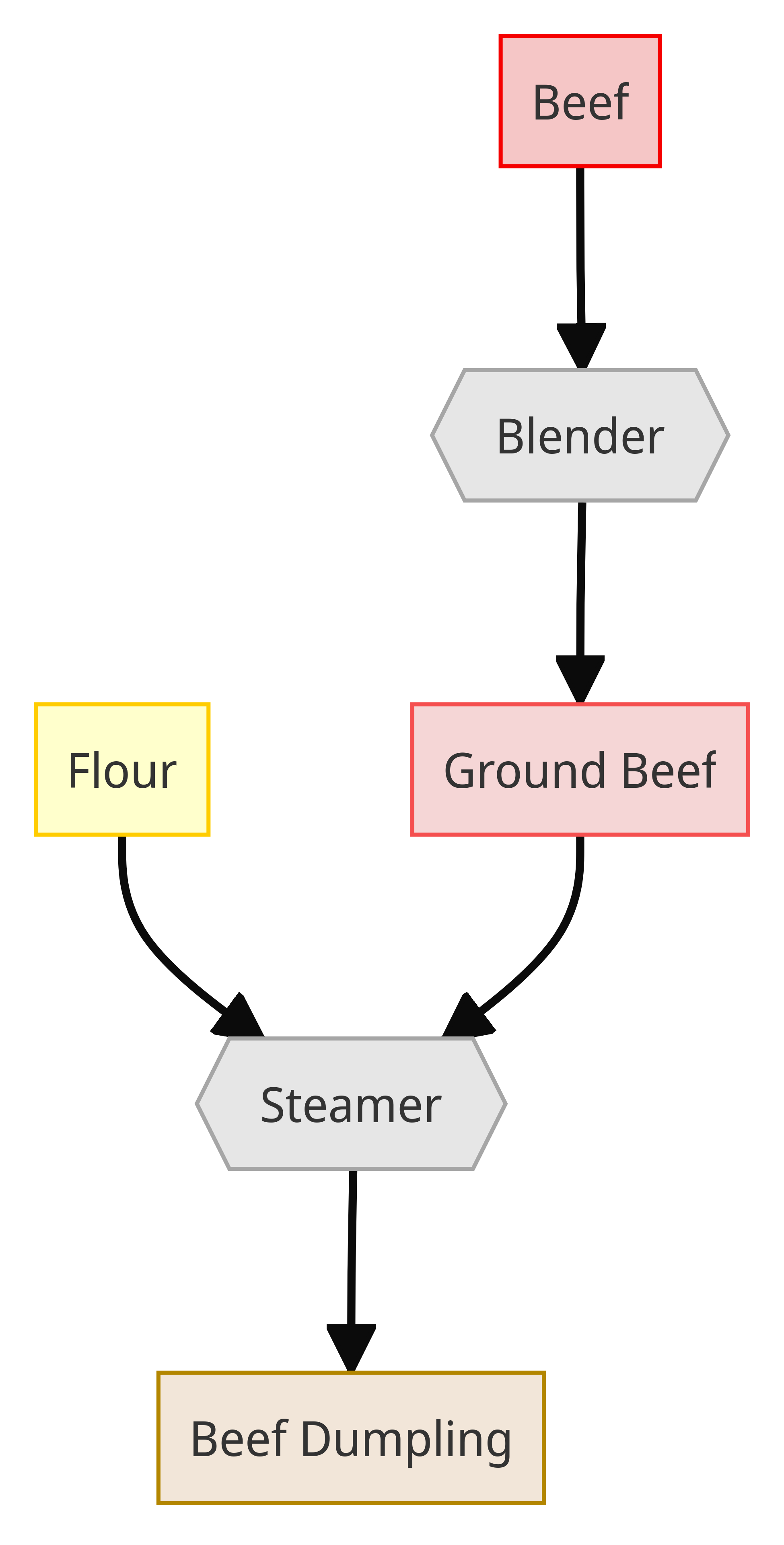}
        \caption{Beef Dumpling}
    \end{subfigure}
    \hfill
    \begin{subfigure}{0.3\textwidth}
        \includegraphics[width=\linewidth]{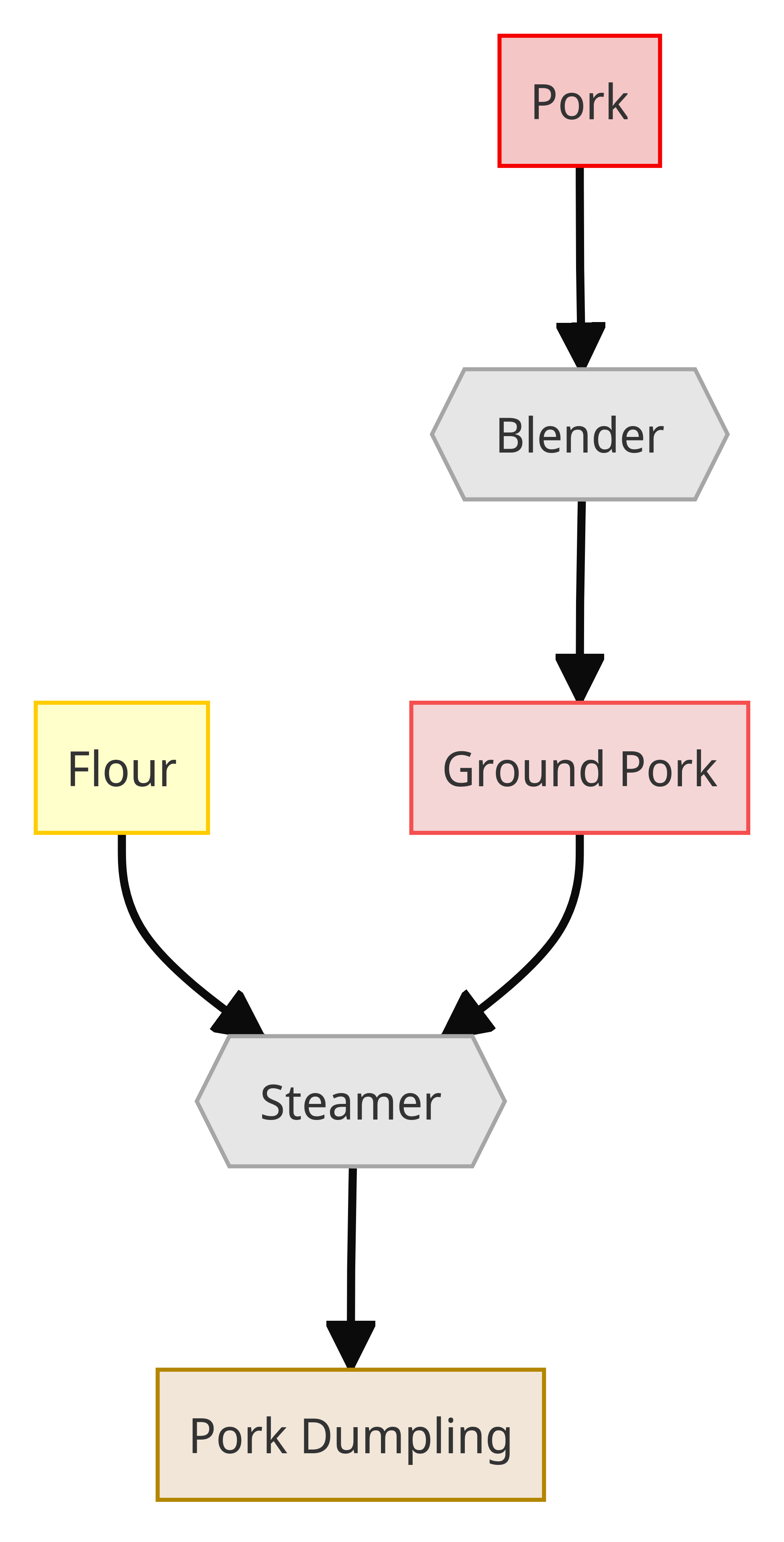}
        \caption{Pork Dumpling}
    \end{subfigure}
    \hfill
    \begin{subfigure}{0.3\textwidth}
        \includegraphics[width=\linewidth]{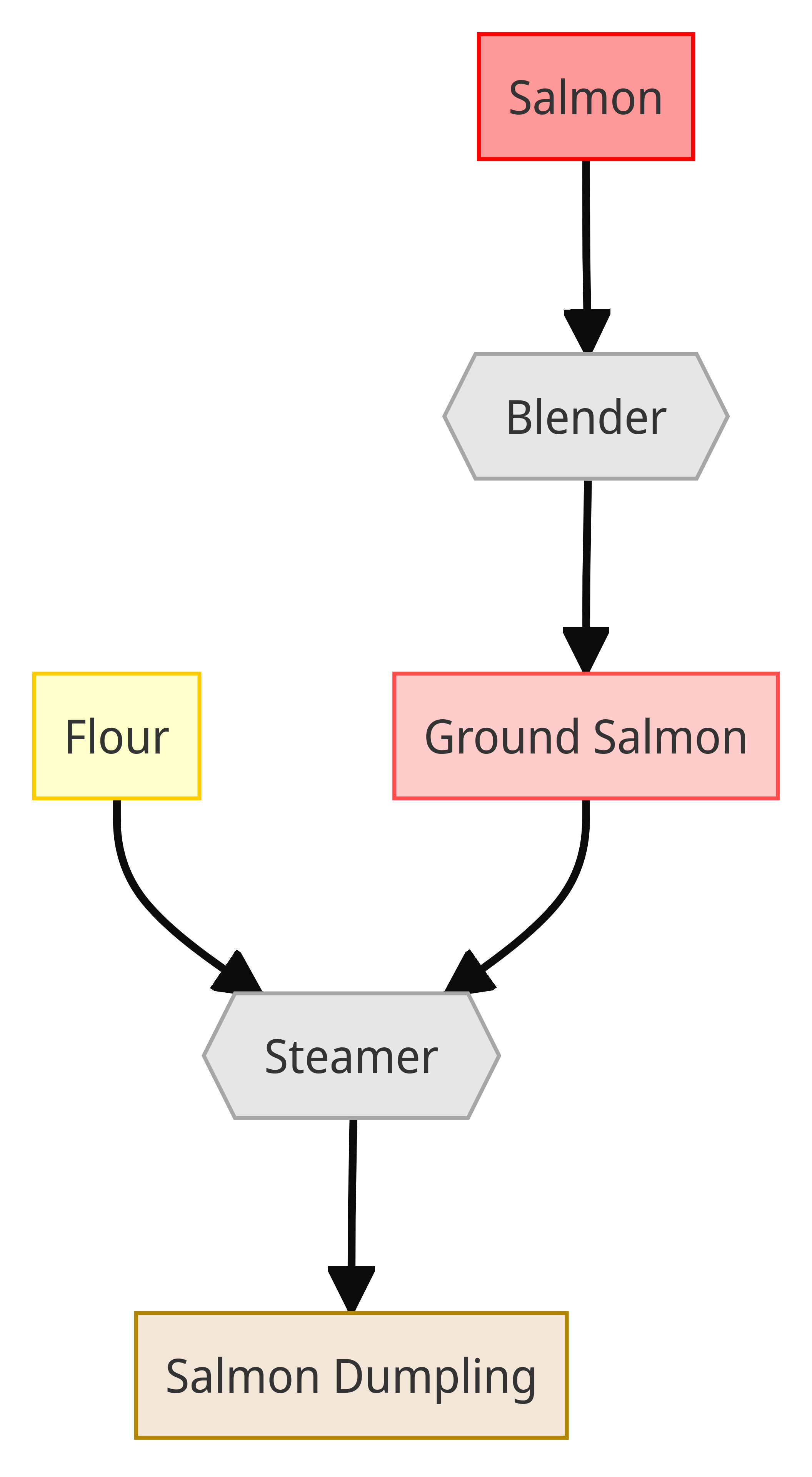}
        \caption{Salmon Dumpling}
    \end{subfigure}
\end{figure}

% Level 9
\subsection{Level 9}
\begin{figure}[H]
    \begin{subfigure}{0.3\textwidth}
        \includegraphics[width=\linewidth]{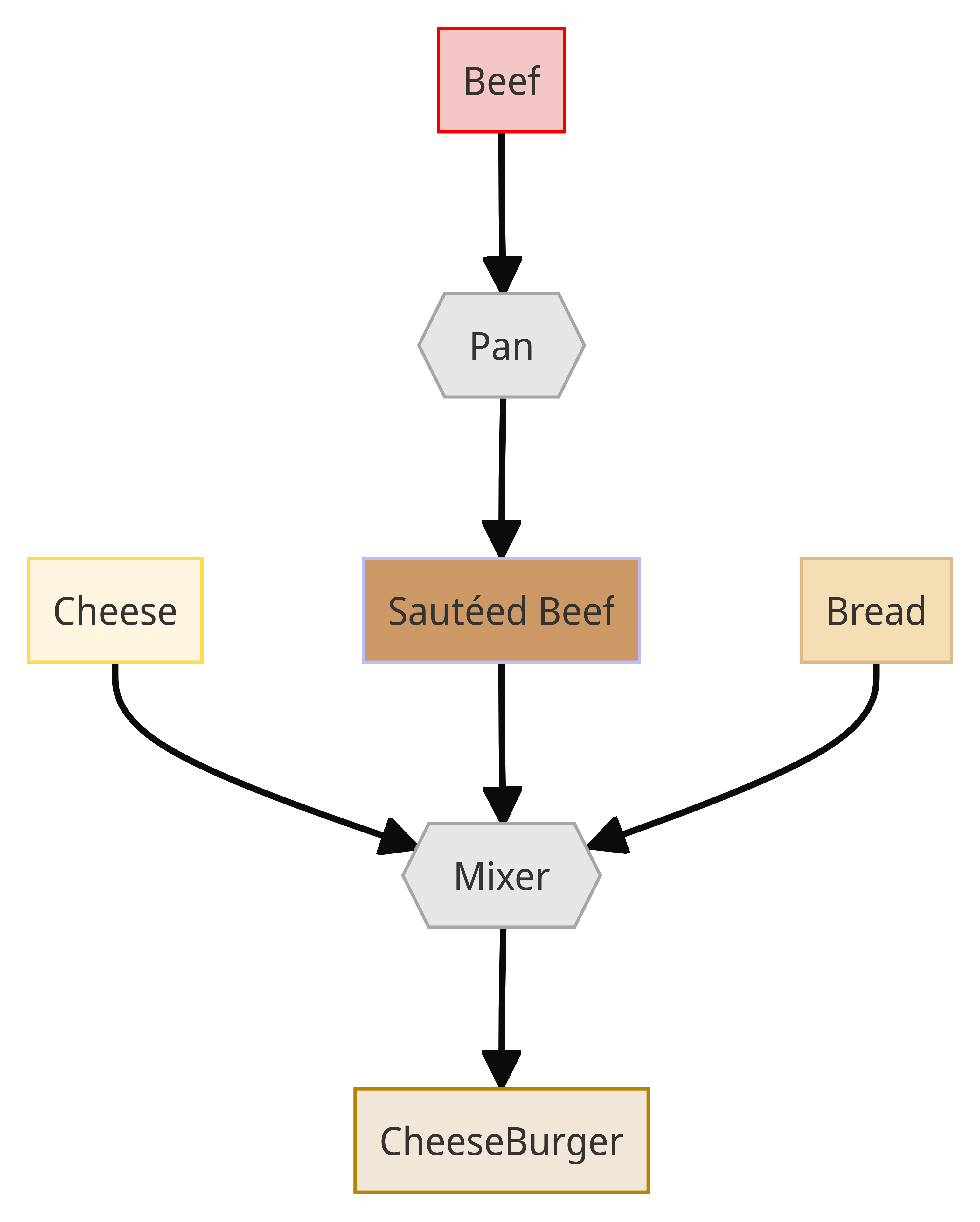}
        \caption{Cheese Burger}
    \end{subfigure}
    \hfill
    \begin{subfigure}{0.3\textwidth}
        \includegraphics[width=\linewidth]{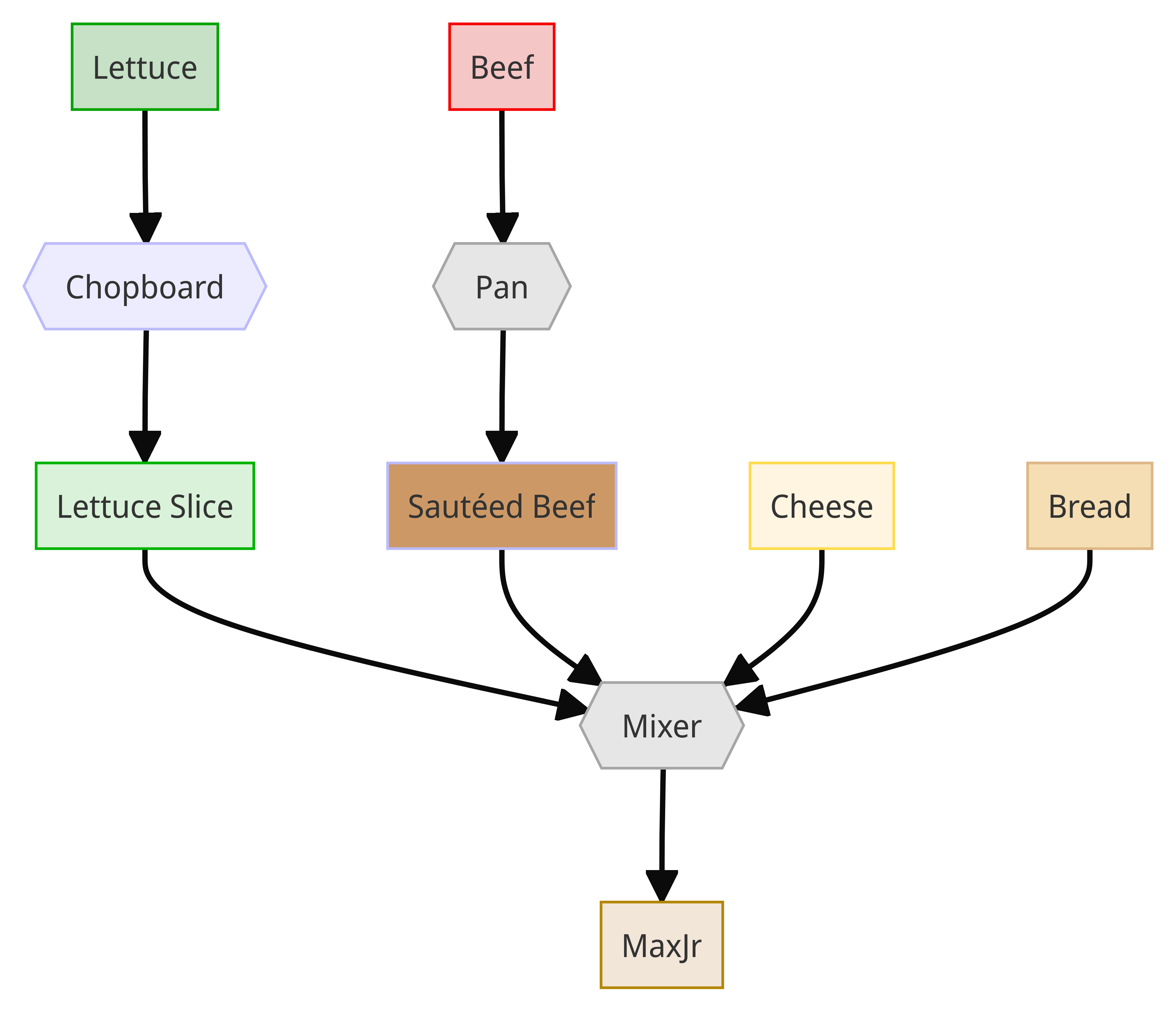}
        \caption{MaxJr}
    \end{subfigure}
    \hfill
    \begin{subfigure}{0.3\textwidth}
        \includegraphics[width=\linewidth]{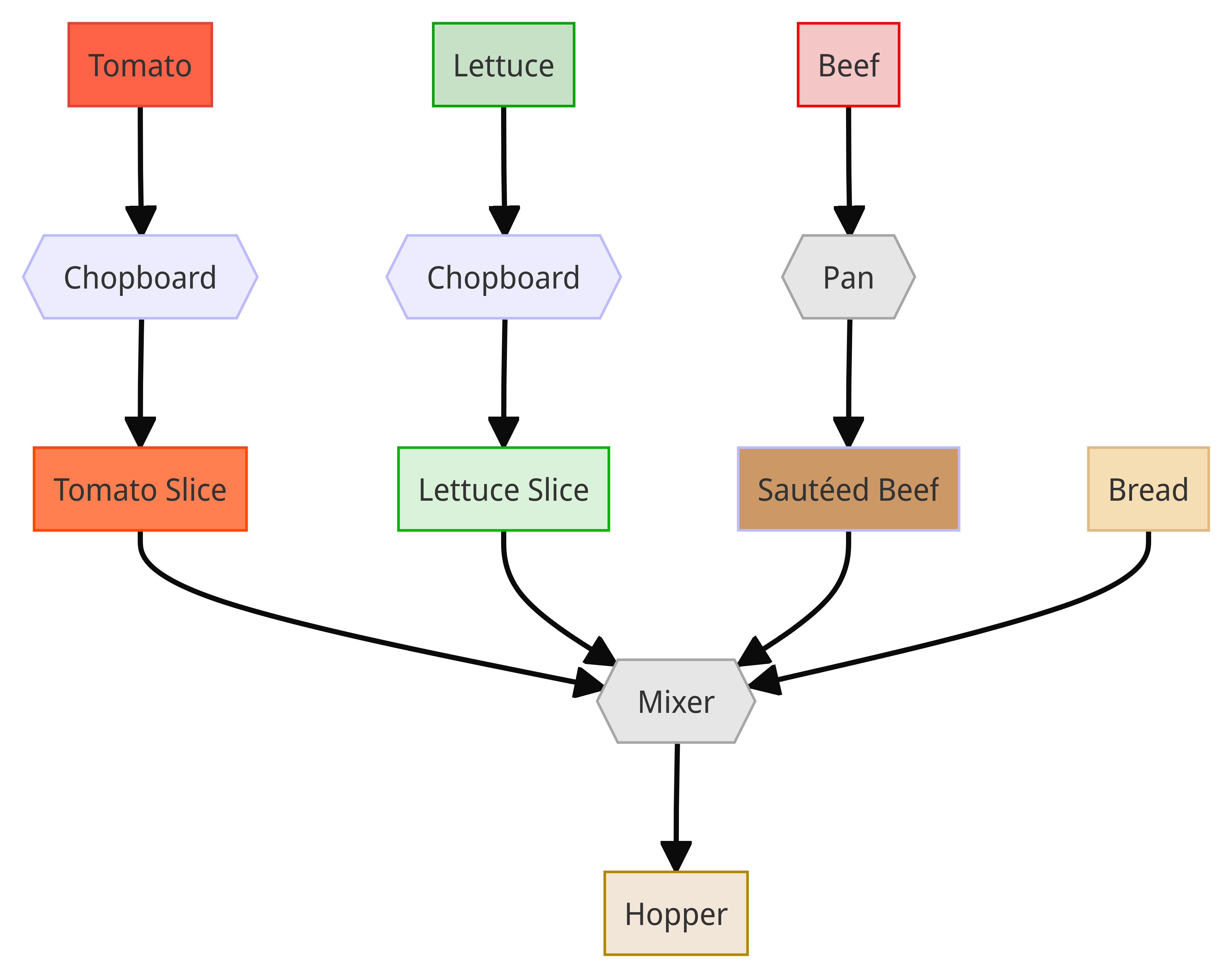}
        \caption{Hopper}
    \end{subfigure}
\end{figure}

% Level 10
\subsection{Level 10}
\begin{figure}[H]
    \begin{subfigure}{0.3\textwidth}
        \includegraphics[width=\linewidth]{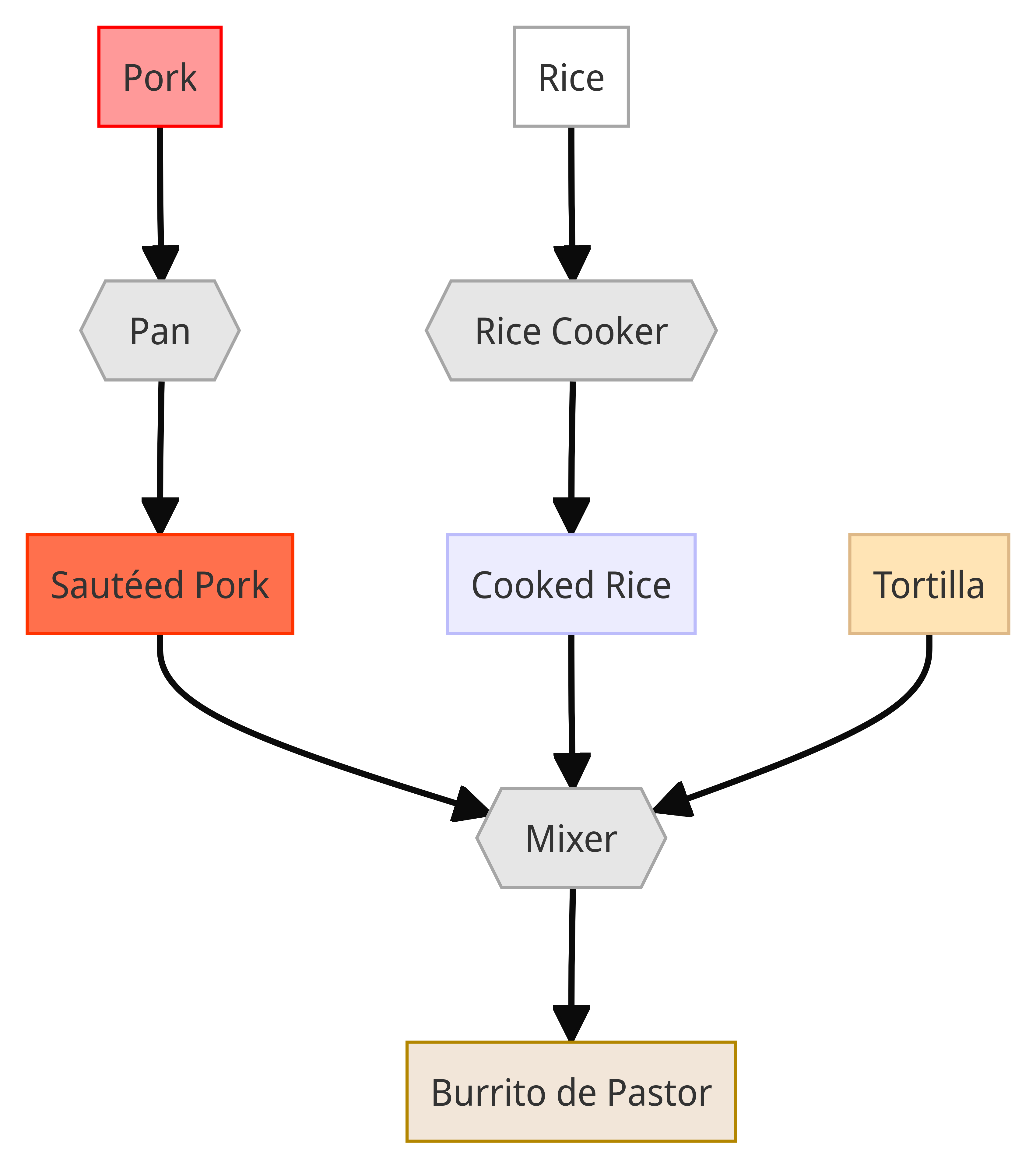}
        \caption{BurritodePastor}
    \end{subfigure}
    \hfill
    \begin{subfigure}{0.3\textwidth}
        \includegraphics[width=\linewidth]{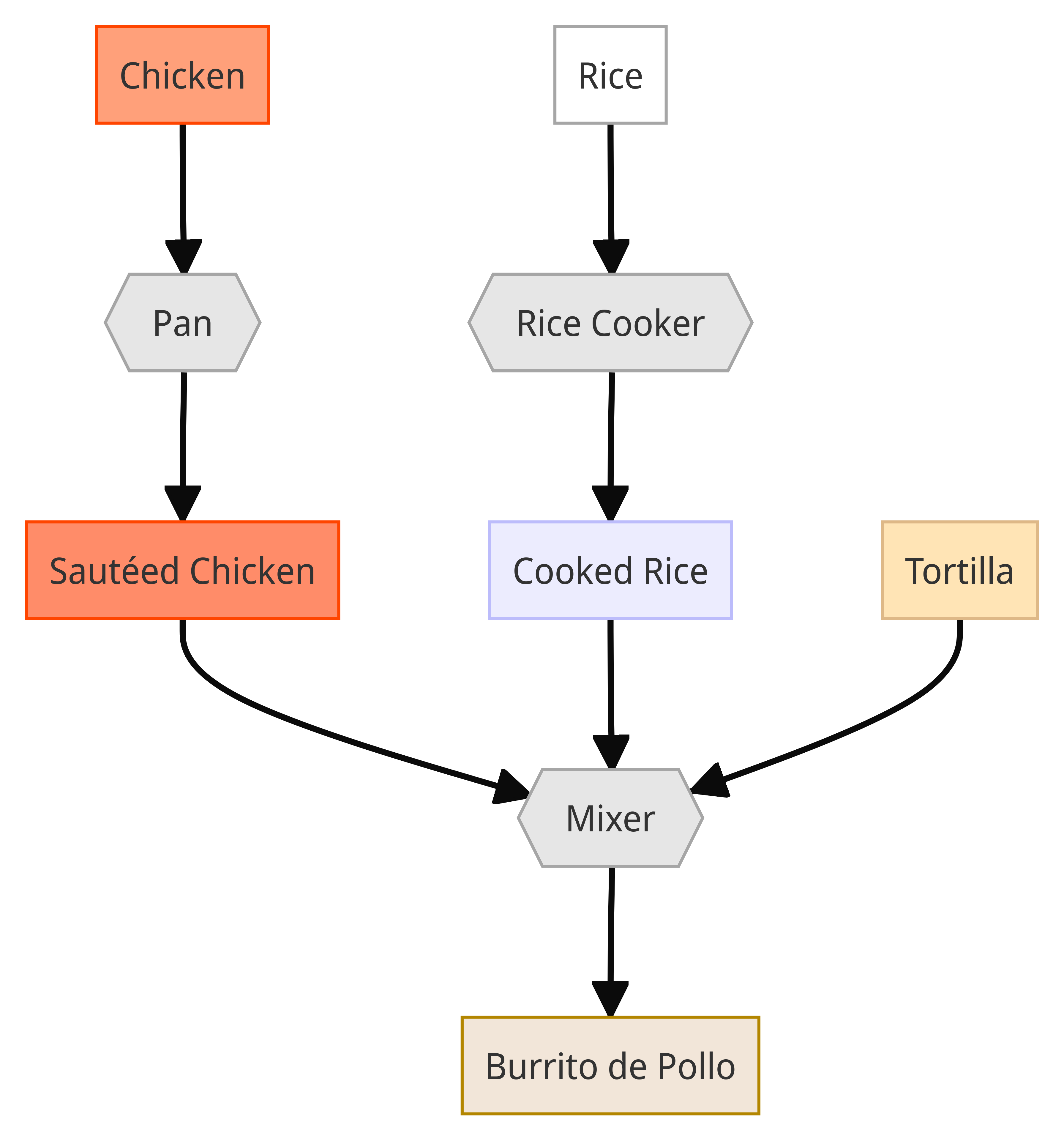}
        \caption{BurritodePollo}
    \end{subfigure}
    \hfill
    \begin{subfigure}{0.3\textwidth}
        \includegraphics[width=\linewidth]{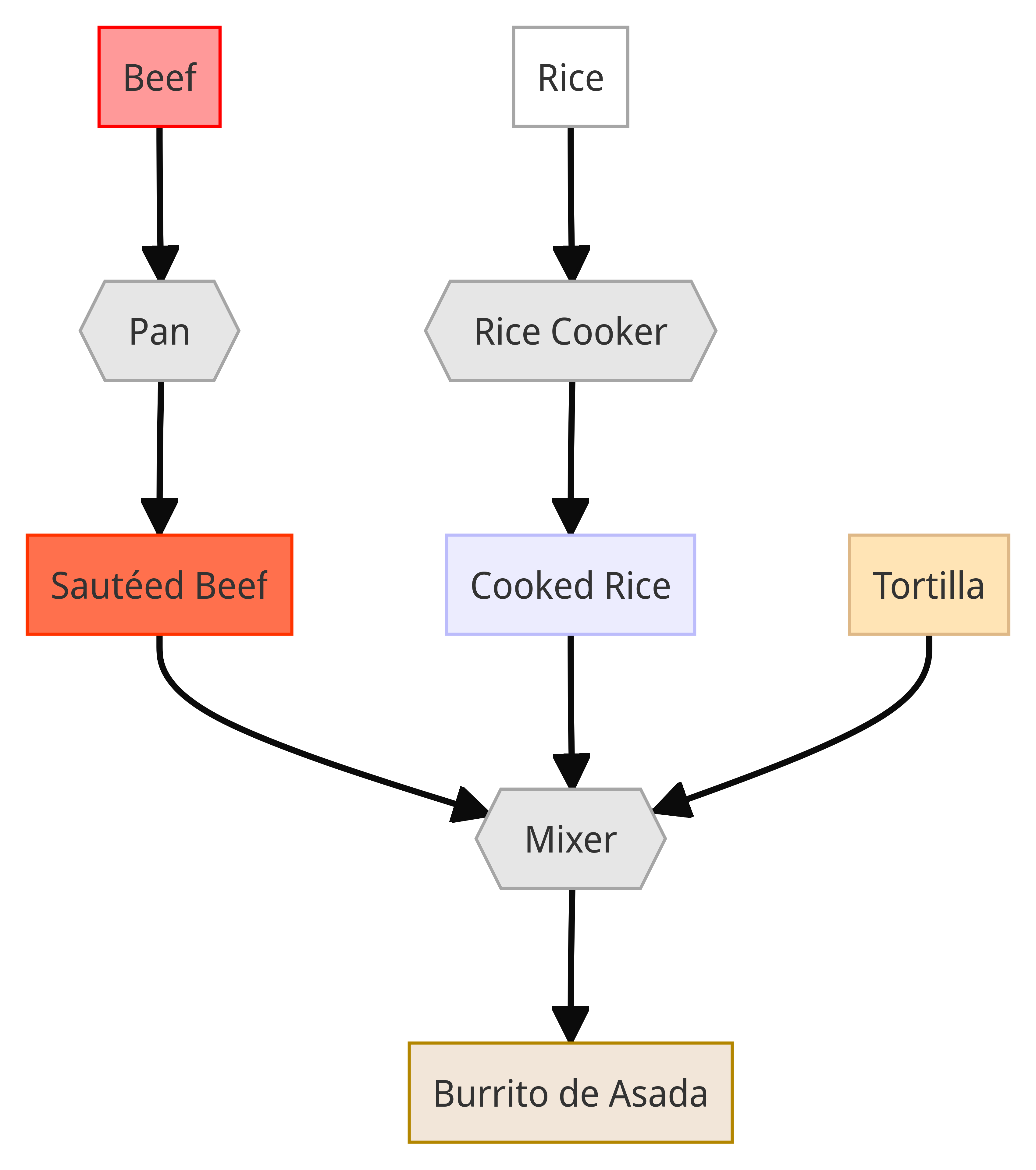}
        \caption{BurritodeAsada}
    \end{subfigure}
\end{figure}

% Level 11
\subsection{Level 11}
\begin{figure}[H]
    \begin{subfigure}{0.32\textwidth}
        \includegraphics[width=\linewidth]{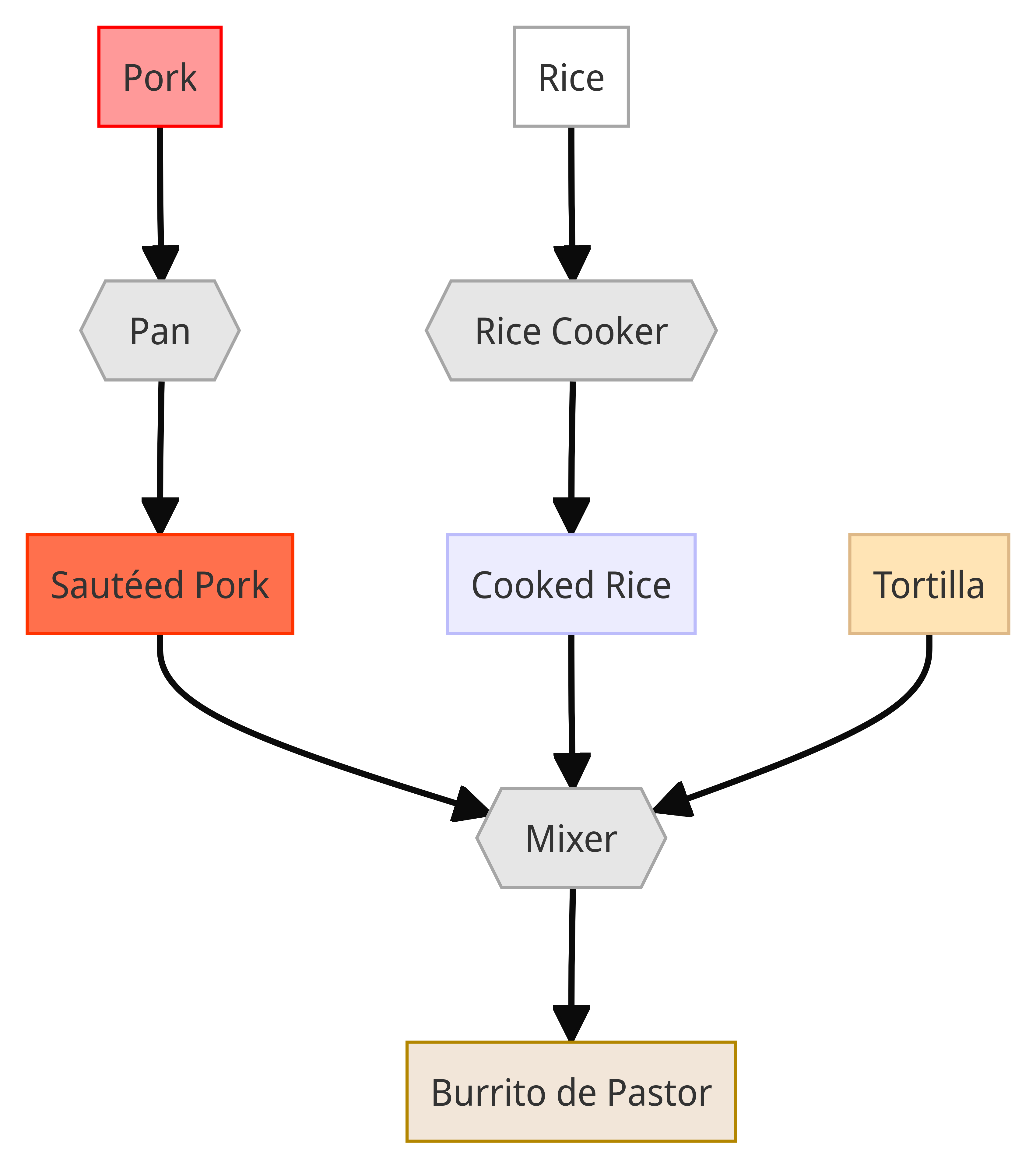}
        \caption{BurritodePastor}
    \end{subfigure}
    \hfill
    \begin{subfigure}{0.32\textwidth}
        \includegraphics[width=\linewidth]{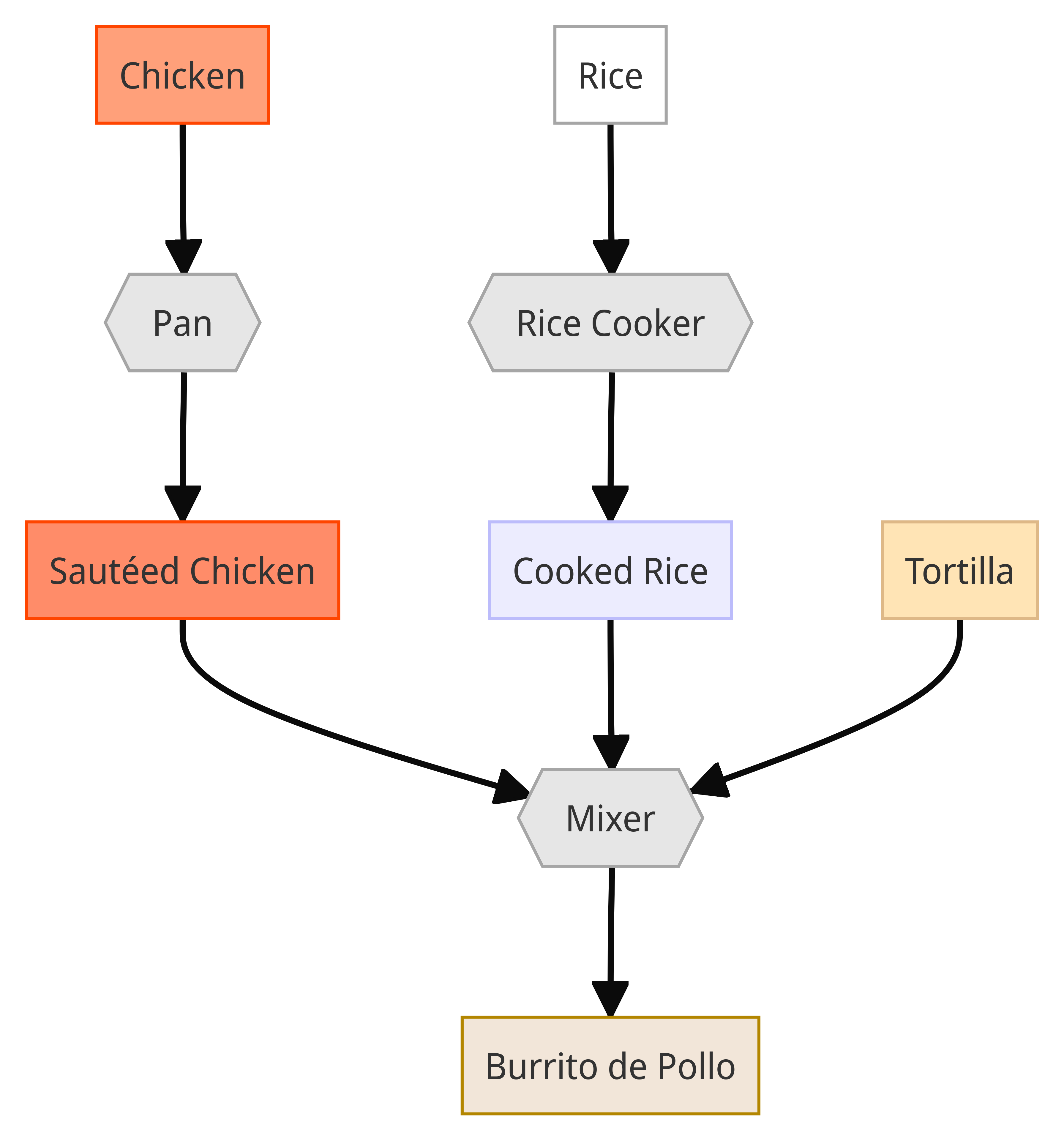}
        \caption{BurritodePollo}
    \end{subfigure}
    \hfill
    \begin{subfigure}{0.32\textwidth}
        \includegraphics[width=\linewidth]{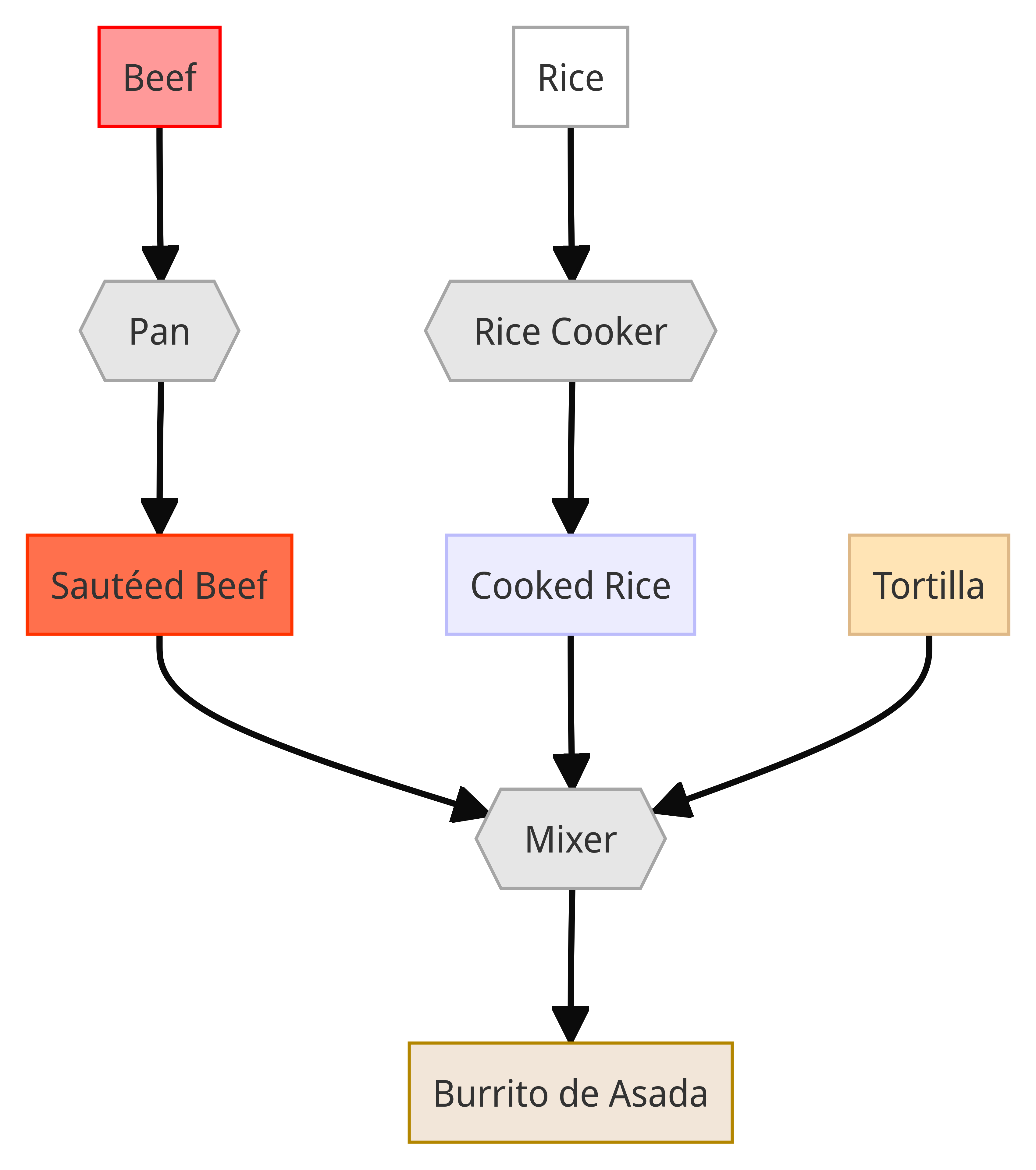}
        \caption{BurritodeAsada}
    \end{subfigure}
    \hfill
    \begin{subfigure}{0.3\textwidth}
        \includegraphics[width=\linewidth, height=0.23\textheight]{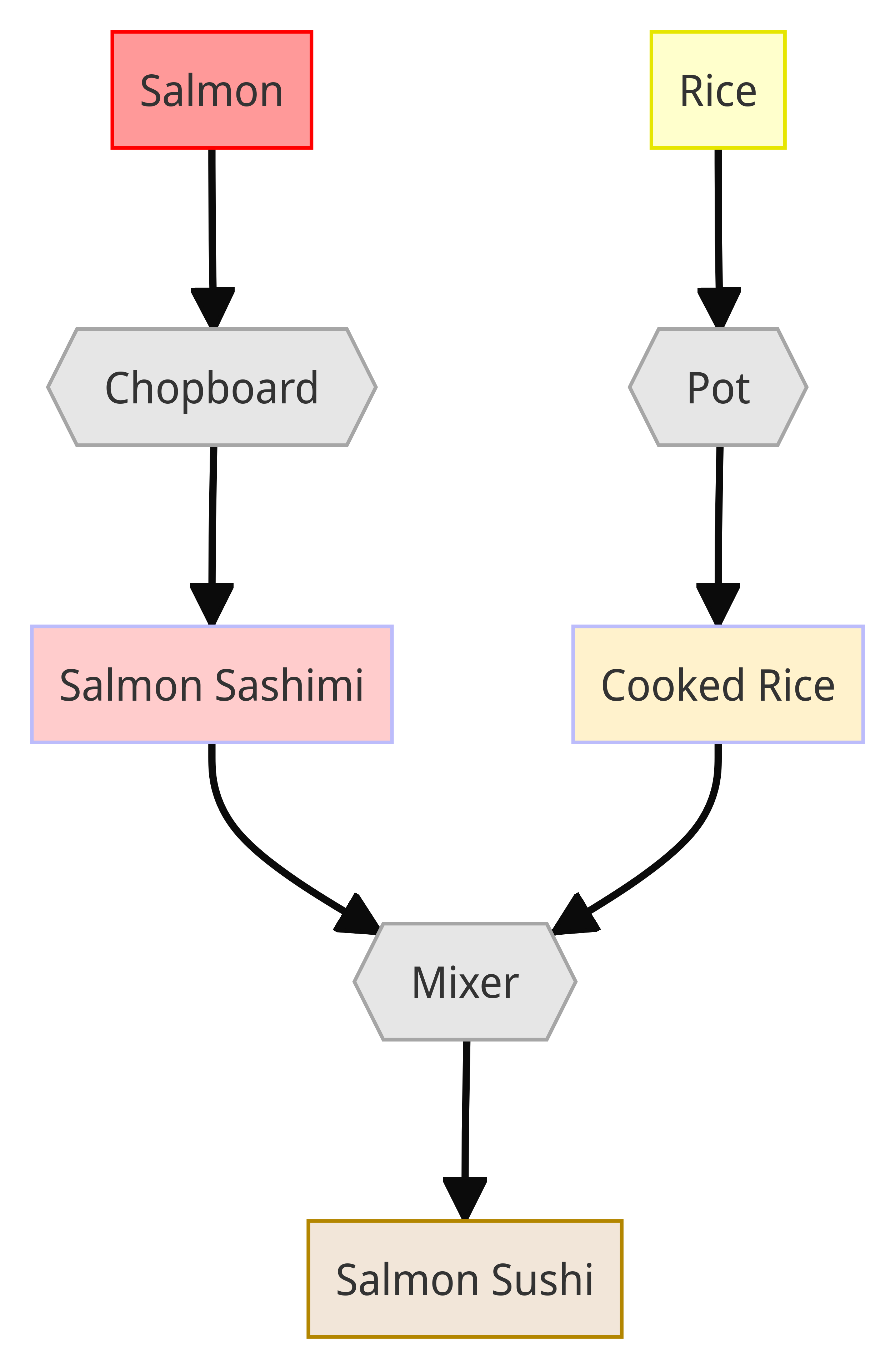}
        \caption{SalmonSushi}
    \end{subfigure}
    \hfill
    \begin{subfigure}{0.3\textwidth}
        \includegraphics[width=\linewidth, height=0.23\textheight]{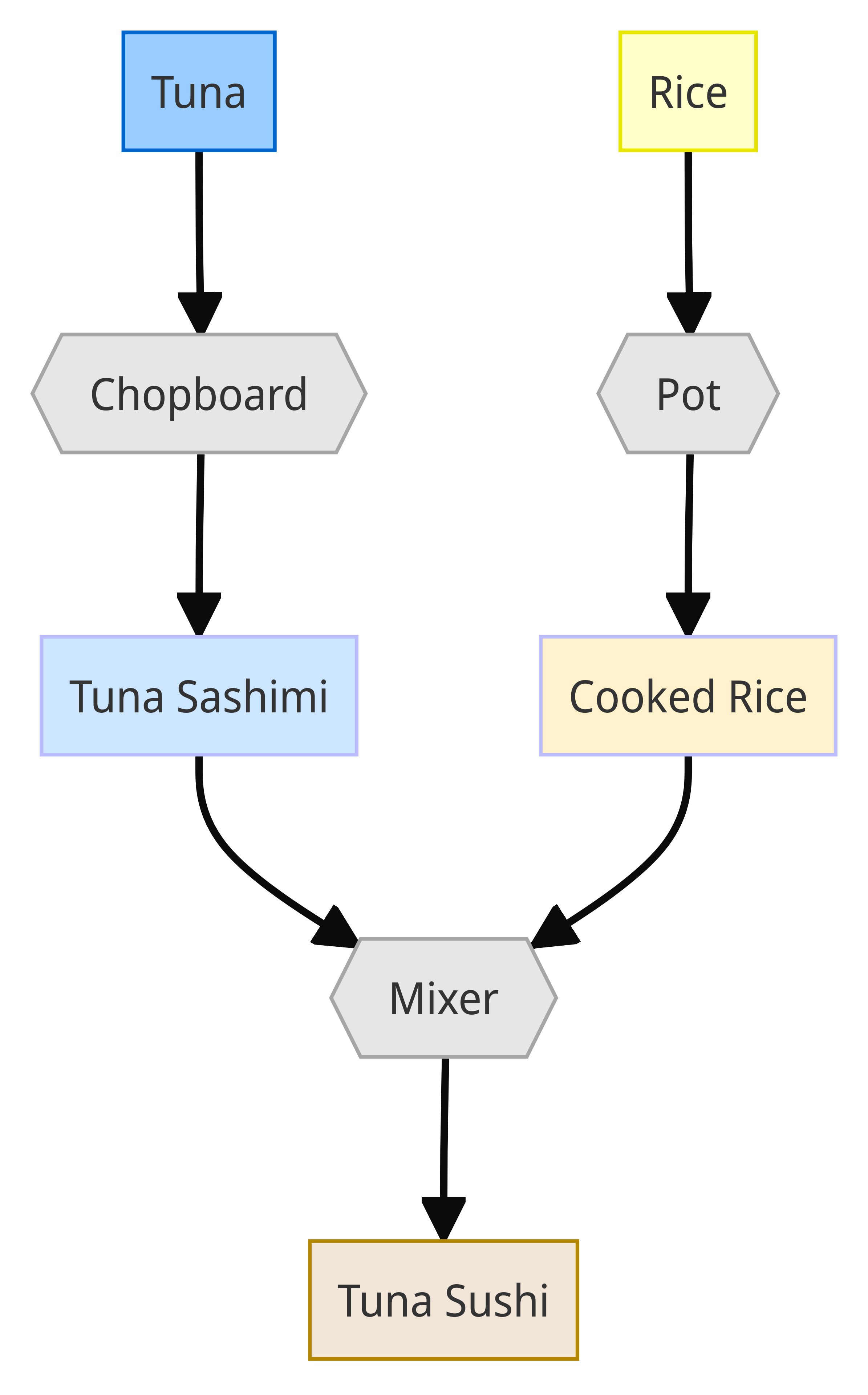}
        \caption{TunaSushi}
    \end{subfigure}
\end{figure}

% Level 12
\subsection{Level 12}
\begin{figure}[H]
    \centering
    \begin{subfigure}{0.3\textwidth}
        \includegraphics[width=\linewidth, height=0.3\textheight]{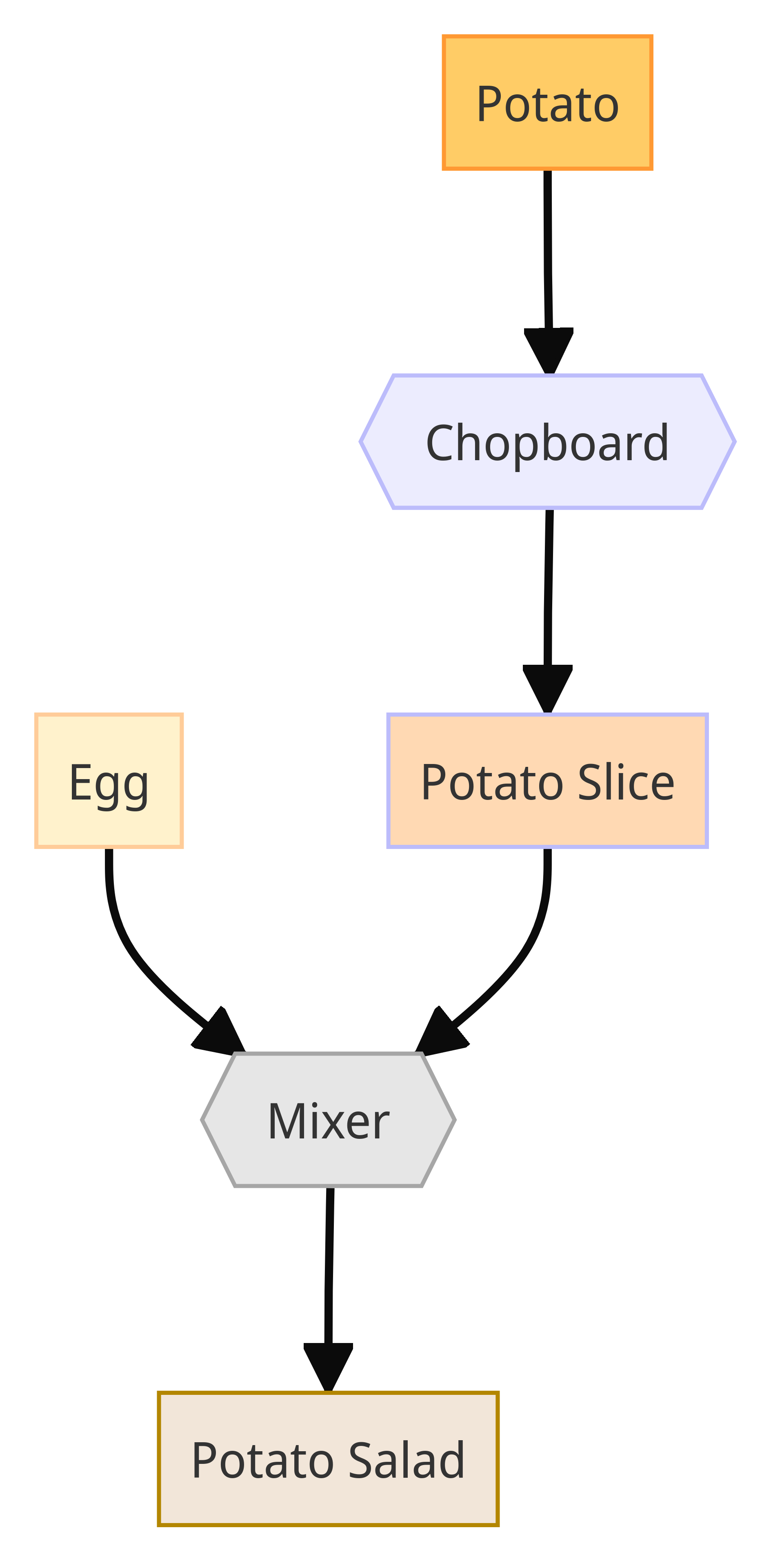}
        \caption{Potato Salad}
    \end{subfigure}
    \hfill
    \begin{subfigure}{0.3\textwidth}
        \includegraphics[width=\linewidth, height=0.35\textheight]{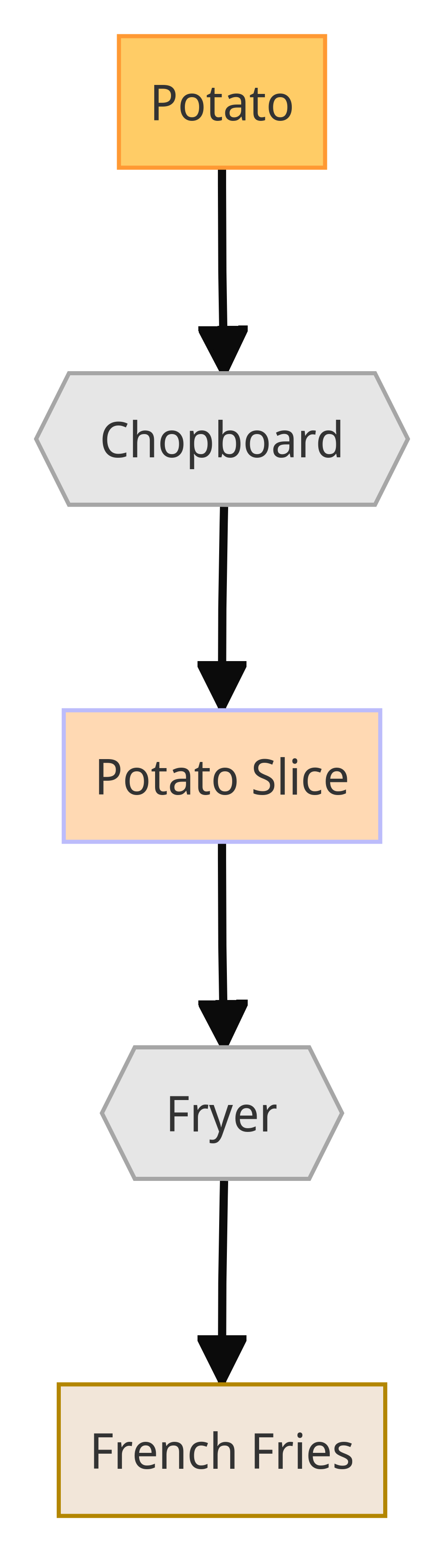}
        \caption{French Fries}
    \end{subfigure}
    \hfill
    \begin{subfigure}{0.3\textwidth}
        \includegraphics[width=\linewidth, height=0.35\textheight]{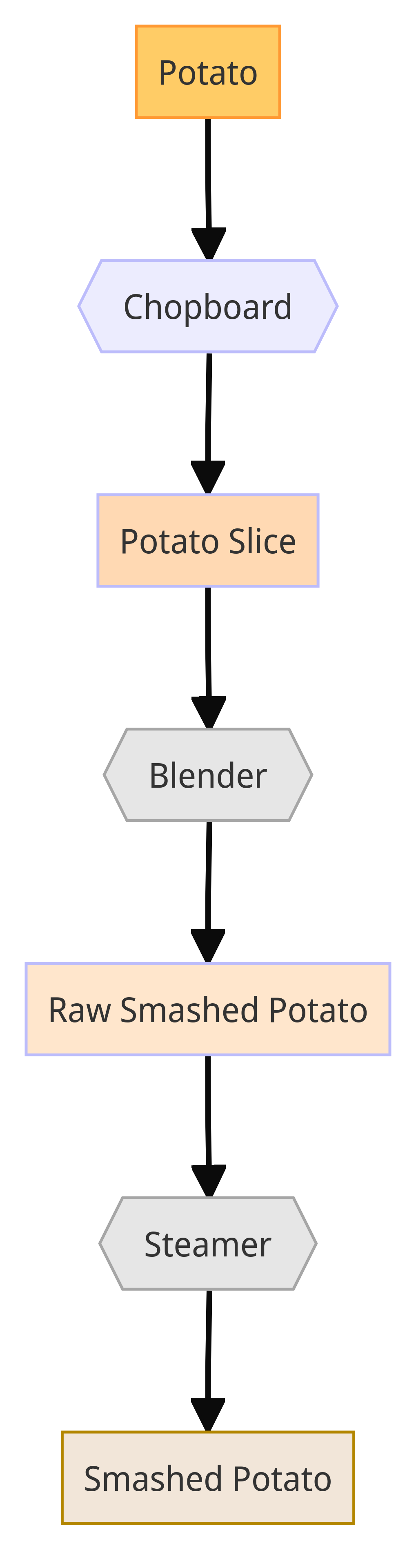}
        \caption{Smashed Potato}
    \end{subfigure}
\end{figure}

\pagebreak
\section{Minecraft}
Here we visualize the task graphs for different tasks in Minecraft.
\begin{figure}[H]
    \centering
    \begin{subfigure}{0.6\textwidth}
        \includegraphics[width=\linewidth]{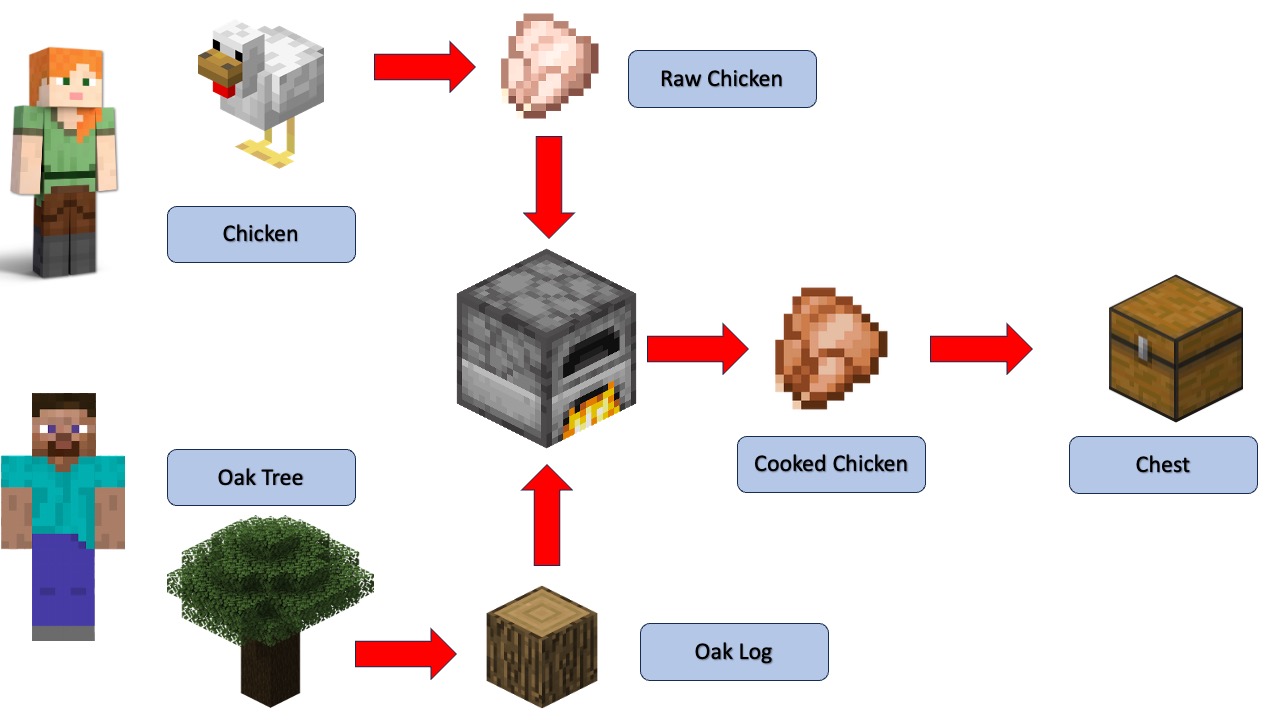}
        \caption{Cooking chicken in Minecraft}
    \end{subfigure}
    \hfill
    \centering
    \begin{subfigure}{0.6\textwidth}
        \includegraphics[width=\linewidth]{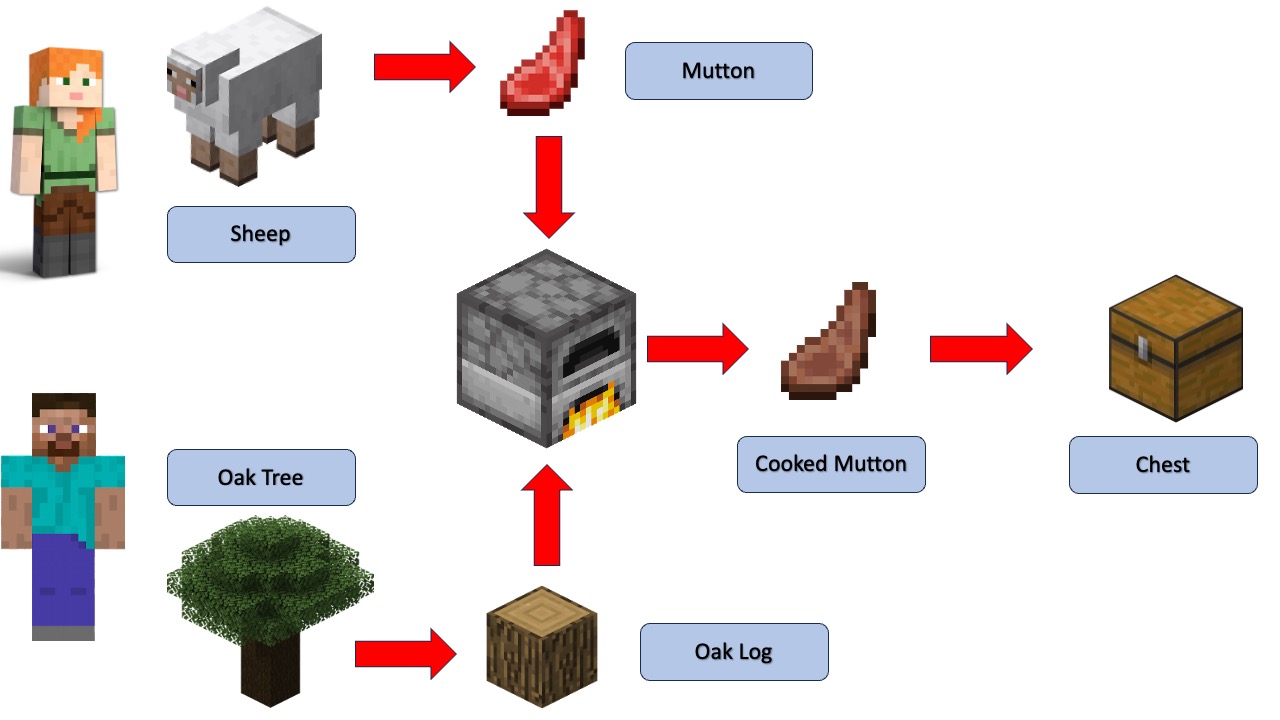}
        \caption{Cooking mutton in Minecraft}
    \end{subfigure}
    \hfill
    \centering
    \begin{subfigure}{0.6\textwidth}
        \includegraphics[width=\linewidth]{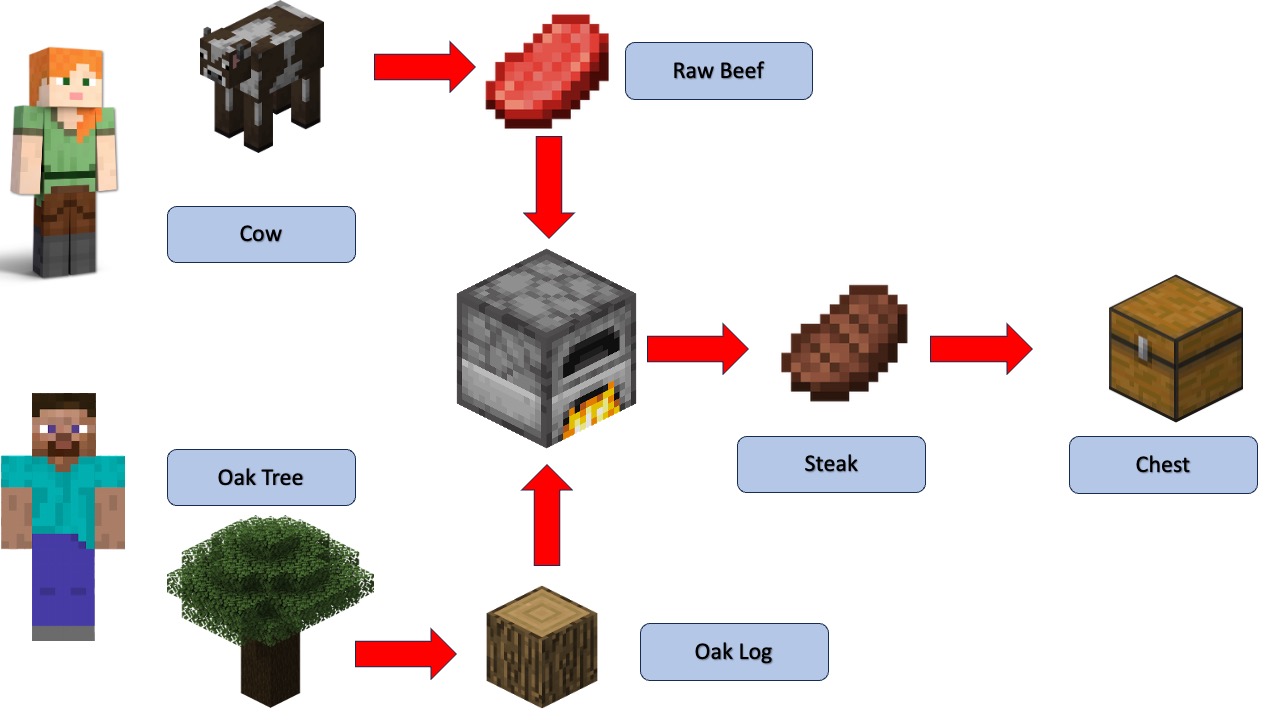}
        \caption{Cooking steak in Minecraft}
    \end{subfigure}
    \hfill
    \centering
    \begin{subfigure}{0.6\textwidth}
        \includegraphics[width=\linewidth]{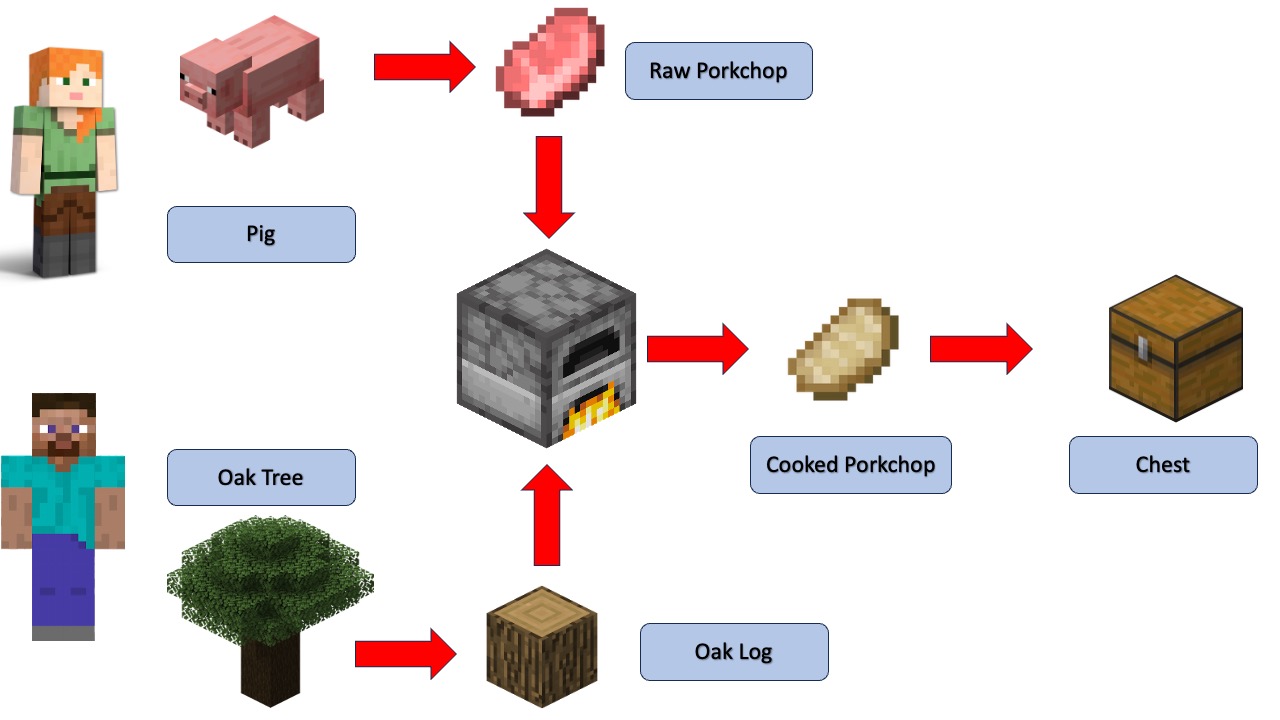}
        \caption{Cooking porkchop in Minecraft}
    \end{subfigure}
\end{figure}

\pagebreak
\section{Human Evaluation Interface}
We use the human evaluation interface to test the human's perception of collaborative agents. This gives us a more controlled environment so users' perception of collaborative agents does not depend on their ability to control the keyboard and mouse, and their perception of collaborative agents does not depend on the latency and rate limits of GPT-4.
\begin{figure}[H]
    \begin{subfigure}{0.46\textwidth}
        \includegraphics[width=\linewidth]{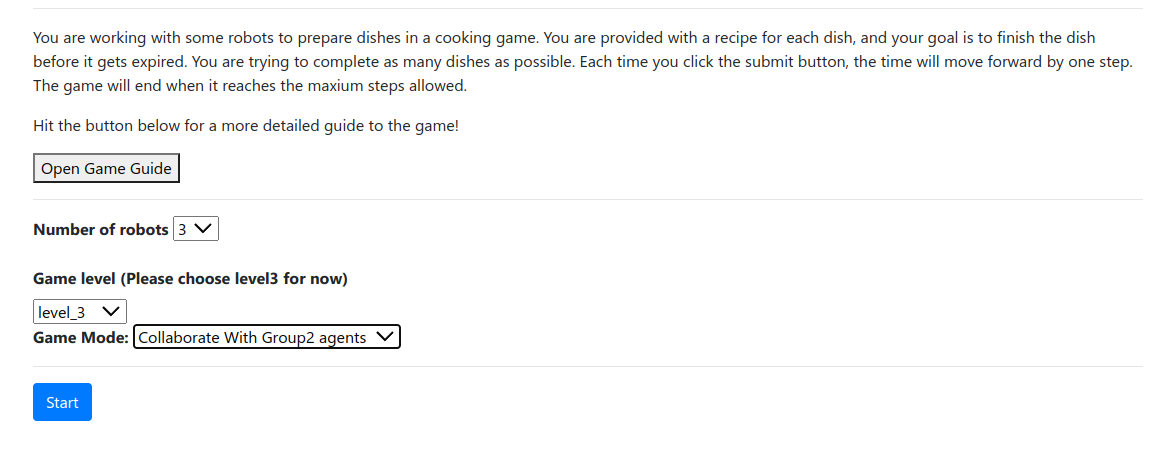}
        \caption{Welcom Screen for human evaluation}
    \end{subfigure}
    \hfill
    \begin{subfigure}{0.46\textwidth}
        \includegraphics[width=\linewidth]{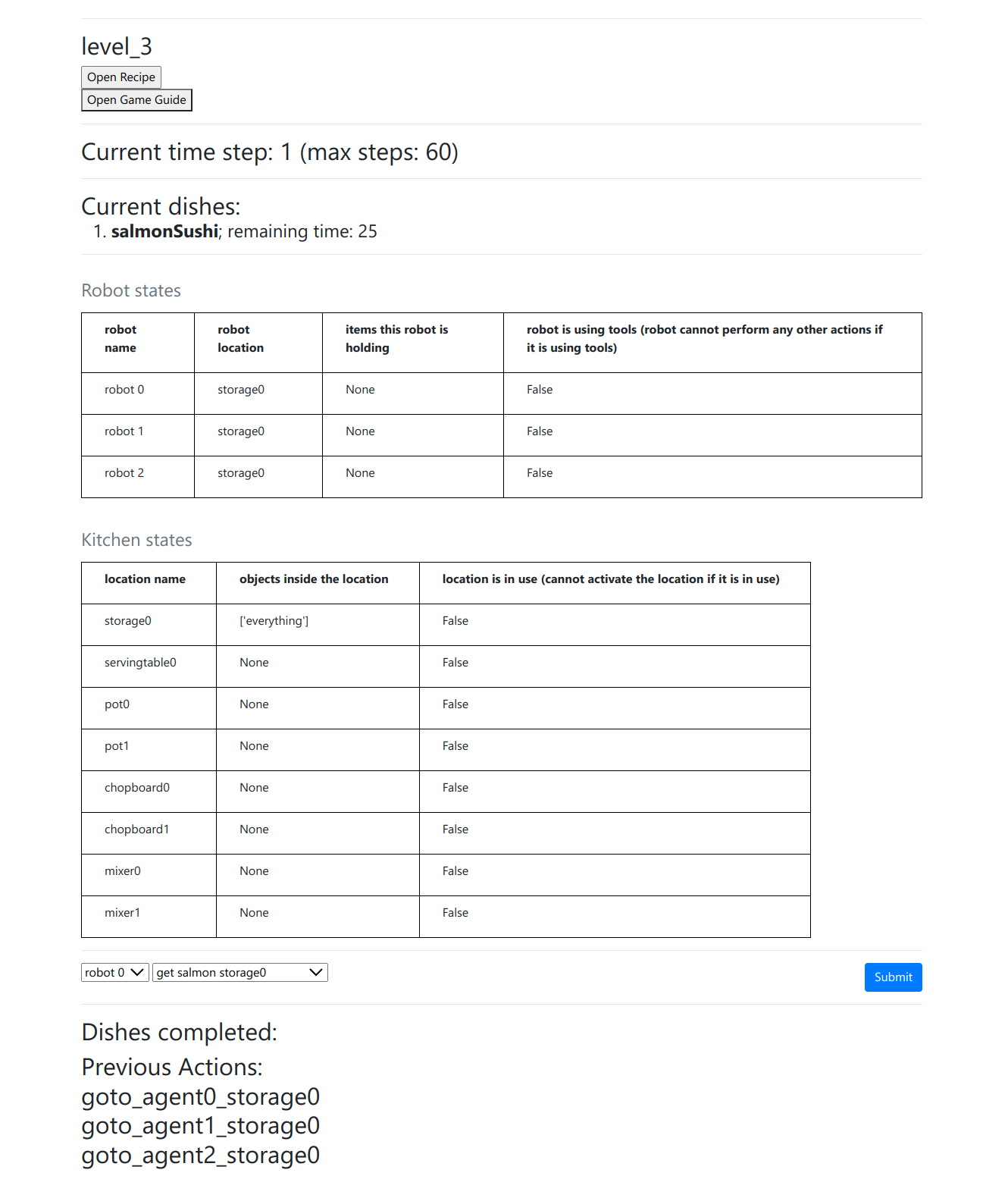}
        \caption{Human Evaluation Example}
    \end{subfigure}
    \hfill
    \begin{subfigure}{0.46\textwidth}
        \includegraphics[width=\linewidth]{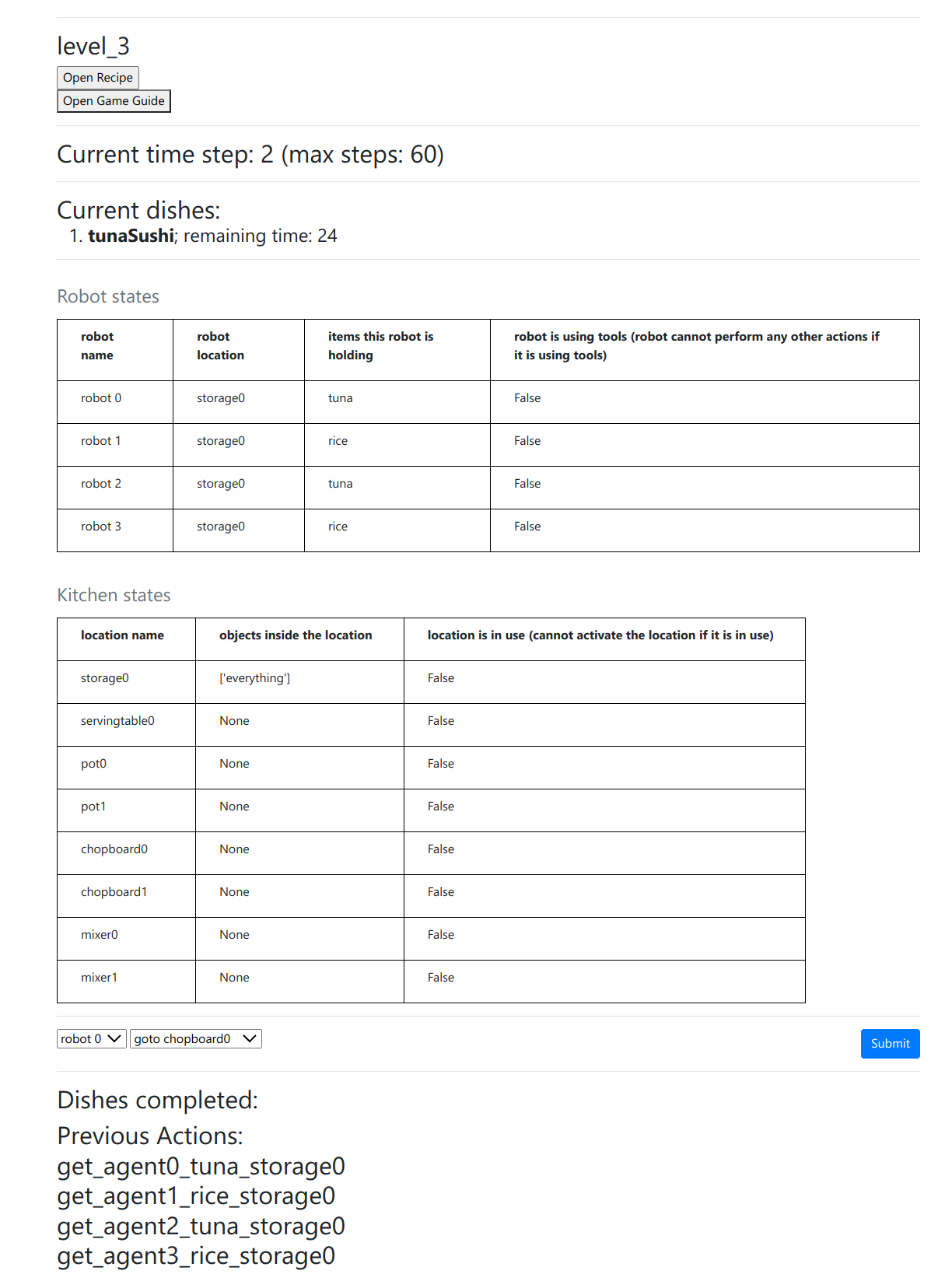}
        \caption{Human Evaluation Example}
    \end{subfigure}
    \hfill
    \begin{subfigure}{0.46\textwidth}
        \includegraphics[width=\linewidth]{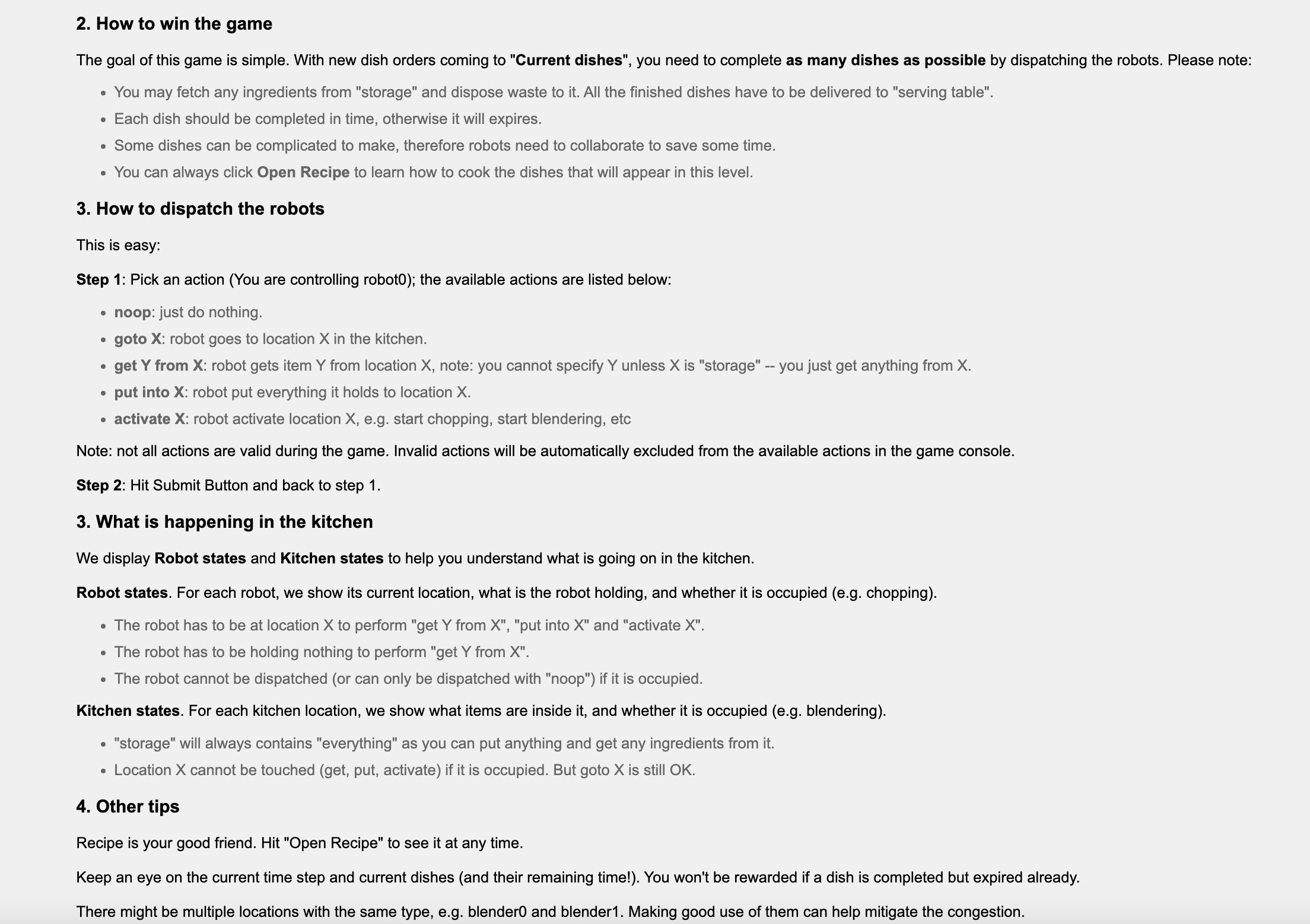}
        \caption{Human Instructions}
    \end{subfigure}
    
\end{figure}

\end{document}

%% file: math_commands.tex
\usepackage{amsmath,amsfonts,bm}

\def\eqref#1{equation~\ref{#1}}
\def\1{\bm{1}}

\DeclareMathAlphabet{\mathsfit}{\encodingdefault}{\sfdefault}{m}{sl}
\SetMathAlphabet{\mathsfit}{bold}{\encodingdefault}{\sfdefault}{bx}{n}

\DeclareMathOperator*{\argmax}{arg\,max}

%% file: config.tex
\definecolor{Gray}{gray}{0.9}
\definecolor{mygreen}{rgb}{0.0, 0.5, 0.0}
\definecolor{myred}{rgb}{0.8, 0.25, 0.33}
\definecolor{myblue}{rgb}{0.19, 0.55, 0.91}
\definecolor{uclablue}{rgb}{0.15, 0.45, 0.68}
\definecolor{boxgreen}{rgb}{0.02, 0.66, 0.02}
\definecolor{boxred}{rgb}{0.66, 0.1, 0.1}
\definecolor{boxblue}{rgb}{0.01, 0.01, 0.73}
\usepackage{url}
\usepackage{etoc}
\usepackage{amsmath}
\usepackage{booktabs}
\usepackage{algorithm}
\usepackage{algorithmic}
\usepackage{lipsum}
\usepackage{pifont}
\usepackage{wrapfig}
\usepackage{colortbl}
\usepackage{acronym}
\usepackage{multicol}
\usepackage{multirow}
\usepackage{makecell}
\usepackage{enumitem}
\usepackage{tabularx,colortbl,array}
\usepackage[skip=3pt,font=small]{subcaption}
\usepackage[skip=3pt,font=small]{caption}
\usepackage[pagebackref,breaklinks,citecolor=uclablue,colorlinks=true,linkcolor=black]{hyperref}
\usepackage{xspace}
\usepackage{mdframed}

\usepackage{mathtools}
\usepackage{amssymb}
\usepackage{amsthm}

\renewcommand{\paragraph}[1]{\noindent\textbf{#1.}}
\newcommand{\supp}{\textit{appendix}}

\makeatletter
\DeclareRobustCommand\onedot{\futurelet\@let@token\@onedot}
\def\@onedot{\ifx\@let@token.\else.\null\fi\xspace}
\def\eg{\emph{e.g}\onedot} 

\def\ie{\emph{i.e}\onedot}

\def\etc{\emph{etc}\onedot} 

\def\wrt{w.r.t\onedot}

\makeatother

\acrodef{llms}[LLMs]{Large Language Models}
\acrodef{mlms}[MLMs]{Multimodal Language Models}
\newcommand{\overcook}{\textsc{CuisineWorld}\xspace}
\newcommand{\loc}{\texttt{location}\xspace}
\newcommand{\agent}{\texttt{agent}\xspace}
\newcommand{\interval}{$\tau_\text{int}$\xspace}
\newcommand{\lifetime}{$\tau_\text{lft}$\xspace}
\newcommand{\colab}{\textbf{CoS}\xspace}
\newcommand{\mindagent}{\textsc{MindAgent}\xspace}

\usepackage{xcolor}

\definecolor{mygray}{gray}{0.4}
\newcommand{\cmark}{\color{mygray}\ding{51}}%
\newcommand{\xmark}{\color{mygray}\ding{55}}%